\documentclass[acmtog]{acmart}
\usepackage{multirow}
\usepackage{xcolor}
\usepackage{soul}

\usepackage{subfig}

\hypersetup{pdfstartview={FitH}}

\makeatletter
\def\blfootnote{\xdef\@thefnmark{}\@footnotetext}
\makeatother

\AtBeginDocument{%
  \providecommand\BibTeX{{%
    \normalfont B\kern-0.5em{\scshape i\kern-0.25em b}\kern-0.8em\TeX}}}

\acmSubmissionID{427}

\begin{document}

\title{Controllable Group Choreography using Contrastive Diffusion}


\author{Nhat Le}
\email{nhat.minh.le@aioz.io}
\orcid{https://orcid.org/0009-0007-1122-4981}
\affiliation{%
  \institution{AIOZ}
  \country{Singapore}
}
\author{Tuong Do}
\email{tuong.khanh-long.do@aioz.io}
\orcid{https://orcid.org/0000-0002-3290-3787}
\affiliation{%
  \institution{AIOZ}
  \country{Singapore}
}
\author{Khoa Do}
\email{19110348@student.hcmus.edu.vn}
\orcid{https://orcid.org/0009-0008-3819-1312}
\affiliation{%
  \institution{VNUHCM-University of Science}
  \country{Vietnam}
}
\author{Hien Nguyen}
\email{hien.nguyen@aioz.io}
\orcid{https://orcid.org/0009-0003-1589-6583}
\affiliation{%
  \institution{AIOZ}
  \country{Singapore}
}
\author{Erman Tjiputra}
\email{erman.tjiputra@aioz.io}
\orcid{https://orcid.org/0009-0003-6909-4623}
\affiliation{%
  \institution{AIOZ}
  \country{Singapore}
}
\author{Quang D.Tran}
\email{quang.tran@aioz.io}
\orcid{https://orcid.org/0000-0001-5839-5875}
\affiliation{%
  \institution{AIOZ}
  \country{Singapore}
}
\author{Anh Nguyen}
\email{anh.nguyen@liverpool.ac.uk}
\orcid{https://orcid.org/0000-0002-1449-211X}
\affiliation{%
  \institution{University of Liverpool}
  \country{United Kingdom}
}

\renewcommand{\shortauthors}{Nhat Le, et al.}


% \twocolumn[{%
% \renewcommand\twocolumn[1][]{#1}%
% \begin{center}
%     \centering
%     \captionsetup{type=figure}
%      \includegraphics[width=\textwidth, keepaspectratio=true]{images/graph/IntroVis_v3.pdf}
%     \captionof{figure}{Test caption}
% \end{center}%
% }]
\begin{teaserfigure}
    \centering
\begin{center}
  \centering
  \captionsetup{type=figure}
  \Large
\resizebox{\linewidth}{!}{
\setlength{\tabcolsep}{2pt}
\begin{tabular}{ccccc}
\shortstack{\includegraphics[width=0.33\linewidth]{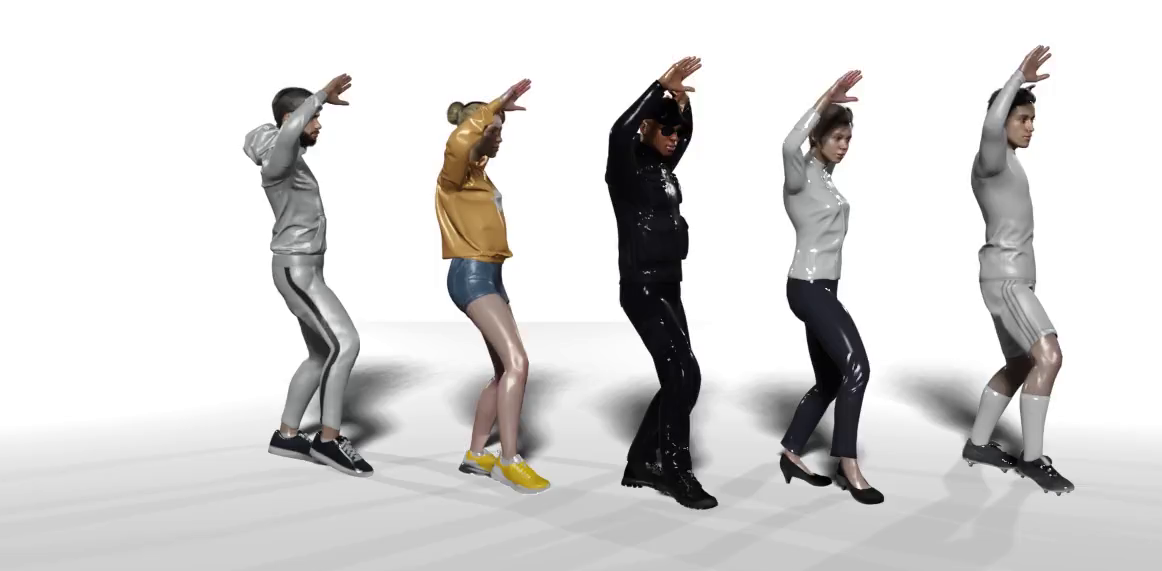}}&
\shortstack{\includegraphics[width=0.33\linewidth]{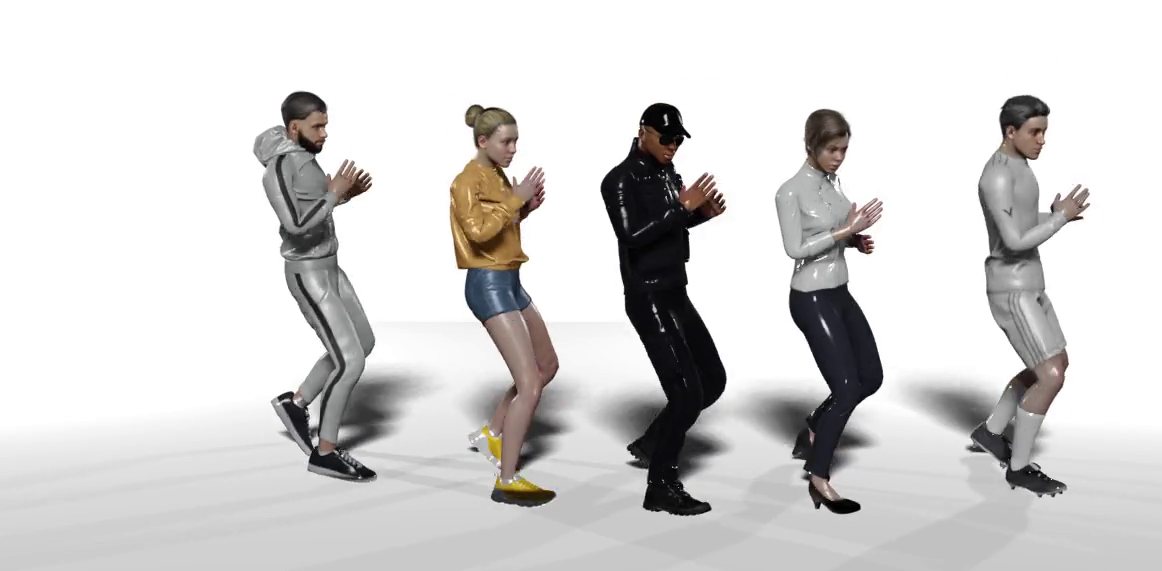}}&
\shortstack{\includegraphics[width=0.33\linewidth]{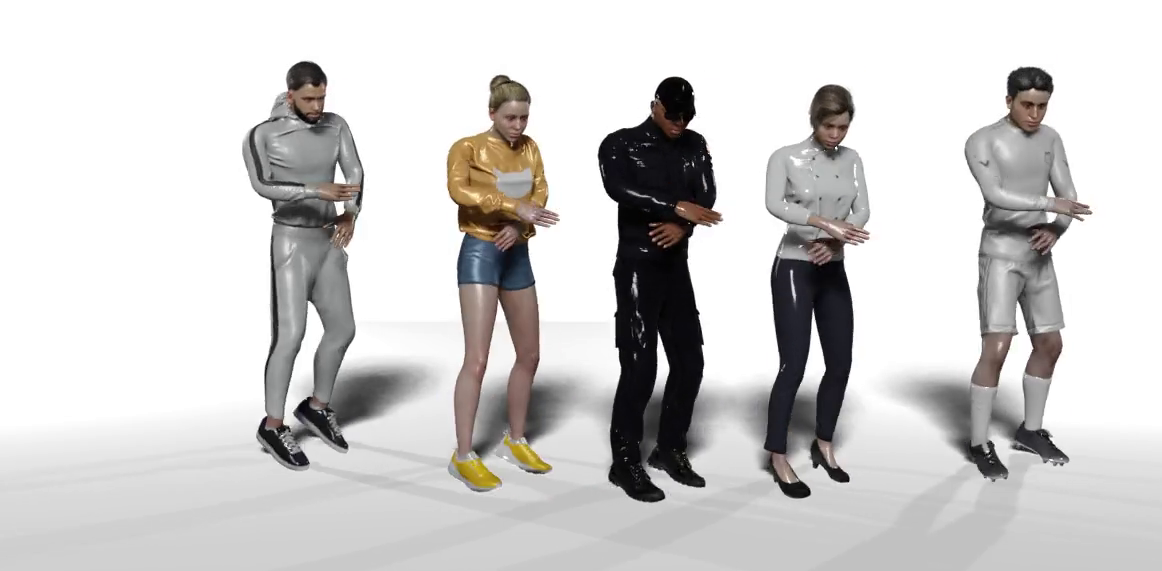}}&
\shortstack{\includegraphics[width=0.33\linewidth]{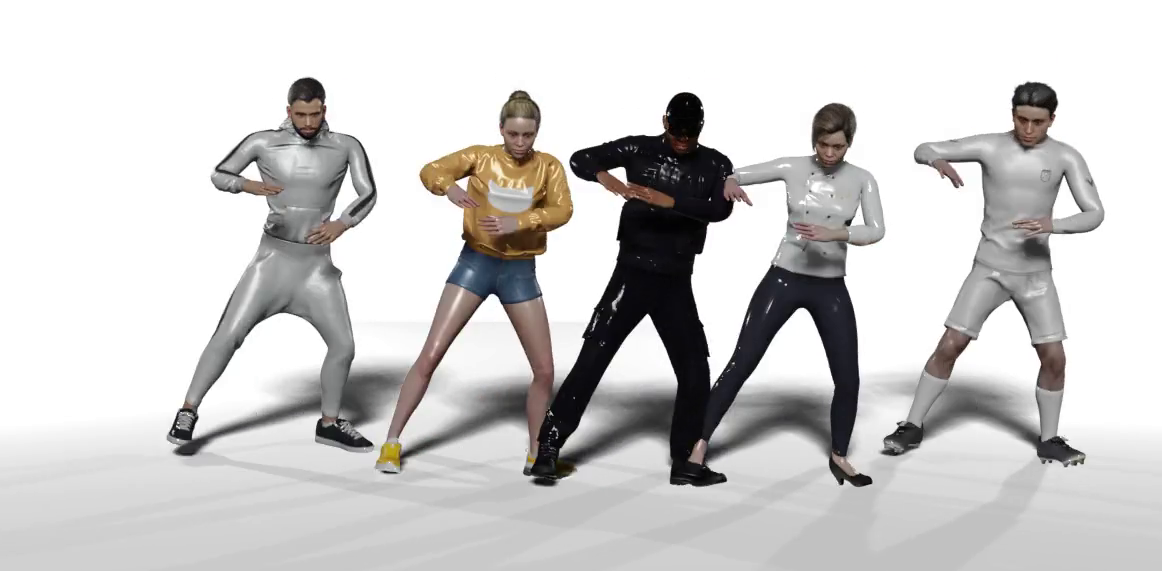}}&
\shortstack{\includegraphics[width=0.33\linewidth]{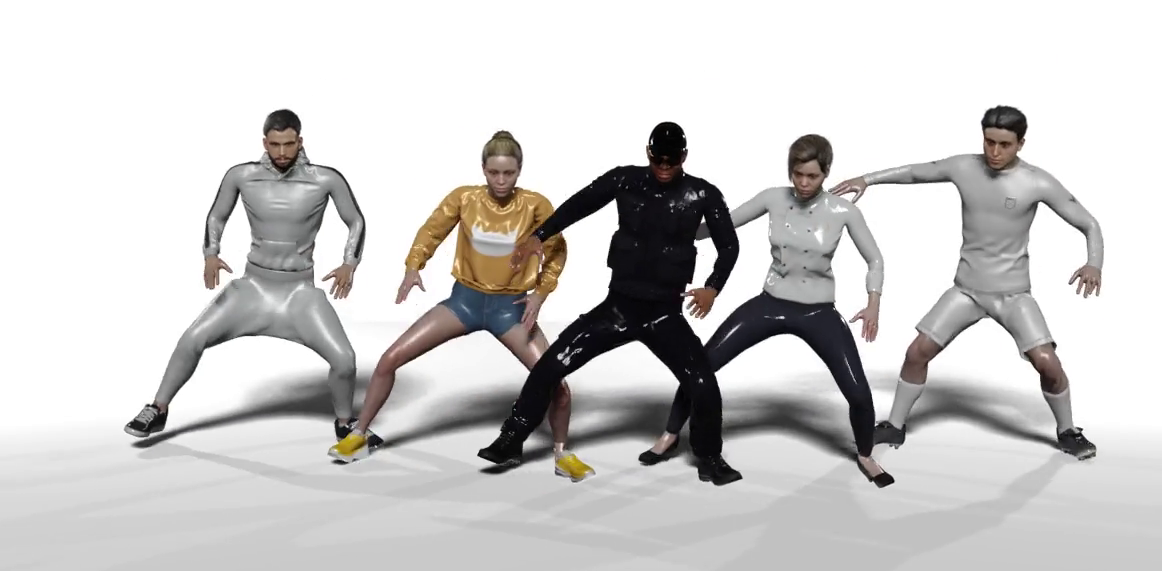}}\\[15pt]
% \hline\\
\shortstack{\includegraphics[width=0.33\linewidth]{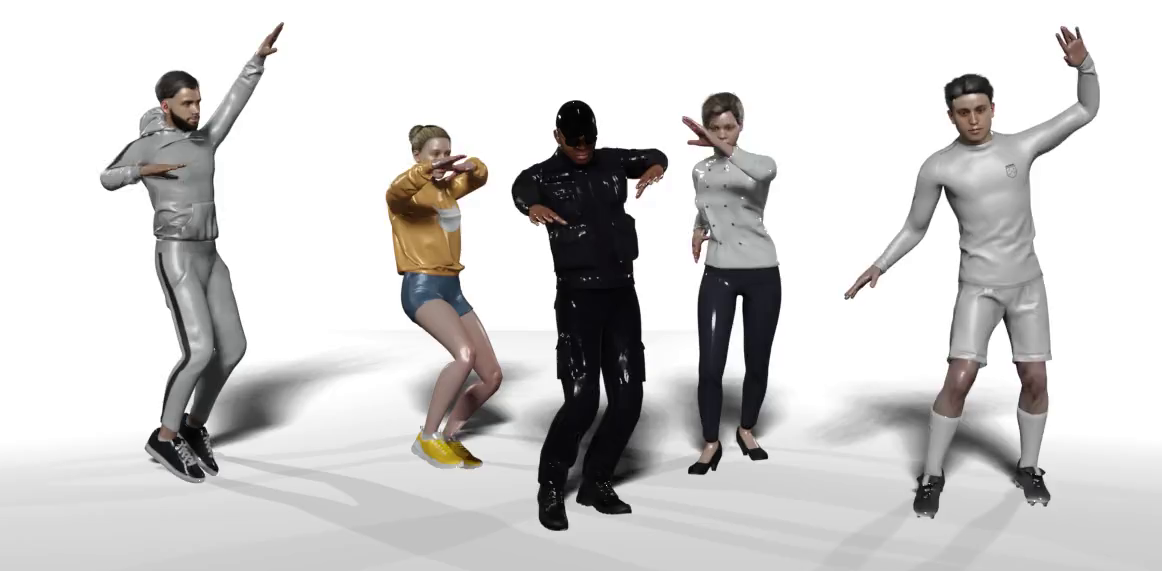}}&
\shortstack{\includegraphics[width=0.33\linewidth]{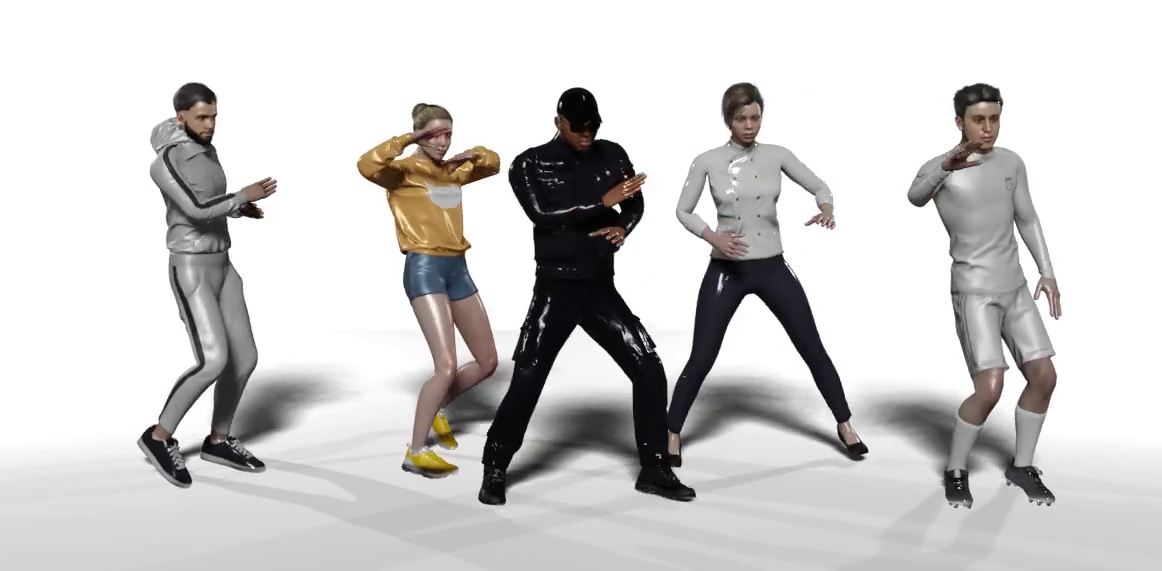}}&
\shortstack{\includegraphics[width=0.33\linewidth]{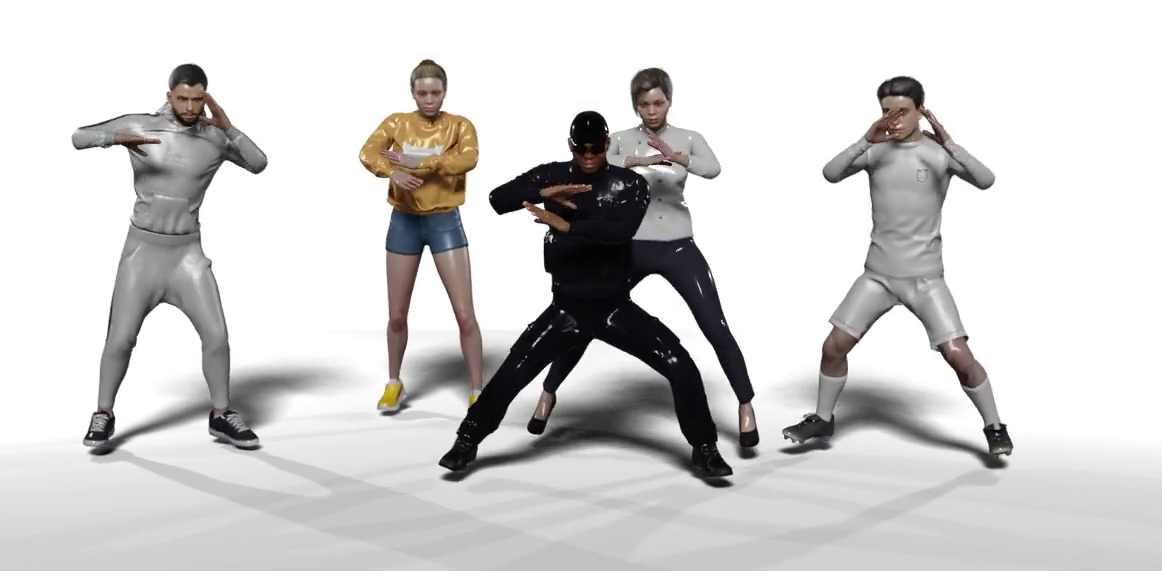}}&
\shortstack{\includegraphics[width=0.33\linewidth]{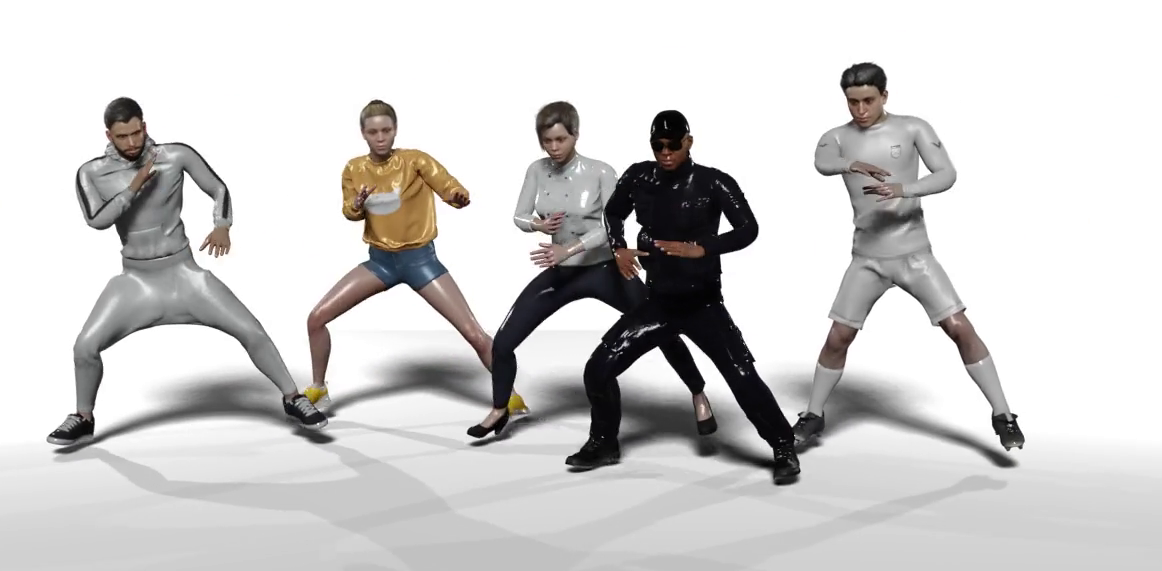}}&
\shortstack{\includegraphics[width=0.33\linewidth]{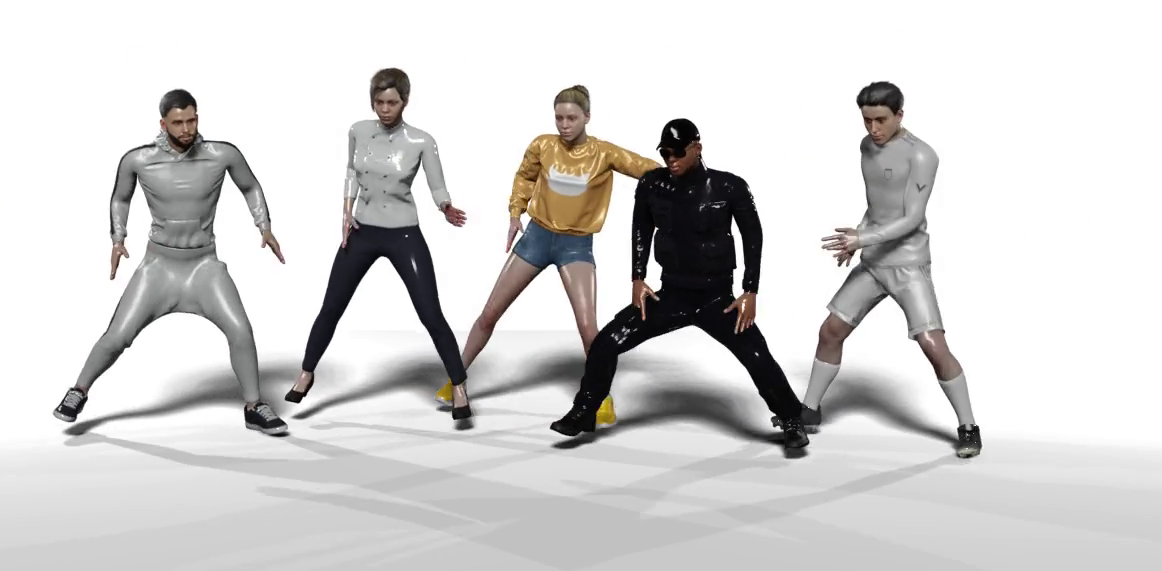}}
% \hline\\
\end{tabular}
}
%\vspace{3ex}
    \captionof{figure}{We present a contrastive diffusion method that controls the consistency (top row) and diversity (second row) in group choreography.}
\end{center}%
\label{fig:introvis}
\end{teaserfigure}

\begin{abstract}
Music-driven group choreography poses a considerable challenge but holds significant potential for a wide range of industrial applications. The ability to generate synchronized and visually appealing group dance motions that are aligned with music opens up opportunities in many fields such as entertainment, advertising, and virtual performances. However, most of the recent works are not able to generate high-fidelity long-term motions, or fail to enable controllable experience. In this work, we aim to address the demand for high-quality and customizable group dance generation by effectively governing the consistency and diversity of group choreographies. In particular, we utilize a diffusion-based generative approach to enable the synthesis of flexible   {number of dancers} and long-term group dances, while ensuring coherence to the input music. Ultimately, we introduce a Group Contrastive Diffusion (GCD) strategy to enhance the connection between dancers and their group, presenting the ability to control the consistency or diversity level of the synthesized group animation via the classifier-guidance sampling technique. Through intensive experiments and evaluation, we demonstrate the effectiveness of our approach in producing visually captivating and consistent group dance motions. The experimental results show the capability of our method to achieve the desired levels of consistency and diversity, while maintaining the overall quality of the generated group choreography.
\end{abstract}


% \ccsdesc[300]{Machine learning approaches}



\setcopyright{acmlicensed}
\acmJournal{TOG}
\acmYear{2023} \acmVolume{42} \acmNumber{6} \acmArticle{224} \acmMonth{12} \acmPrice{15.00}\acmDOI{10.1145/3618356}

\maketitle
\begin{sloppypar}
\blfootnote{Our project page is available at: \url{https://aioz-ai.github.io/GCD/}}
\section{Introduction}
With the widespread presence of digital social media platforms, the act of creating and editing dance videos has gained immense popularity among social communities. This surge in interest has resulted in the daily production and watching of millions of dancing videos across online platforms~\cite{FINK2021351, Visualization-folk-dance}. Recently, researchers from computer vision, computer graphics, and machine learning communities have devoted considerable attention to developing techniques that can generate natural dance movements from music~\cite{bisig2022generative}. These advancements have far-reaching implications and find applications in various domains, such as animation~\cite{li2021AIST++}, the creation of virtual idols~\cite{perez2021_transflower,pham2023style}, the development of virtual meta-verse~\cite{survey-dancing-deep-metaverse}, and dance education~\cite{physic-perform, soga2005automatic, shi2021application}. These techniques empower artists, animators, and educators alike, providing them with powerful tools to enhance their creative endeavors and enrich the dance experience for both performers and audiences.

While significant progress has been made in generating dancing motions for single dancer~\cite{li2021AIST++,ferreira2021_learn2dance_gcn,perez2021_transflower,siyao2022_bailando,kim2022brandnew_dance,Dance_Revolution,tseng2022edge_diffusion}, the task of producing cohesive and expressive choreography for a group of dancers has received limited attention~\cite{le2023music}. The generation of synchronized group dance motions that are both realistic and aligned with music remains a challenging problem in the field of computer animation and motion synthesis~\cite{yalta2019_weaklyrnn,chen2021_choreomaster}. This is primarily due to the complex relationship between music and human motion, the diverse range of motions required for group performances, and the insufficient of a suitable dataset~\cite{le2023music}. 
At present, AIOZ-GDance~\citep{le2023music} stands as the most recent extensive dataset available to facilitate the task of generating group choreography. Besides, while current algorithms can generate individual movements and choreographic sequences, ensuring that these elements align seamlessly with the overall group performance is also paramount~\cite{AISTDanceDB}.

Different from solo dance, group dance involves coordination and interaction between dancers, making it crucial and challenging to establish correlations between motion series within a group~\cite{le2023music}. Besides, group dance can involve complex and diverse choreographies among participating dancers while still maintaining a semantic relationship between the motion and input music. Exploring the \textit{\textbf{consistency}} and \textit{\textbf{diversity}} between the movements of dancers of the synthesized group choreography is of vital importance to create a natural and expressive performance. The ability to control the consistency and diversity in group dance generation holds great potential across various applications~\cite{bisig2022generative}. One such application is in the realm of entertainment and performance. Choreographers and creative teams can leverage this control ability to design captivating group dance routines that seamlessly blend synchronized movements with moments of individual expression. Second, in the context of animation and virtual metaverses, the control over consistency and diversity allows for the creation of visually stunning and immersive virtual dance performances. By balancing the synchronization of dancers' movements, while also introducing variations and unique flourishes, the generated group dances can captivate audiences and evoke a sense of realism and authenticity. Last but not least, in dance education and training, the ability to regulate consistency and diversity in group dance generation can be invaluable. It enables instructors to provide students with a diverse range of generated dance routines and samples that challenge their abilities, promote collaboration, and foster creativity. By dynamically adjusting the level of consistency and diversity, educators can cater to the unique need and skill level of each individual dancer, creating more inclusive instructions and enriching the learning environment~\cite{phillips2009dancing}. Although plenty of applications can be listed, due to some limitations of data establishment~\cite{le2023music}, investigating the consistency and diversity in group choreography has not been carefully explored.

In this paper, our goal is to develop a controllable technique for group dance generation. We present a \textbf{G}roup \textbf{C}ontrastive \textbf{D}iffusion (GCD) strategy that learns an encoder to capture the key targets between group dance movements. Diffusion modeling provides a flexible framework for manipulating the dance distribution, which allows us to modulate the degree of diversity and consistency in the generated dances. By using denoising diffusion probabilistic model~\cite{ho2020ddpm} as a key technique, we can effectively control the trade-off between diversity and consistency during the group dance generation, thanks to the guided sampling process. With this approach, we can guide the generation process toward a desired balance between diversity and consistency levels. Moreover, incorporating the encoder, which learns the association between the dancers and their group, can help to maintain the generated dance moves so that they are consistent with a specific dance style, music genre, or any long-term chorus. We empirically show that this approach has the potential to enhance the quality and naturalness of generated group dance performances, making it more appealing for various applications. 

To summarize, our key contributions are as follows:
\begin{itemize}
    \item We introduce contrastive diffusion, the first denoising diffusion approach for music-driven group choreography. Our model is able to generate high-fidelity and diverse group dancing motions that are aligned with the input music.
    \item We develop a method to trade-off between the consistency and diversity of generated group motions. Our framework allows users to control and generate different outputs from a single piece of input music. 
    \item   {Extensive experiments along with user study evaluations demonstrate state-of-the-art performance of our model in synthesizing group choreography animation, as well as creating long dance motion sequences while maintaining the coherency among dancers.}
\end{itemize}

\section{Related Work}
\label{Sec:relatedwork}
\subsection{Music-driven Choreography} 
Creating natural and authentic human choreography from music is a complex task~\cite{joshi2021extensive}. One commonly employed technique involves using a motion graph derived from a vast motion database to generate new motions~\cite{motion_graph1}. This involves combining various motion segments and optimizing transition costs along the graph path. Alternatively, there are other methods that incorporate music-motion similarity matching constraints to ensure consistency between the motion and the accompanying music~\cite{motion_graph2,motion_graph3}. Previous studies have extensively explored these methodologies~\cite{motion_graph4, motion_graph5, motion_graph6, moion_graph7}.
However, most of these approaches relied on heuristic algorithms to stitch together pre-existing dance segments sourced from a limited music-dance database~\cite{moion_graph7}. While these methods are successful in generating extended and realistic dance sequences, they face limitations when trying to create entirely novel dance fragments~\cite{ofli2011learn2dance}.

In recent years, several signs of progress have been made in the field of music-to-dance motion generation using   {Convolutional Network (CNN)}~\cite{chan2019everybody,zhuang2020_music2dance, sun2020_deepdance, ye2020_choreonet,ahn2020_autoregressive, yin2022dance},   {Recurrent Network (RNN)}~\cite{tang2018_dancemelody, sun2020_deepdance, alemi2017_groovenet, Dance_Revolution,yalta2019_weaklyrnn},   {Graph Neural Network (GNN)}~\cite{ferreira2021_learn2dance_gcn, ren2020_ssl_gcn,au2022choreograph,zhou2022spatio_temporal_gcn},   {Generative Adversarial Network (GAN)}~\cite{sun2020_deepdance,lee2019_dancing2music}, or Transformer~\cite{siyao2022_bailando, li2021AIST++, li2022_phantomdance, perez2021_transflower,kim2022brandnew_dance,li2022danceformer}. Typically, these methods rely on multiple inputs such as the current music and a brief history of past dance movements to predict the future sequence of human poses. 
  {Recently,  {\citet{gong2023tm2d}} propose an interesting task of generating dance by simultaneously utilizing both music and text instruction. A music-text feature fusion module was designed to fuse the inputs into a motion decoder to generate dance conditioned on both music and text.}
However, although these methods have the potential to produce natural and realistic dancing motion, they are often unable to create synchronized and harmonious movements between multiple dancers~\cite{le2023music}. Ensuring coordination and synchronization between dancers is a complicated problem, as it involves not only individual pose predictions but also the seamless integration of these poses within the context of a group. Achieving synchronized and harmonious group movements requires considering spatial and temporal relationships among dancers, their interactions, and the overall choreographic structure~\cite{AISTDanceDB}. Thus, further advances in the field are considered to address these challenges, including works that use deep learning approaches such as   {Variational Autoencoder (VAE)}~\cite{li2021audio2gestures,hong2022avatarclip}, GAN~\cite{zhu2022quantized}, and Normalising Flow~\cite{perez2021_transflower}.~\citet{zhou2019dance} explore the use of VAE to combine motion data with style embeddings so as to generate diverse and stylistically consistent dance movements. Meanwhile,~\citet{huang2021choreography} introduce a conditional GAN-based approach for generating new dance motions.
\citet{perez2021_transflower} combine a multimodal transformer encoder with a normalising-flow-based decoder to estimate a probability distribution encompassing the potential succeeding poses.
Unfortunately, most of these networks are limited by their ability to model long-term dance sequences (e.g., over $8$ seconds) as the generated sequence may freeze or drift towards the end of the music~\cite{sun2022you}.
  {\mbox{\citet{feng2023robust}} learn the dance movements using unpaired data with music style and motion style exemplars of the same style. To facilitate long-term generation, they apply a motion repeat constraint to predict future frames by attending to the historical motions. Nevertheless, this would limit the flexibility of the model by forcing it to always look into the past.}
  {\mbox{\citet{aristidou2022rhythm}} use the motion motifs (clusters of similar short motion sequences) and motion signatures {\cite{aristidou2018deep_motifs}} to guide the dance synthesis to preserve the global consistency following a specific dance style. Consistency in \textit{single dance} (as mentioned in {\cite{aristidou2022rhythm}}) is related to temporal information of a motion sequence itself, whereas diversity and consistency in \textit{group dance} paradigm are factors between motions of two or more dancers within a period.}

\subsection{Group Choreography} Group choreography and its related problem, 
multi-person motion prediction, have been an active research area with numerous studies addressing the challenges of predicting the behaviors of multiple individuals~\cite{arikan2002interactive,mehta2018single,stergiou2019analyzing,chen2020dynamic,aliakbarian2020stochastic,kiciroglu2022long,khaire2022deep,kim2022conditional,guo2022multi,song2022actformer}. One approach from~\citet{alahi2014socially} utilizes a Markov chain model to jointly analyze the trajectories of several pedestrians and predict their destinations in a given scene. Another method presented in~\cite{adeli2020socially} integrates social interactions and the visual context of the environment to forecast the future motion of multiple individuals. Multi-Range Transformers, introduced by~\citet{wang2021multi}, has the capability to predict the movements of groups with more than ten people engaging in social interactions. Recently,~\citet{le2023music} develop a novel approach that utilizes input music sequences and a set of 3D positions of dancers to generate multiple choreographies with group coherency. 
  {\mbox{\citet{wang2022groupdancer}} present a collaboration system to determine which period the dancers should perform dancing with each other and then produce the corresponding motion sequence for each dancer.}
These aforementioned methods leverage various techniques to capture social interactions~\cite{willis2004human,arikan2002interactive}, spatial dependencies~\cite{mehta2018single,aliakbarian2020stochastic}, and temporal dynamics~\cite{stergiou2019analyzing,chen2020dynamic}, generally aiming to predict accurate and socially plausible future motions for multiple individuals in different scenarios.
However, despite the notable advancements achieved, there remains a demand for further investigation of the correlation between the consistency and diversity of motions within group context~\cite{le2023music}. A deeper understanding of how to attain the optimal balance between consistency and diversity holds the potential to unlock new possibilities for creating group choreographies that benefit the users in many circumstances.

\subsection{Diffusion for Music-driven Choreography}
Recently, diffusion-based approaches have shown remarkable results on several generative tasks~\cite{yang2022survey_diffusion_applications} ranging from image generation~\cite{dhariwal2021classifier_guidance,rombach2022high,nichol2022glide,ramesh2022hierarchical,saharia2022imagen_diffusion}, audio synthesis~\cite{kong2020diff_audio_synthesis1, popov2021diff_audio_synthesis2}, pose estimation~\cite{nguyen2023language}, natural language generation~\cite{nichol2021improved}, and motion synthesis~\cite{tevet2022mdm,dabral2022mofusion,alexanderson2023diffusion,ren2023diffusion}, to point cloud generation~\cite{luo2021diff_pointcloud1,nichol2022diff_pointcloud2}, 3D object synthesis~\cite{xiang20233d,seo2023let,poole2022dreamfusion}, and scene creation~\cite{vuong2023language,sharp2022diffusionnet,huang2023diffusion,zeng2022lion}. Diffusion models have shown that they can achieve high mode coverage, unlike GANs, while still maintaining high sample quality~\cite{ulhaq2022survey_diffusion_for_vision, yang2022survey_diffusion_applications}. This ability makes them an ideal method for the music-to-dance generation task.~\citet{dabral2022mofusion} introduce a denoising diffusion-based framework that enables the generation of extended, realistic, and semantically faithful human motion sequences by considering diverse conditioning contexts (e.g., text or music). \citet{tseng2022edge_diffusion} present an editable dance generation model (EDGE), which exploits the capability of a transformer-based diffusion architecture and a strong music feature extractor, to provide flexible editing capabilities for dance applications. 
  {Most existing diffusion-based approaches for human motion/dance synthesis only focus on generating motion sequences for  {\textit{a single character}}, conditioned on information such as text {\cite{tevet2022mdm,zhang2022motiondiffuse}}, audio {\cite{tseng2022edge_diffusion,alexanderson2023diffusion}}, or both audio and text {\cite{dabral2022mofusion, zhou2023ude_diffusion_motion}}. Different from these prior works, we aim to create  {\textit{group of dancing motions}} from music, which includes coordinating multiple characters, avoiding collisions, and maintaining coherence between them.  In addition to the vanilla diffusion loss term used for training in previous works, our method employs a contrastive learning strategy that directly influences the training of the diffusion reverse process, enhancing the association within the dance group. In terms of controllable generation, while the work of {\citet{tevet2022mdm}} can edit individual motion sequence using text prompts or {\citet{alexanderson2023diffusion}} interpolates different motion styles using classifier-free guidance {\cite{ho2022classifier_free}},  our approach provides the means to control the trade-off between consistency and diversity of group dance through the learned contrastive encoder.}
Recently,~\citet{chopin2023bipartite} propose BiGraphDiff, a diffusion approach based on bipartite graph architecture for text-driven human motion interactions between two persons. 
  {Concurently, {\citet{shafir2023prior_mdm_two_person}} train a small communication block between two pre-trained Motion Diffusion Models {\cite{tevet2022mdm}} to coordinate between two instances for two-person motion generation from text prompts. Limited by the architectural designs, these methods can only synthesize motion for only two persons, while our model is capable of generating group dancing motion with flexible number of dancers.}

A prominent issue of the diffusion approach for motion synthesis is that although it is highly effective in generating diverse samples, injecting a large amount of noise during the sampling process can lead to inconsistent results. This issue is particularly problematic for group dance paradigm. Therefore, our desideratum is to design a group dance generation model with the ability to address both diversity and consistency problem. Typically, the work in~\cite{le2023music} mainly addresses the consistency through a cross-entity attention mechanism, but it is not entirely effective as diversity is overlooked due to the deterministic nature of their main training process. 
Different from them, we devise a contrastive diffusion approach to tackle these issues altogether. Our model is not only able to create numerous distinctive dancing motions while preserving their coherency, but it also has the flexibility to allow the user to freely control the diversity or consistency level.

\begin{figure*}[ht]
    \centering
    \includegraphics[width=1\textwidth, keepaspectratio=true]{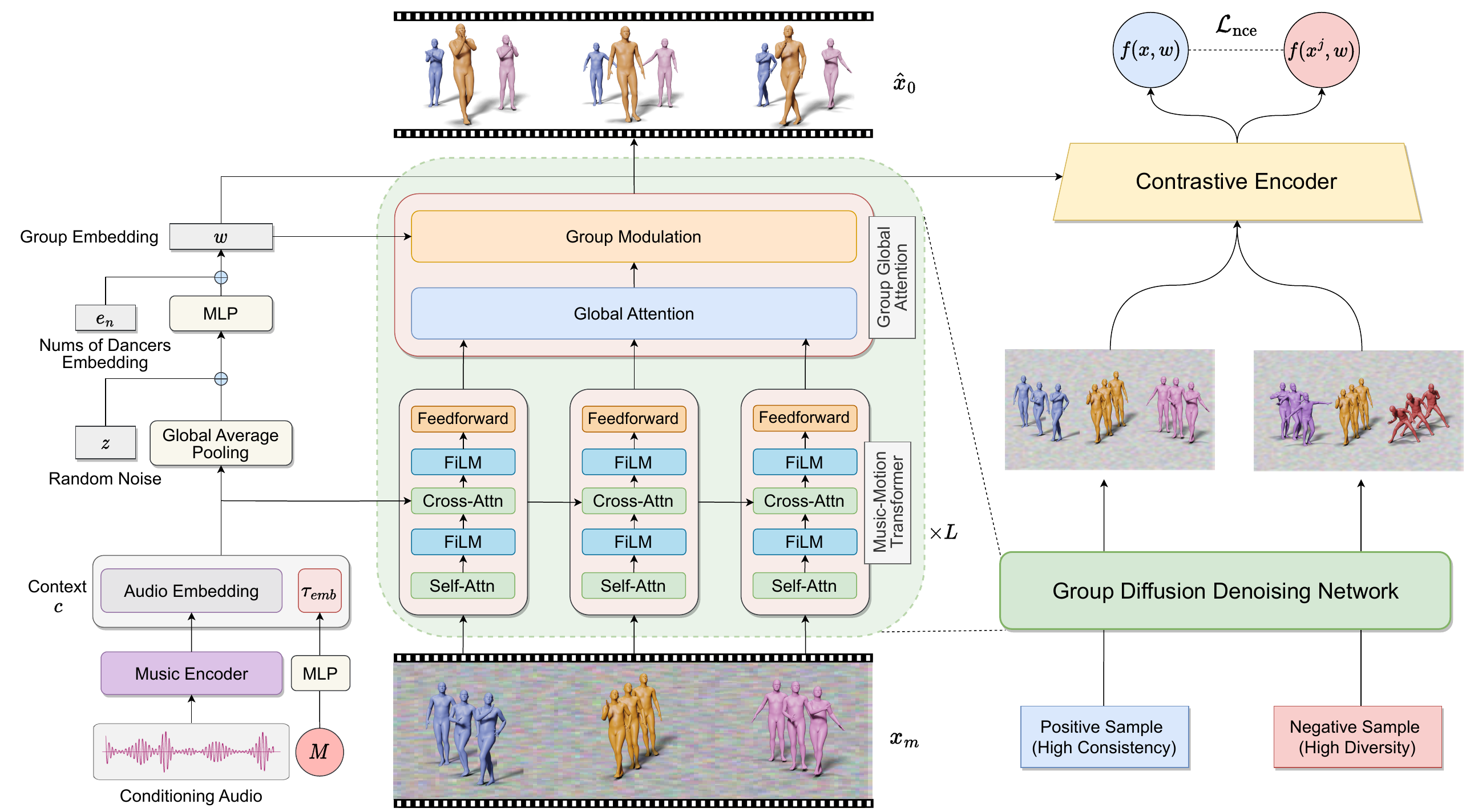}
    \vspace{-3ex}
    \caption{Detailed illustration of our method for group choreography generation. We adopt a transformer architecture to generate the entire sequence all at once. The input of the denoising network is a noisy group motion sample at each step $m$, along with the conditioning music. The model predicts noiseless sample $\hat{x}_0$, which is then diffused back to $x_{m-1}$ to continue the process until reaching $m=0$. We further propose to learn the consistency and diversity of samples through a contrastive learning objective with the Contrastive Encoder. The learned encoder is used as guidance signals to control the generation process.}
    \vspace{-1ex}
    \label{fig:overview}
\end{figure*}
\section{Methodology}

\subsection{Background}
Given an input music sequence $\{a_1, a_t, ...,a_T\}$ with $t = \{1,..., T\}$ indicates the index of the music frames, our goal is to generate the group motion sequences of $N$ dancers: $\{x^1_1,..., x^1_T; ...;x^N_1,...,x^N_T\}$ where $x^i_t$ is the pose of $i$-th dancer at frame $t$. 
We represent dances as sequences of poses in the 24-joint of the SMPL model~\cite{SMPL:2015}, using the 6D continuous rotation~\cite{zhou2019rotation_6d} for every joint, along with a single 3D root translation.
This rotation representation ensures the uniqueness and continuity of the rotation vector, which is more beneficial to the training of deep neural networks. We tackle the group dance generation task by using a diffusion-based framework to synthesize the motions from a random noise distribution, given the music conditioning. Thanks to the sampling process of the diffusion model, we can effectively control the consistency and diversity in the generated sequences.

\textbf{Forward Process of Diffusion Model.} Given an original sample from the real data distribution ${x_0} \sim q({x_0})$, following \cite{ho2020ddpm}, the forward diffusion process is defined as a Markov process that gradually adds Gaussian noise to the data under a pre-defined noise schedule up to $M$ steps.
\begin{equation}
q(x_m | x_{m-1}) = \mathcal{N}(x_m; \sqrt{1-\beta_m} {x_{m-1}}, \beta_mI), \forall m \in \{1,2, ... ,M\}
\end{equation}
If the noise variance schedule $\beta_m$ is small and the number of diffusion step $M$ is large enough, the distribution $q(x_M)$ at the end of the process is well-approximated by a standard normal distribution $\mathcal{N}(0,I)$, which is easy to sample from. Thanks to the nice property of the forward diffusion, we can directly obtain the noised sample at any arbitrary step $m$ without traversing through the whole chain:
\begin{align}
    q(x_m | x_{0}) &= \mathcal{N}(x_m; \sqrt{\bar{\alpha}_m} x_{0} , (1-\bar{\alpha}_m)I), \\
    x_m &= \sqrt{\bar{\alpha}_m} x_{0} + \sqrt{1-\bar{\alpha}_m} \epsilon, \epsilon \sim \mathcal{N}(0, I) \label{eq:nice_forward_sample}
\end{align}
where $\alpha_m = 1-\beta_m$ and $\bar{\alpha}_m = \prod_{s=0}^m \alpha_s$.

\textbf{Reverse Process.} By additionally conditioning on $x_0$, the posterior of the reverse process is tractable and becomes a Gaussian distribution:
\begin{align}
    q(x_{m-1} | x_m, x_0) = \mathcal{N} (x_{m-1}; \tilde{\mu}_m, \tilde{\beta}_mI),
\end{align}
where $\tilde{\mu}_m$ and $\tilde{\beta}_m$ are the posterior mean and variance that depend on both $x_m$ and $x_0$, respectively. We refer the readers to ~\cite{ho2020ddpm} for a detailed derivation of the posterior mean and variance. To obtain a sample from the original data distribution, we start by sampling from the noise distribution $q(x_M)$ and then gradually remove the noise until we reach $x_0$, following the reverse process. Therefore, our goal is to train a neural network to approximate the posterior $q(x_{m-1} | x_m)$ of the reverse process as:
\begin{equation}
    \label{eq:approximate_posterior}
    p_{\theta}(x_{m-1} | x_m) = \mathcal{N}(x_{m-1}; \mu_{\theta}(x_m, m), \Sigma_{\theta}(x_m,m) ) 
\end{equation}
We follow~\cite{ho2020ddpm} to model only the mean $\mu_{\theta}(x_m, m)$ of the reverse distribution while keeping the variance $\Sigma_{\theta}(x_m,m)$ fixed according to the noise schedule. However, instead of predicting the noise $\epsilon_m$ at any arbitrary step $m$ as in their approach, we train the network to learn to predict the original noiseless signal $x_0$. The sample at the previous step $m-1$ can be obtained by noising back the predicted $x_0$ through Equation~\ref{eq:nice_forward_sample}. For conditional generation setting, the network is additionally conditioned with the conditioning signal $c$ as $x_0 \approx \mathcal{G}_\theta(x_m,m,c)$ with model parameters $\theta$.

\subsection{Group Diffusion Denoising Network}
\label{sec:group_denoising_network}

Our model architecture is illustrated in Figure~\ref{fig:overview}.
We utilize a transformer based architecture to generate the whole sequence in one go. Compared with recent auto-regressive approach~\cite{le2023music}, our method does not suffer from the error accumulation problem (i.e., the prediction error accumulates over time since the current-frame outputs are used as inputs to the next frame in the auto-regressive fashion) and thus can generate arbitrary long motion dance sequences without freezing effects~\cite{tseng2022edge_diffusion, petrovich2021action}. 
The input of our network at each diffusion step $m$ is the noisy group sequence $ x_m =\{x^1_{m,1},..., x^1_{m,T}; ...;x^N_{m,1},...,x^N_{m,T}\}$, however, we skip the $m$ index for ease of notation from now on.

\subsubsection{Music-Motion Transformer} Given an input extracted audio sequence $a = \{a_1, a_2, ...,a_T\}$,  we employ a transformer encoder architecture~\cite{vaswani2017attention} to encode the music into the sequence of hidden audio representation $\{c_1, c_2, ...,c_T\}$, which will be used as the conditioning context to the diffusion denoising network. Specifically, we follow the encoder layer as in~\cite{vaswani2017attention} which consists of multi-head self-attention layers and feed-forward layers to effectively encode the multi-scale rhythmic patterns and long-term dependencies between music frames. The diffusion time-step $m$ is also projected to the transformer dimension through a separate Multi-layer Perceptron (MLP) with 3 hidden layers to get the embedding $\tau_{emb}$, then concatenated with the music feature sequence to obtain the final conditioning context $c = \{c_1, c_2, ...,c_T, \tau_{emb}\}$.

\begin{figure}[!t]
 \centering	
\subfloat[]{
	\begin{minipage}[c]{
	   0.23\textwidth}
	   \label{fig_stat_music_genre}
	   \centering
	   \includegraphics[width=1.0\textwidth]{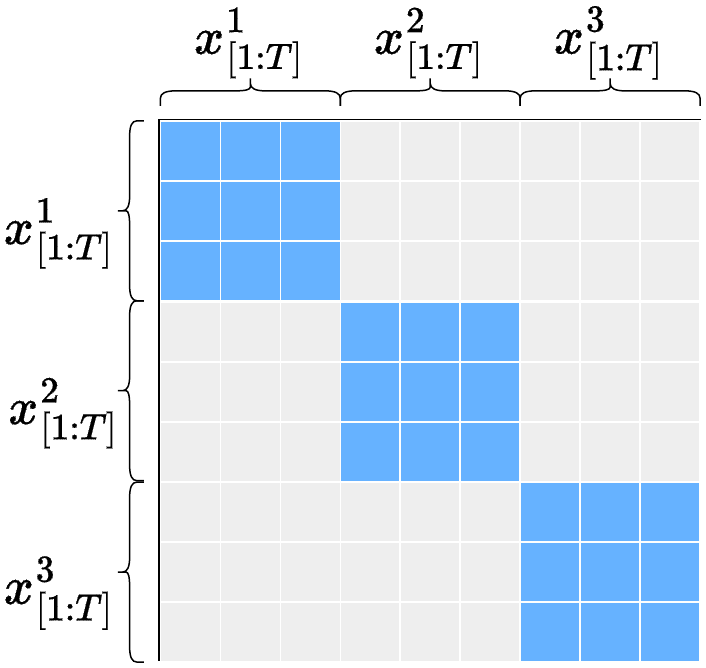}
	\end{minipage}}
 \hspace{0ex}	
  \subfloat[]{
	\begin{minipage}[c]{
	   0.23\textwidth}
	   \label{fig_stat_dance_style}
	   \centering
	   \includegraphics[width=1.0\textwidth]{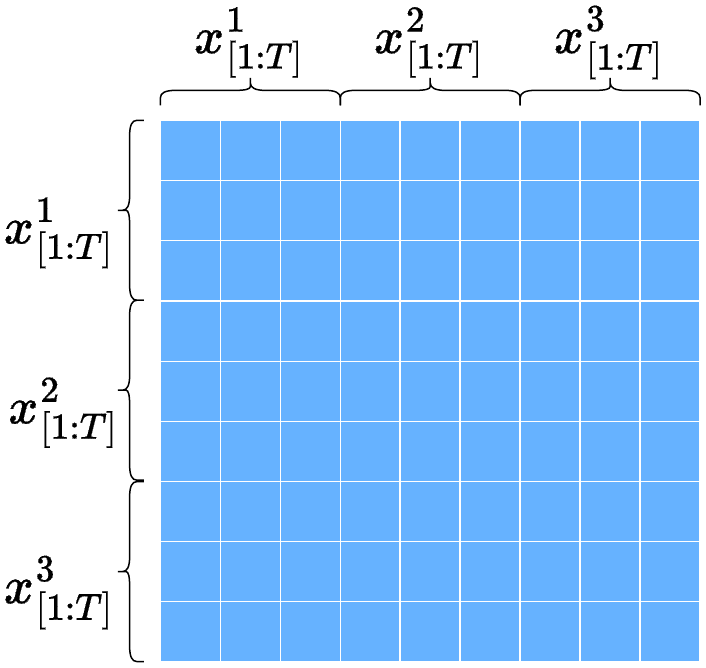}
	\end{minipage}}
\caption{The local attention mask $m_{local}$ (a) and global attention mask $m_{global}$ (b). The blue cell indicates where frames can attend to each other. Blue color represents zero value of the mask while gray color represents minus infinity. $x^i_{[1:T]}$ indicate motion sequence of $i$-th dancer.}
\label{fig:attention_mask}
\end{figure}

Although group choreography incorporates the problem of learning the interaction between dancers, we still need to learn the correlation between the dance movements and the accompanying music audio for each dancer. Therefore, we design the Music-Motion Transformer to essentially focus on learning the direct connection between the motion and the music of \textit{each individual dancer} (and not considering the interconnection among dancers yet). Each frame of the noised input motion $x^i_t$ is projected into the transformer dimension by a linear layer followed by an additive positional encoding~\cite{vaswani2017attention}. Given the whole group sequence including all dancers $\{x^1_1,..., x^1_T; ...;x^n_1,...,x^n_T\}$, we separately encode the motion features of each individual dancer by utilizing the multi-head self-attention~\cite{vaswani2017attention} with masking strategy. We implement the masked self-attention (MSA) mechanism as follows:
\begin{align}
    \label{eq:masked_self_attention}
    \text{MSA}(Q,K,V) = \text{softmax}\left(\frac{{Q}{K}^\top}{\sqrt{d_k}} + {m}_{local} \right) {V}, \\
    {Q} = x {W}^{Q}, \quad {K} = x {W}^{K}, \quad  {V} = x {W}^{V}
\end{align}
where ${W}^{Q}, {W}^{K} \in \mathbb{R}^{d \times d_k}$ and ${W}^{V} \in \mathbb{R}^{d \times d_v}$ are learnable projection matrices to transform the input to query, key, and value, respectively.
${m}_{local}$ is the local attention mask illustrated in Figure~\ref{fig:attention_mask}a. This mask ensures each individual can only attend to their own motion. 
Subsequently, to incorporate the music conditioning context $c$ into each individual motion features, we adopt a transformer decoder architecture~\cite{vaswani2017attention} with cross-attention mechanism (CA)~\cite{vaswani2017attention, tseng2022edge_diffusion, saharia2022imagen_diffusion}, where the motion is the query and the music is the key/value. 
\begin{align}
    \label{eq:cross_attention}
    \text{CA}(\tilde{Q}, \tilde{K}, \tilde{V}) = \text{softmax}\left(\frac{{\tilde{Q}}{\tilde{K}}^\top}{\sqrt{d_k}} \right) {\tilde{V}}, \\
    {\tilde{Q}} = \tilde{x} {\tilde{W}}^{Q}, \quad {\tilde{K}} = c {\tilde{W}}^{K}, \quad {\tilde{V}} = c {\tilde{W}}^{V}
\end{align}
where $\tilde{x}$ is the output activation of the MSA block, and ${\tilde{W}}^{Q}$, ${\tilde{W}}^{K}$, ${\tilde{W}}^{V}$ are the learnable projection matrices that have similar behavior to the MSA mechanism.

\subsubsection{Group Global Attention} To ensure the coherency and non-collision in the movements of \textit{all dancers} within the group, such that their dances should correlate with each other under the music condition instead of dancing asynchronously  {, we} first perform global attention via a masked attention mechanism similar to Equation~\ref{eq:masked_self_attention} with a full masking strategy $m_{global}$. The attention mask is illustrated in Figure~\ref{fig:attention_mask}b. It allows a dancer to fully attend to all other dancers under the global receptive field. Then, we propose the Group Modulation to enforce the group constraints within the group embedding information.

Inspired by StyleGan~\cite{karras2019stylegan}, in which the synthesized image can be manipulated via a latent style vector, we aim to learn a group embedding information from the input music in order to control the group dance generation process. We first apply temporal average pooling to the encoded music feature sequence to obtain a compact representation of the input music $\bar{c} = \frac{1}{T}\sum_{t=1}^T c_t$. To increase the variation and diversity of the group information (i.e., avoid limiting the group embedding to only one style of the input music), we inject a random noise drawn from a standard gaussian distribution $z \sim \mathcal{N}(0,I)$ into $\bar{c}$. We use an 8-layer MLP to learn a mapping from the audio representation to the group embedding. We also add a learnable embedding token $e_{n}$ from a variable-size lookup table $E \in \mathbb{R}^{N\times D}$ up to $N$ maximum dancers, to represent the variation of dancers in the sequence since each sequence may contain different number of dancers. In summary, the process can be written as follows:
\begin{equation}
    \label{eq:mapping_network}
    w = \text{MLP}\left(z + \frac{1}{T}\sum_{t=1}^T c_t\right) + e_{n}, \quad z \sim \mathcal{N}(0,I)
\end{equation}

\textbf{Group Modulation.}
To better apply the group information constraints to the learned hidden features of the dancers, we adopt a Group Modulation layer that learns to adaptively influence the output of the transformer attention block by applying an affine transformation to the intermediate features based on the group embedding $w$. More specifically, we utilize two separate linear layers to learn the affine transformation parameters $\{S(w); b(w)\} \in \mathbb{R}^d$ from the group embedding $w$. The predicted affine parameters are then used to modulate the activations sequence $h = \{h^1_1\dots h^1_T;\dots;h^N_1\dots h^N_T\}$ as follows:
\begin{equation}
    \tilde{h} = S(w) * \frac{h-\mu(h)}{\sigma(h)}+ b(w)
\end{equation}
where each channel of the whole activation sequence is first normalized separately by calculating the mean $\mu$ and $\sigma$, and then scaled and biased using the outputted affine parameters $S(w)$ and $b(w)$. Intuitively, this operation shifts the activated hidden motion features of each individual motion towards a unified group representation  to further encourage the association between them. 
Finally, the output features are then projected back to the original motion dimensions via a linear layer, to obtain the predicted outputs $\hat{x}_0$.

\subsection{Contrastive Diffusion for Controllable Group Dance}
\label{sec:contrastive_approach}

\subsubsection{Contrastive Diffusion} 
We follow 
~\cite{oord2018contrastive_cpc,zhu2022diffusion_contrastive}  to learn the representations that encode the underlying shared
information between the group embedding information $w$ and the group sequence $x$.  Specifically, we model a density ratio that preserves the mutual information between $x$ and $w$ as:
\begin{equation}
    f(x, w) \propto \frac{p({x}|{w})}{p({x})}
\end{equation}
$f(\cdot)$ is a model (i.e., a neural network) to predict a positive score (how well $x$ is related to $w$) for a pair of $({x}, {w})$.

To enhance the association between the generated group dance (data) and the group embedding (context), we aim to maximize their mutual information with a Contrastive Encoder $f(\hat{x},w)$ via the contrastive learning objective as in Equation \ref{eq:info_nce}. The encoder takes both the generated group dance sequence $\hat{x}$ and a group embedding $w$ as inputs, and it outputs a score indicating the correspondence between these two. 
\begin{equation}
\label{eq:info_nce}
\mathcal{L}_{\rm nce} = - \mathbb{E} \left[ \log\frac{f(\hat{x},w)}{f(\hat{x},w) + \Sigma_{x^j \in X'}f(\hat{x}^j, w)}\right] 
\end{equation}
where $X'$ is a set of randomly constructed negative sequences. In general, this loss is similar to the cross-entropy loss for classifying the positive sample, and optimizing it leads to the maximization of the mutual information between the learned context representation and the data~\cite{oord2018contrastive_cpc}. Using the contrastive objective, we expect the Contrastive Encoder to learn to distinguish between the two quantities: \textit{consistency} (the positive sequence) and \textit{diversity} (the negative sequence). This is the key factor that enables the ability to control diversity and consistency in our framework.

Here, we will describe our strategy to construct contrastive samples to achieve our target. Recall that we use reverse distribution $p_\theta(x_{t-1} | x_t)$ of Gaussian Diffusion with the mean as the prediction of the model while the variance is fixed to a scheduler (Equation~\ref{eq:approximate_posterior}). To obtain the contrastive samples, given the true pair is $(x_0,w)$, we first leverage forward diffusion process $q(x_m|x_0)$ to obtain the noised sample $x_m$. Then, our \textit{positive sample} is $p_\theta(x_{m-1} |x_m, w)$. Subsequently, we construct the negative sample from the positive pair by randomly replacing dancers from other group dance sequences ($x^j_0 \neq x_0$) with some probabilities, feeding it through the forward process to obtain $x^j_m$, then our \textit{negative sample} is $p_\theta(x^j_{m-1} | x^j_m, w)$. 
By constructing contrastive samples this way, the positive pair $(x_0,w)$ represents a group sequence with high consistency, whereas the negative one represents a high diversity sample. This is because mixing a sample with dancers from different groups is likely to result in substantially distinctive movements between each dancer, making it a group dance sample with high degree of diversity. 
  {Note that negative sequences should also match the music because they are motions generated by the network whose inputs are manipulated to increase diversity. Particularly,  negative samples are acquired from outputs of the denoising network whose inputs are both the current music and the noised mixed group with some replaced dancers. As the network is trained to reconstruct only positive samples, its outputs will likely follow the music. Therefore, negative samples are not just random bad samples but are the valid group dance generated from the network that is trained to generate group dance conditioned on the music. This is because our main diffusion training objective (Section~{\ref{subsec:training}}) is calculated only for ground-truth dances (positive samples) that are consistent with the music. }
Our proposed strategy also allows us to learn a more powerful group representation as it directly affects the reverse process, which is beneficial to maintaining consistency in long-term synthesis.

\subsubsection{Diversity vs. Consistency}
Using the Contrastive Encoder $f(x_m,w)$, we extend the classifier guidance~\cite{dhariwal2021classifier_guidance} to control the generation process. Accordingly,   {we} incorporate $f(x_m,w)$ in the contrastive framework to replace the guiding classifier in the original formula, since it provides a score of how consistent the sample is with the group information. In particular, we shift the mean of the reverse diffusion process with the log gradient of the Contrastive Encoder with respect to the generated data as follows:
\begin{align}
    \hat{\mu}_\theta(x_m,m) = \mu_\theta(x_m,m) &+ \gamma \cdot \Sigma_{\theta}(x_m,m) \nabla_{x_m}\log f(x_m,w) 
\label{eq_gamma}
\end{align}
where $\gamma$ is the control parameter that uses the encoder to enforce consistency and connection with the group embedding. Since the Contrastive Encoder is trained to classify between high-consistency and high-diversity samples, its gradients yield meaningful guidance signals to control the trade-off process.
Intuitively, a positive value of $\gamma$ encourages more consistency between dancers while a negative value (which corresponds to shifting the distribution with a negative gradient step) boosts the diversity between each individual dancer.

\section{Experiments}
\subsection{Implementation Details}
\subsubsection{Network Parameters}
The hidden layer of all the MLPs consists of $512$ units followed by GELU activation.  The hidden dimension of all attention layers is set to $d=512$, and the attention adopts a multi-head scheme~\cite{vaswani2017attention} with $8$ attention heads. We also use a feature-wise linear modulation (FiLM)~\cite{perez2018film,tseng2022edge_diffusion} after each attention layer to strengthen the influence of the conditioning context. At the end of each attention block, we append a $2$-layer feed-forward network~\cite{vaswani2017attention} with a feed-forward size of $1024$ to enhance the expressivity of the learned features. We extract the features from the raw audio signal by leveraging the representations from the frozen Jukebox~\cite{dhariwal2020jukebox}, a pre-trained generative model for music, to enhance the model's generalization ability to several kinds of in-the-wild music. In total, the Group Diffusion Denoising Network is comprised of $L=5$ stacked Music-Motion Transformer and Group Global Attention blocks, along with $2$ transformer encoder layers to encode the music features. We implement the architecture of the Contrastive Encoder similarly to the Denoising Network but without cross attention since it does not take the music as input. The output sequence of the Contrastive Encoder is then averaged out and fed into an output layer with one unit. We also make the Contrastive Encoder aware of the current step in the diffusion chain by appending the diffusion timestep embedding to the motion sequence so that it can provide correct guidance signals in the sampling process. Overall, our model has approximately $62M$ trainable parameters. 

\subsubsection{Training}
\label{subsec:training}
To train the denoising diffusion network, we use the "simple" objective as introduced in~\cite{ho2020ddpm}. 
\begin{equation}
\label{eq:simple}
    \mathcal{L}_{\rm simple} = \mathbb{E}_{x_0\sim q(x_0|c), m\sim [1,M]} \left[\Vert x_0 - \mathcal{G}_\theta(x_m,m,c) \Vert^2_2\right]
\end{equation}
To improve the physical plausibility and prevent artifacts of the generated motion, we also utilize auxiliary geometric losses similar to~\cite{tevet2022mdm}.
\begin{equation}
    \mathcal{L}_{\rm geo} = \lambda_{\rm pos}\mathcal{L}_{\rm pos} + \lambda_{vel}\mathcal{L}_{\rm vel} + \lambda_{\rm foot}\mathcal{L}_{\rm foot}
\end{equation}
In particular, geometric losses mainly consist of \textit{(i)} a joint position loss  $\mathcal{L}_{\rm pos}$ to better constrain the global joint hierarchy via forward kinematics; \textit{(ii)} a velocity loss $\mathcal{L}_{\rm vel}$ to increase the smoothness and naturalness of the motion by penalizing the difference between the differences between the velocities of the ground-truth and predicted motions; and \textit{(iii)} a foot contact loss $\mathcal{L}_{\rm foot}$ to mitigate foot skating artifacts and improve the realism of the generated motions by ensuring the feet to stay stationary when ground contact occurs.

Our total training objective is the combination of the "simple" diffusion objective, the auxiliary geometric losses, and the contrastive loss (Equation~\ref{eq:info_nce}):
\begin{equation}
    \mathcal{L} = \mathcal{L}_{\rm simple} + \mathcal{L}_{\rm geo} + \lambda_{\rm nce}\mathcal{L}_{\rm nce}
\end{equation}
We train our model on $4$ NVIDIA V100 GPUs using Adam optimizer~\cite{kingma2014adam} with a learning rate of $1\mathrm{e}{-4}$ and a batch size of $64$ per GPU, which took about $7$ days for $500$k iterations. The models are trained with $M=1000$ diffusion noising steps and a cosine noise  schedule~\cite{nichol2021improved}. During training, group dance motions are randomly sampled with sequence length $T=150$ at $30$ Hz, which corresponds to $5$-second pieces of music. For geometric losses, the loss weights are empirically set to $\lambda_{\rm pos}=1.0$, $\lambda_{\rm smooth}=1.0$, and $\lambda_{\rm foot}=0.005$, respectively. For the contrastive loss $\mathcal{L}_{\rm nce}$, its weight is $\lambda_{\rm nce} = 0.001$, the probability of replacing dancers for negative sequences is $0.5$, and the number of negative samples empirically is selected to $10$.

\subsubsection{Testing} 
At test time, we use the DDIM sampling technique~\cite{song2021ddim} with $50$ steps to accelerate the sampling speed of the reverse diffusion process. Accordingly, our model can achieve real-time generation at $30$ Hz on a single RTX 2080Ti GPU (excluding the music features extracting step), thanks to the parallelization of the Transformer architecture.

To enable long-term generation, we adopt a strategy that is similar to the one described in~\cite{tseng2022edge_diffusion}. Specifically, we divide the input music sequence into multiple overlapping chunks, with each chunk having a maximum window size of $5$ seconds and overlapped by half with the adjacent chunk. The group dance motions are then generated for each chunk along with the corresponding audio. Subsequently, we merge the outputs by blending the overlapped region between two consecutive chunks using spherical linear interpolation, with the interpolation weight gradually decaying from the current chunk to the next chunk. However, for group choreography synthesis, our model generates dance motions for each dancer in random order. Therefore, we need to establish correspondences between dancers across the chunks (i.e., identifying which one of the $N$ dancers in the next chunk corresponds to a dancer in the current chunk). To accomplish this, we organize all dancers in the current chunk into one set and the dancers in the next chunk into another set, forming a bipartite graph between the two chunks. We can then utilize the Hungarian algorithm~\cite{kuhn1955hungarian} to find the optimal matching, where the Euclidean distance between the two pose sequences serves as the matching weights.   {Our blending technique is applied at each step of the diffusion sampling process, starting from pure noise, thus it allows the model to gradually denoise the chunks to make them compatible for blending.}

\subsection{Experimental Settings}
\label{sec:metric}

\subsubsection{Dataset}
We use AIOZ-GDance dataset~\cite{le2023music} in our experiments. AIOZ-GDance is a large-scale group dance dataset including paired music and 3D group motions captured from in-the-wild videos using a semi-automatic method, covering 7 dance styles and 16 music genres. We follow the training and testing split as in~\cite{le2023music} in our experiments.

\subsubsection{Evaluation Protocol}
We use the following metrics to evaluate the quality of single dancing motion:~ Frechet Inception Distance (FID)~\cite{heusel2017ganfid,li2021AIST++}, Motion-Music Consistency (MMC)~\cite{li2021AIST++}, Generation Diversity (GenDiv)~\cite{Dance_Revolution, lee2019_dancing2music,li2021AIST++}   {,  Physical Foot Contact score (PFC) {\cite{tseng2022edge_diffusion}}}.
Concretely, FID score measures the realism of individual dance movements against the ground-truth dance. The MMC evaluates the matching similarity between the motion and the music beats, i.e., how well generated dances follow the beat of the music. The generation diversity (GenDiv) is evaluated as the average pairwise distance of the kinetic features of the motions~\cite{onuma2008fmdistance}.   {The PFC evaluates the physical plausibility of the foot movements by calculating the agreement between the acceleration of the character's center of mass and the foot's velocity.}

To evaluate the group dance quality, we follow three metrics introduced in ~\cite{le2023music}: Group Motion Realism (GMR), Group Motion Correlation (GMC), and Trajectory Intersection Frequency (TIF).  
In general, the GMR measures the realism between generated and ground-truth group motions by calculating Frechet Inception Distance on the extract group motion features. The GMC evaluates the synchrony between dancers within the generated group by calculating their cross-correlation. The TIF measures how often the generated dancers collide with each other in their dance movements.

\subsubsection{Baselines}   {We compare our GCD method with several recent approaches on music-driven dance generation: FACT {\cite{li2021AIST++}}, Transflower {\cite{perez2021_transflower}}, and EDGE {\cite{tseng2022edge_diffusion}}, all of which are adapted for benchmarking in the context of group dance generation {\cite{le2023music}} since the original methods were specifically designed for single-dance. We also evaluate against GDanceR {\cite{le2023music}}, a recent model specifically designed for generating group choreography.}

\subsection{Experimental Results}
\label{sec:quantitative}

\subsubsection{Quality Comparison}

Table~\ref{tab:consistencyVsDiversity} shows a comparison among the baselines FACT~\cite{li2021AIST++}, Transflower~\cite{perez2021_transflower}, EDGE~\cite{tseng2022edge_diffusion}, GDanceR~\cite{le2023music}, and our proposed GCD. The results clearly demonstrate that our default model setting with ``neutral" mode outperforms the baselines significantly across all evaluations.   {We also observe that EDGE, a recent diffusion dance generation model, can yield very competitive performance on single-dance metrics (FID, MMC, GenDiv, and PFC). This suggests the advantages of diffusion approaches in motion generation tasks. However, it is still inferior to our model under several group dance metrics, showing the limitations of single dance methods in the context of group dance creation.} Experimental results highlight the effectiveness of our approach in generating high-quality group dance motions.

\begin{table}[!t]
\centering
\caption{Performance comparison. High Consistency: parameter $\gamma=1$; High Diversity: parameter $\gamma=-1$; Neutral: parameter $\gamma=0$}
\resizebox{\linewidth}{!}{
\setlength{\tabcolsep}{0.2 em} % for the horizontal padding
{\renewcommand{\arraystretch}{1.2}% for the vertical padding
\begin{tabular}{lc|cccc|ccc}
\hline
\multicolumn{2}{c|}{\textbf{Method}} & \multicolumn{1}{c|}{FID$\downarrow$} & \multicolumn{1}{c|}{MMC$\uparrow$} & \multicolumn{1}{c|}{GenDiv$\uparrow$} &   {PFC}$\downarrow$ & \multicolumn{1}{c|}{GMR$\downarrow$} & \multicolumn{1}{c|}{GMC$\uparrow$} & TIF$\downarrow$ \\   \hline
\multicolumn{2}{l|}{FACT~\cite{li2021AIST++}} & \multicolumn{1}{c|}{56.20} & \multicolumn{1}{c|}{0.222} & \multicolumn{1}{c|}{8.64} & 3.52 & \multicolumn{1}{c|}{101.52} & \multicolumn{1}{c|}{62.68} & 0.321 \\ 
\multicolumn{2}{l|}{  {Transflower {\cite{perez2021_transflower}}}} & \multicolumn{1}{c|}{37.73} & \multicolumn{1}{c|}{0.217} & \multicolumn{1}{c|}{8.74} & 3.07 & \multicolumn{1}{c|}{81.17} & \multicolumn{1}{c|}{60.78} & 0.332 \\ 
\multicolumn{2}{l|}{  {EDGE {\cite{tseng2022edge_diffusion}}}} & \multicolumn{1}{c|}{31.40} & \multicolumn{1}{c|}{0.264} & \multicolumn{1}{c|}{9.57} & 2.63 & \multicolumn{1}{c|}{63.35} & \multicolumn{1}{c|}{61.72} & 0.356 \\ 
\hline
\multicolumn{2}{l|}{GDANCER~\cite{le2023music}} & \multicolumn{1}{c|}{43.90} & \multicolumn{1}{c|}{0.250} & \multicolumn{1}{c|}{9.23} & 3.05 & \multicolumn{1}{c|}{51.27} & \multicolumn{1}{c|}{79.01} & 0.217  \\
\hline\hline
\multicolumn{1}{l|}{\multirow{3}{*}{\textbf{\begin{tabular}[c]{@{}c@{}}GCD\\(Ours)\end{tabular}}}} & High Consistency & \multicolumn{1}{c|}{31.48} & \multicolumn{1}{c|}{\textbf{0.272}} & \multicolumn{1}{c|}{8.78} & 2.55 & \multicolumn{1}{c|}{39.22} & \multicolumn{1}{c|}{\textbf{82.01}} & \textbf{0.115}\\ 
\multicolumn{1}{l|}{} & Neutral & \multicolumn{1}{c|}{{\textbf{31.16}}} & \multicolumn{1}{c|}{0.261} & \multicolumn{1}{c|}{10.87} & \textbf{2.53} &\multicolumn{1}{c|}{\textbf{31.47}} & \multicolumn{1}{c|}{80.97} & 0.167\\ 
\multicolumn{1}{l|}{} & High Diversity & \multicolumn{1}{c|}{33.37} & \multicolumn{1}{c|}{0.255} & \multicolumn{1}{c|}{\textbf{11.34}} & 2.58 & \multicolumn{1}{c|}{35.63} & \multicolumn{1}{c|}{78.19} & 0.209\\  \hline
\end{tabular}
}}
\vspace{2ex}

    \label{tab:consistencyVsDiversity}
\end{table}

\begin{figure*}[ht] 
   \centering
  \huge
\resizebox{\linewidth}{!}{
\setlength{\tabcolsep}{2pt}
\begin{tabular}{cccccccc}

\rotatebox[origin=l]{90}{\hspace{0.3cm} \textbf{FACT}} &
\shortstack{\includegraphics[width=0.33\linewidth]{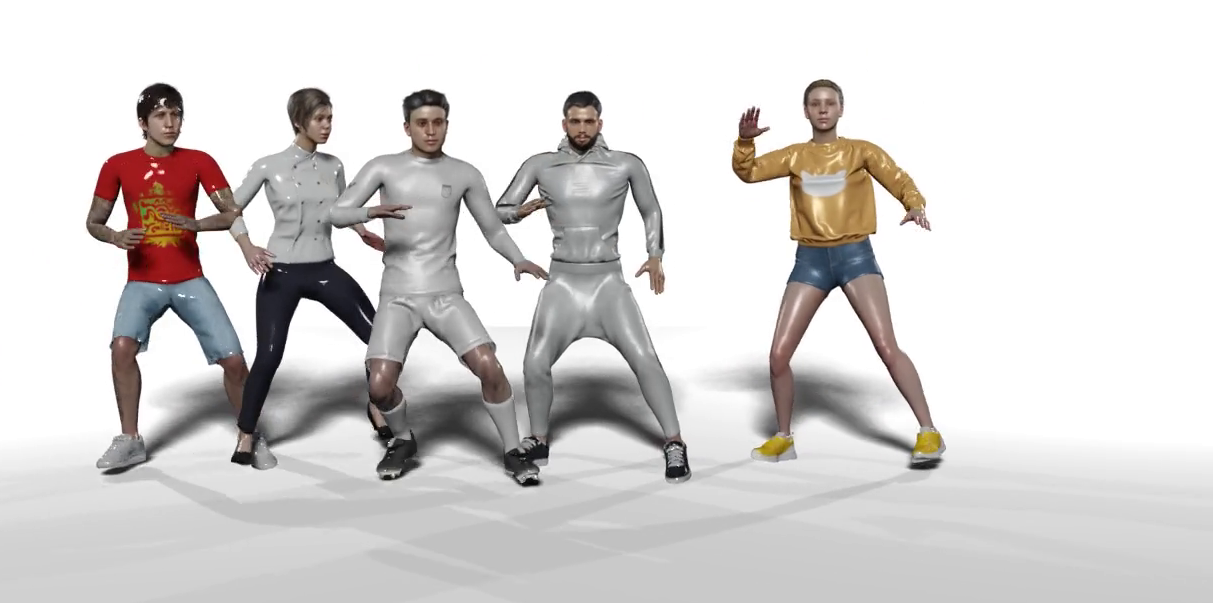}}&
\shortstack{\includegraphics[width=0.33\linewidth]{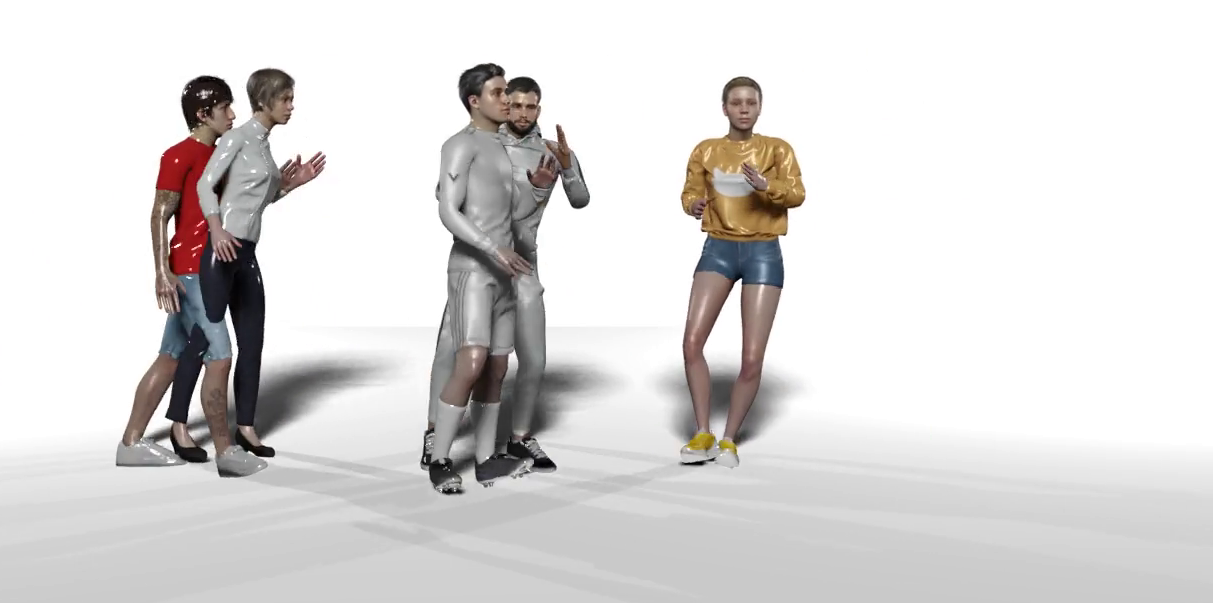}}&
\shortstack{\includegraphics[width=0.33\linewidth]{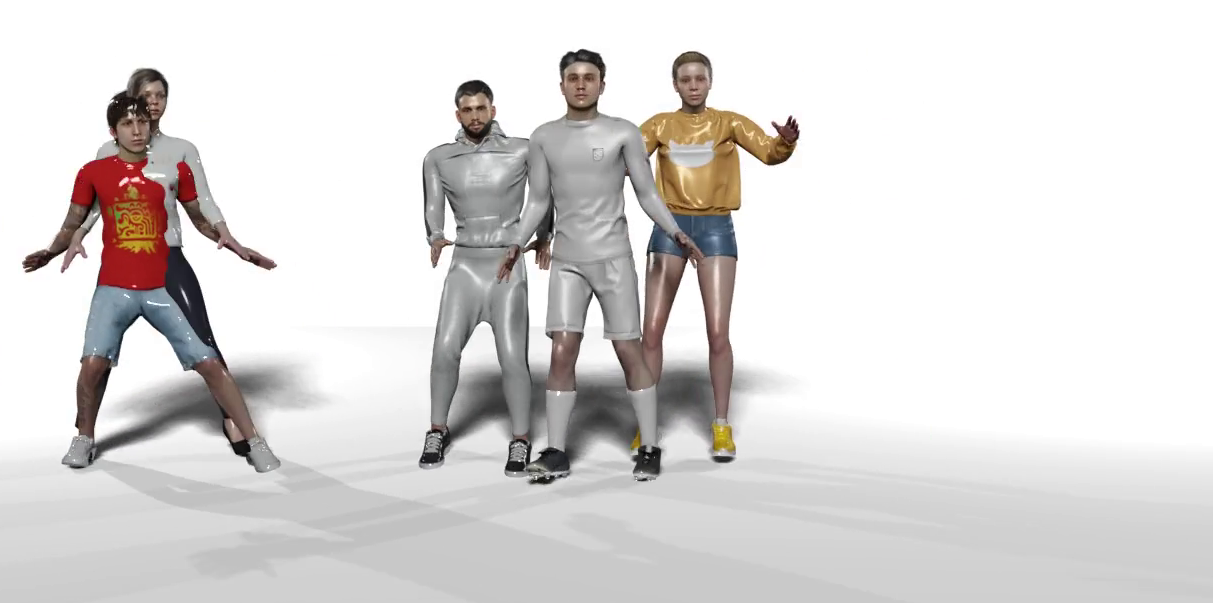}}&
\shortstack{\includegraphics[width=0.33\linewidth]{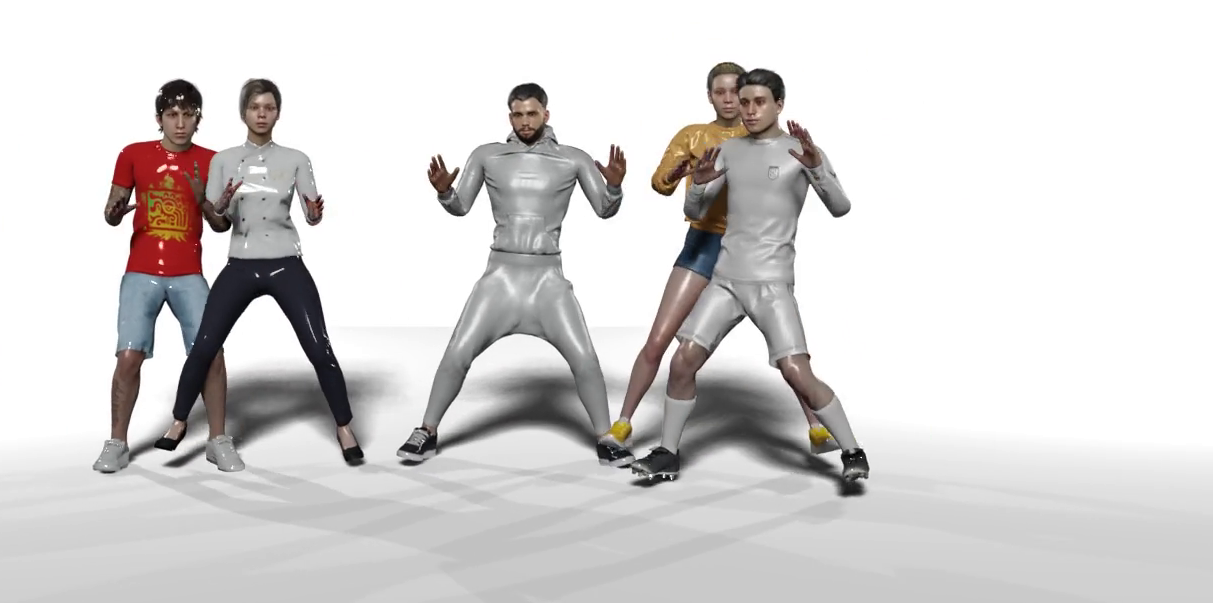}}&
\shortstack{\includegraphics[width=0.33\linewidth]{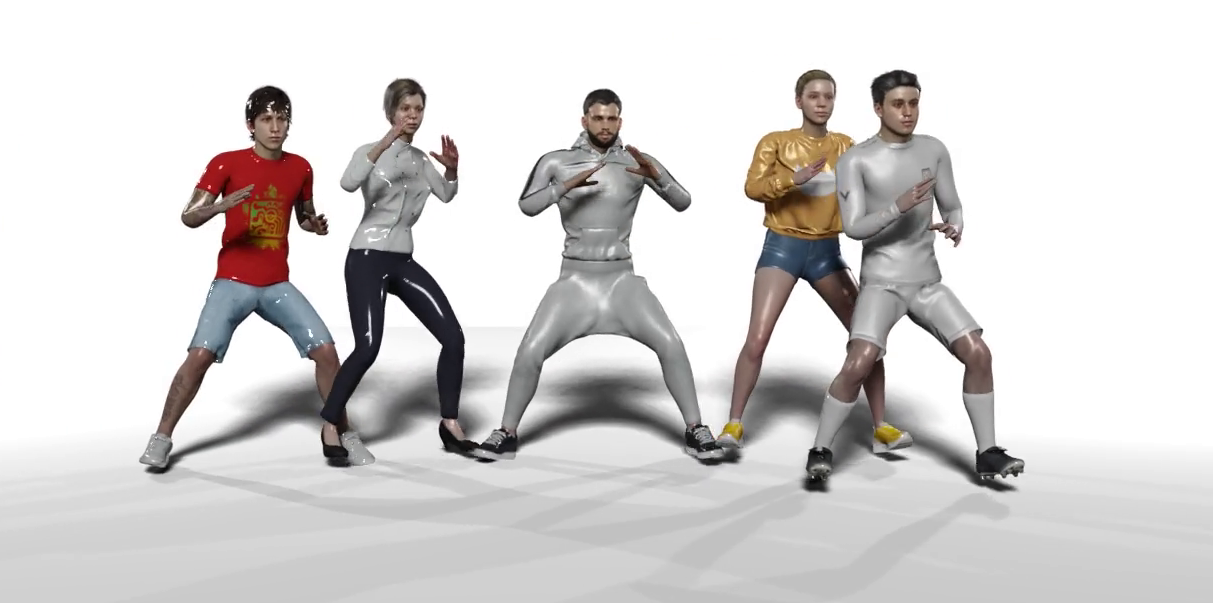}}\\[30pt]
\rotatebox[origin=l]{90}{\hspace{0cm} \textbf{GDanceR}} &
\shortstack{\includegraphics[width=0.33\linewidth]{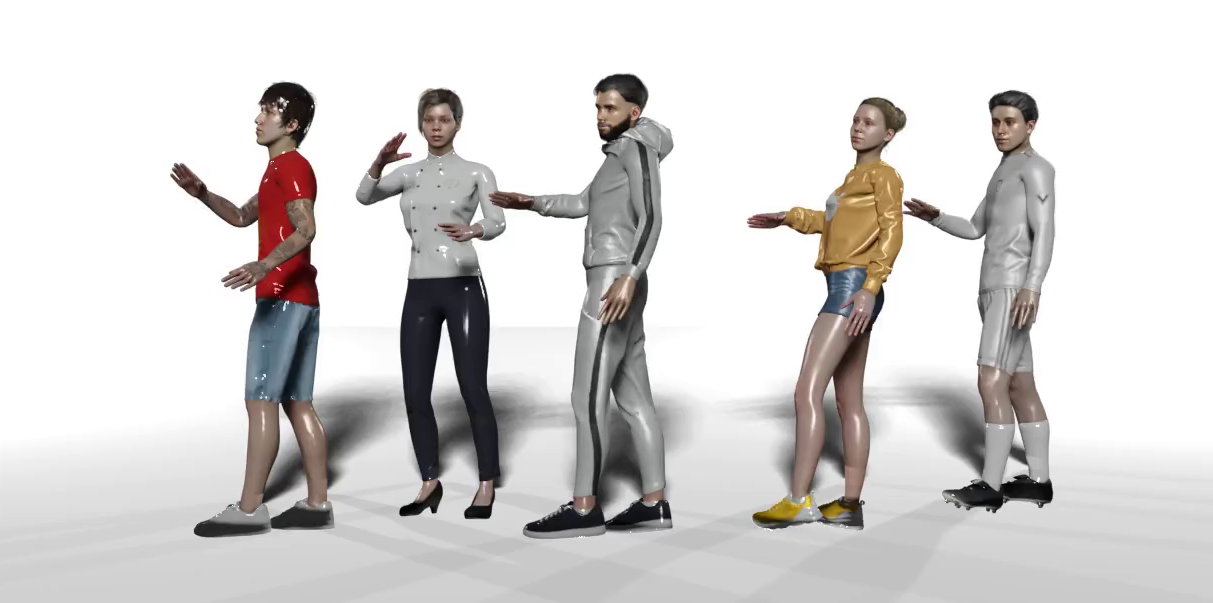}}&
\shortstack{\includegraphics[width=0.33\linewidth]{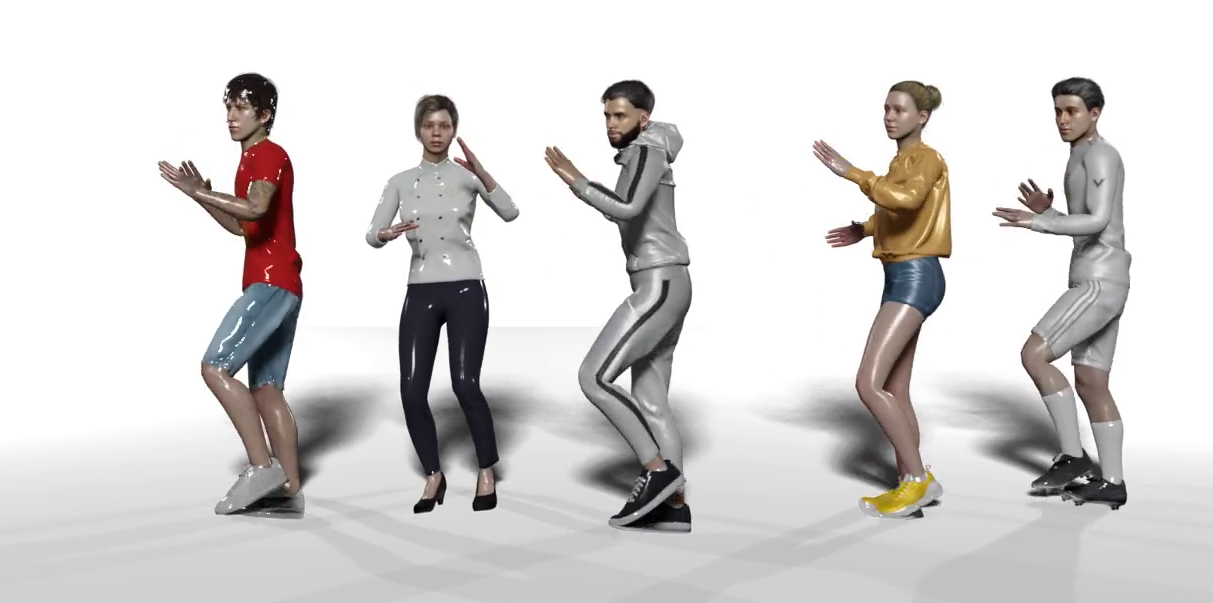}}&
\shortstack{\includegraphics[width=0.33\linewidth]{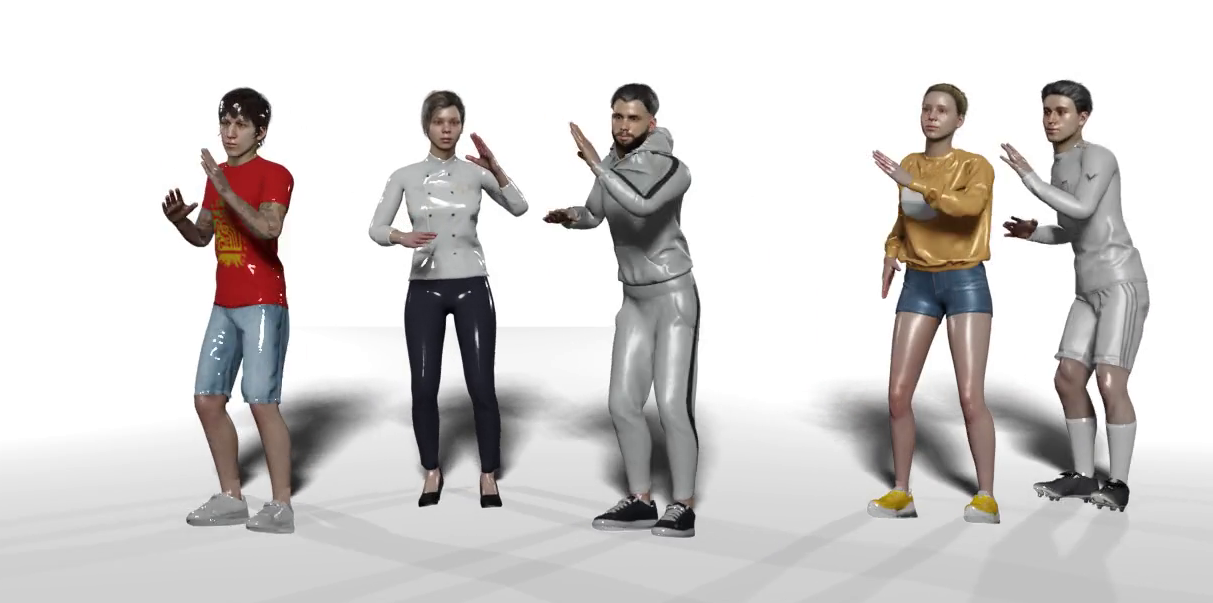}}&
\shortstack{\includegraphics[width=0.33\linewidth]{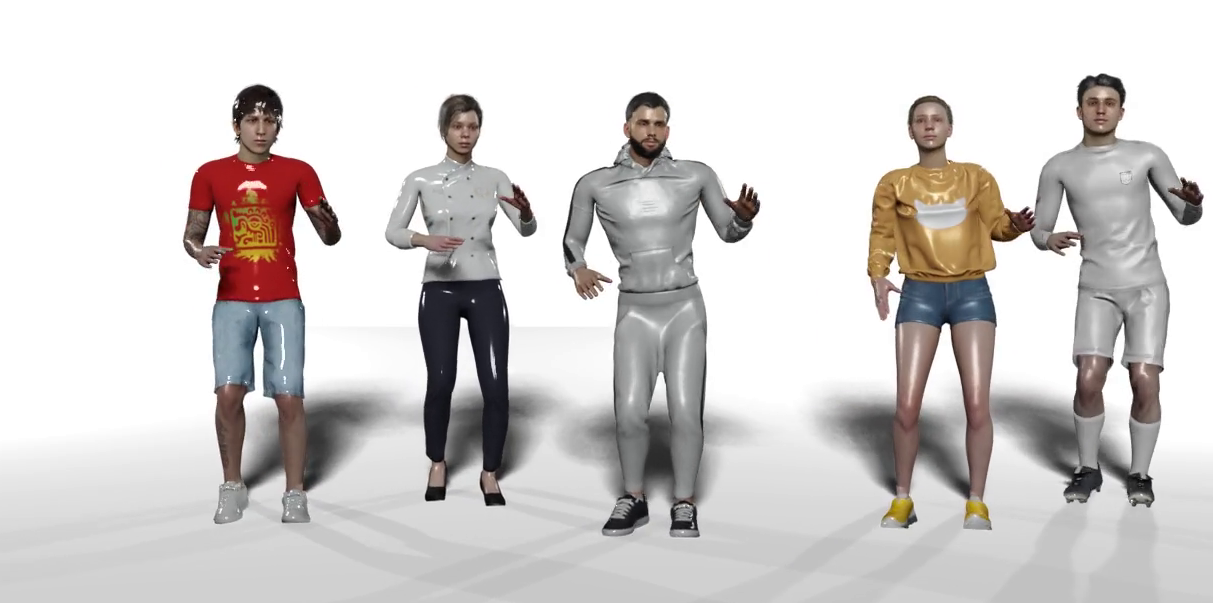}}&
\shortstack{\includegraphics[width=0.33\linewidth]{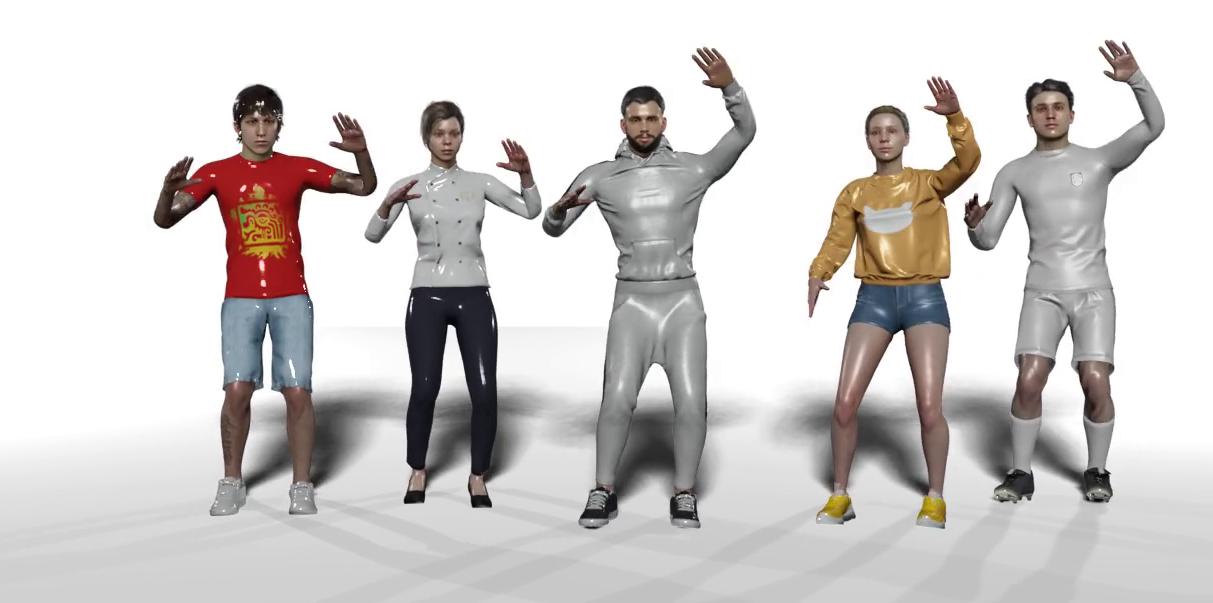}}\\[30pt]
\rotatebox[origin=l]{90}{\hspace{0.5cm} \textbf{Ours}} &
\shortstack{\includegraphics[width=0.33\linewidth]{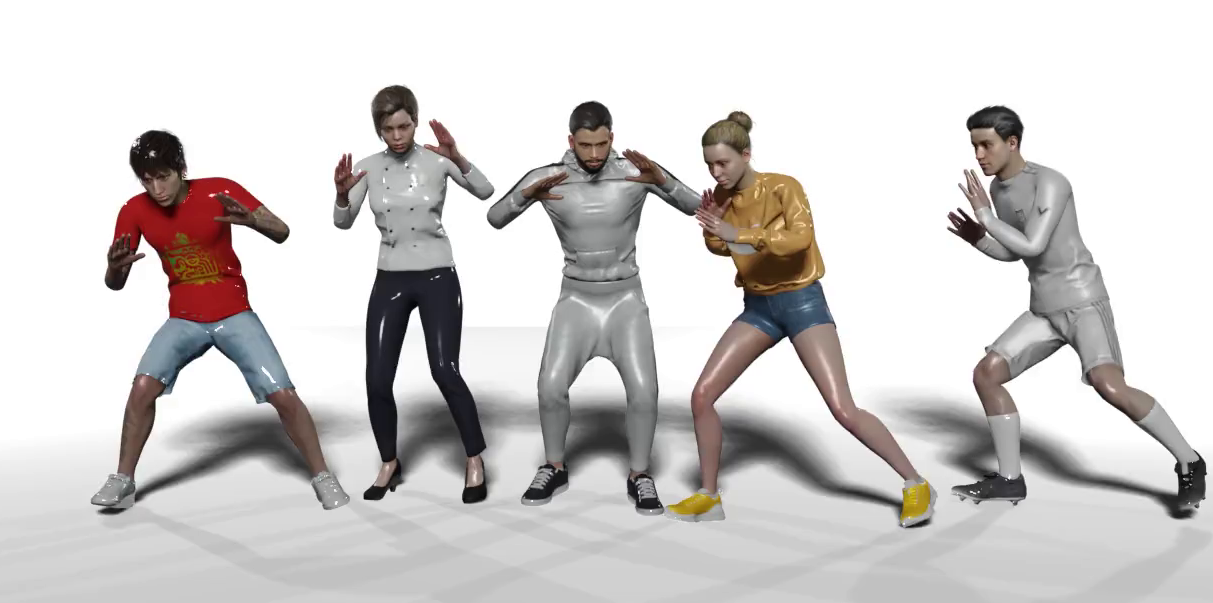}}&
\shortstack{\includegraphics[width=0.33\linewidth]{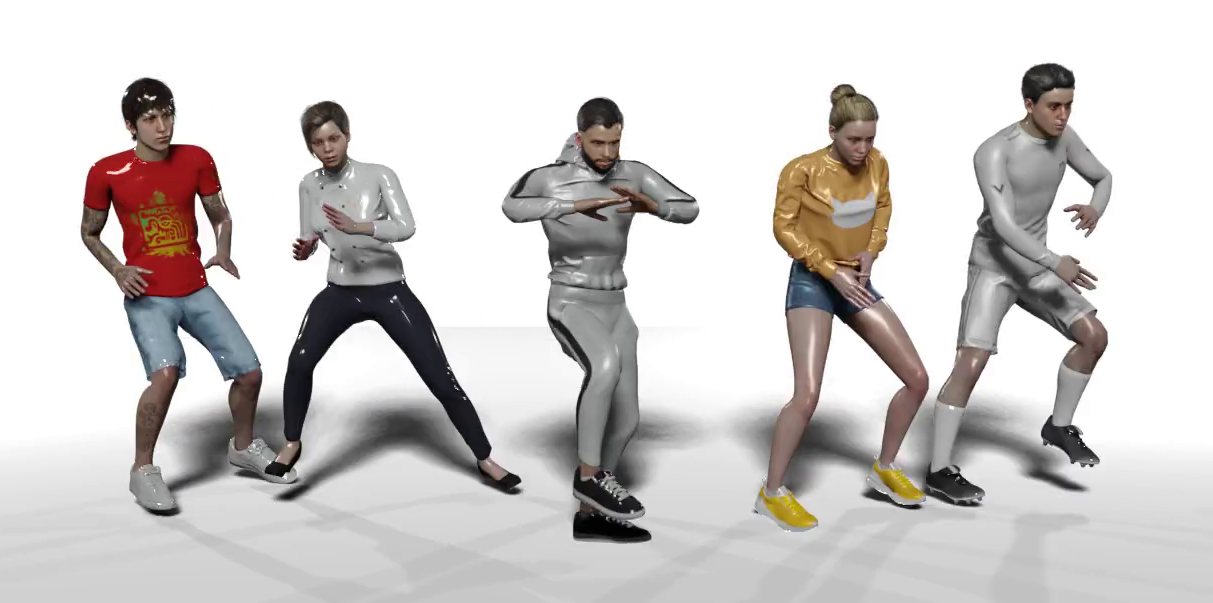}}&
\shortstack{\includegraphics[width=0.33\linewidth]{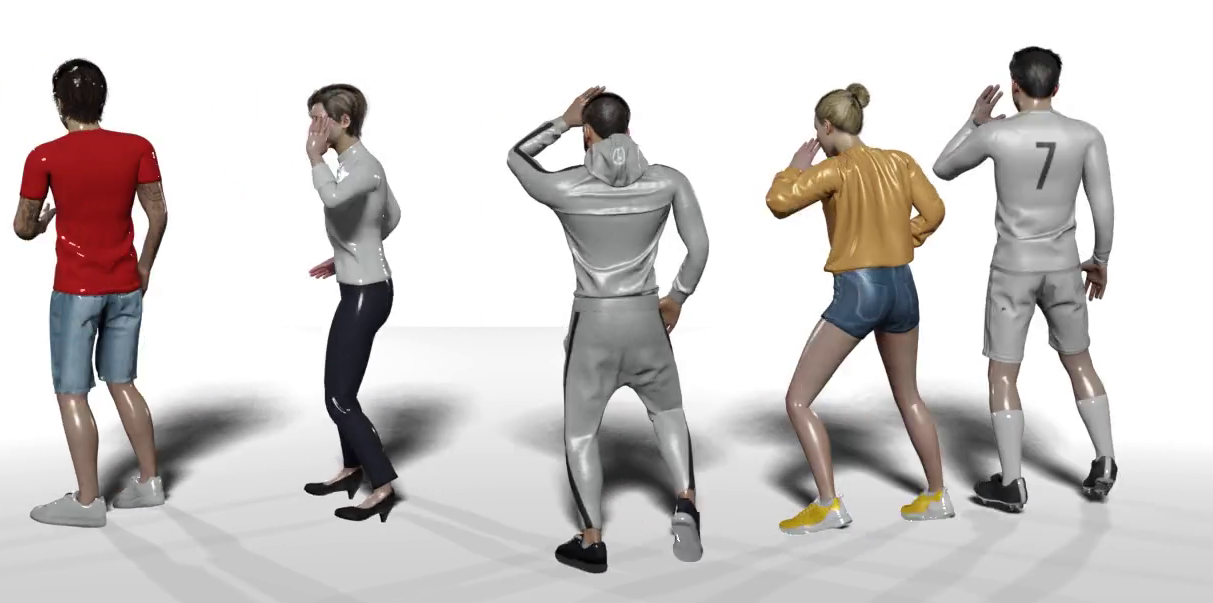}}&
\shortstack{\includegraphics[width=0.33\linewidth]{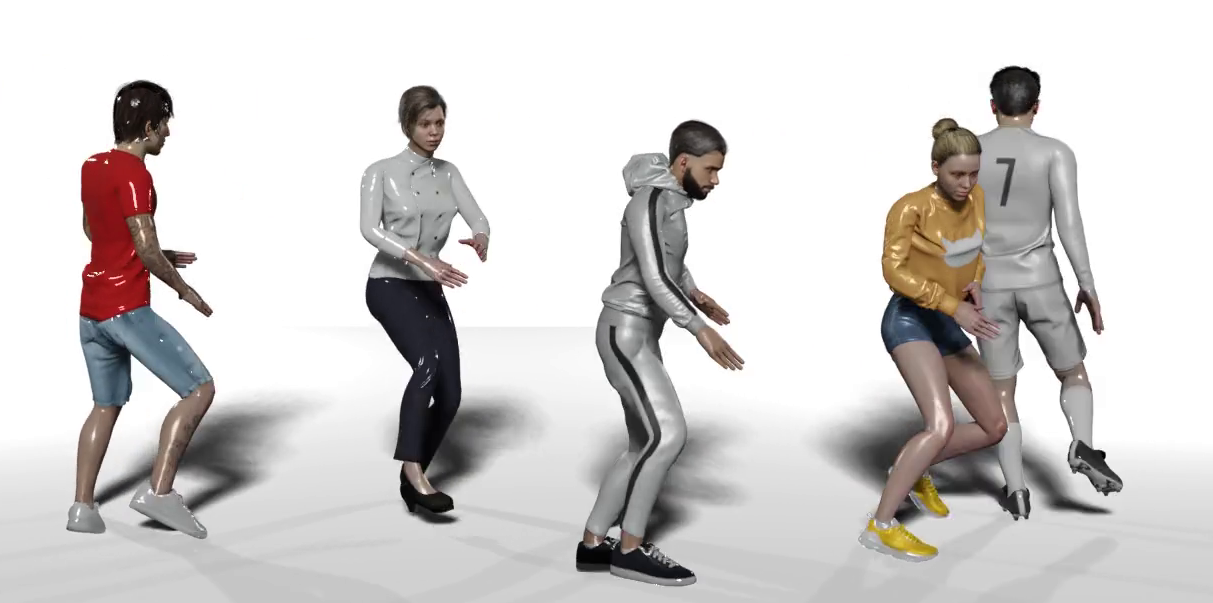}}&
\shortstack{\includegraphics[width=0.33\linewidth]{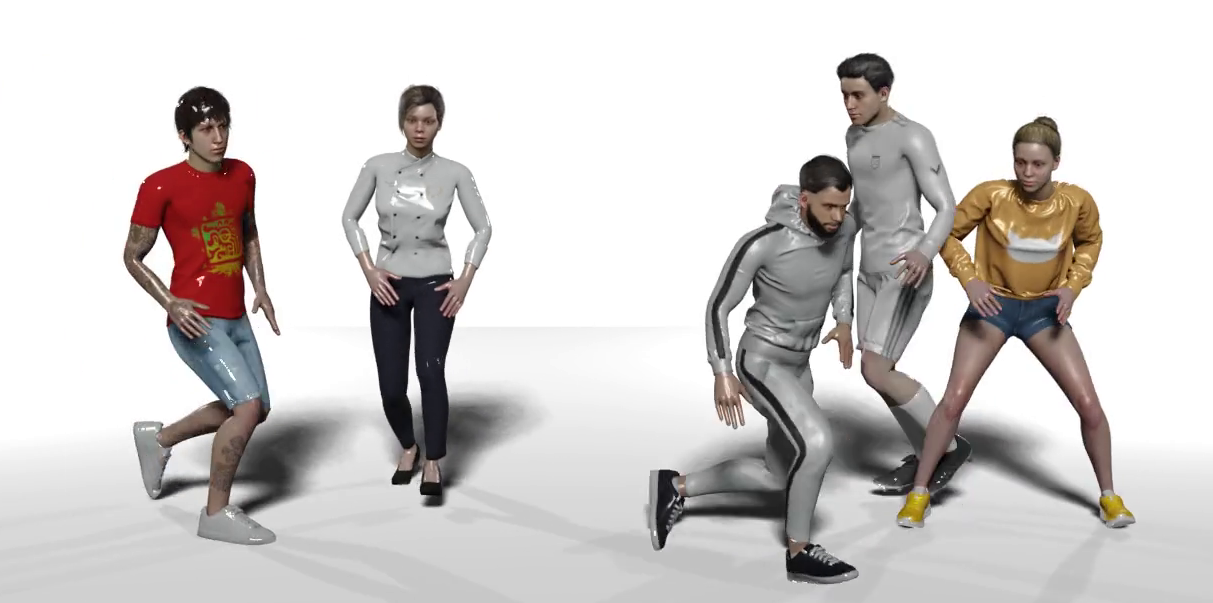}}\\[1pt]

\end{tabular}
}
    \caption{Comparison between different dance generation methods when generating dancing in groups.}
    \label{fig:OverallCompare}
\end{figure*}

To complement the quantitative analysis, we present qualitative examples from FACT, GDanceR, and our GCD method in Figure~\ref{fig:OverallCompare}. Notably, FACT struggles to deal with the intersection problem, which is reasonable given that it was not originally designed for group dance generation. As a result, the generated motions from FACT lack coordination and synchronization in most cases. 
While GDanceR shows improvements in terms of motion quality compared to FACT, the generated motions appear floating, unnatural, and sometimes unsynchronized in many cases. These drawbacks indicate that GDanceR's effort on generating group choreography would still require more refinement to produce consistent and cohesive movements among the dancers. 
In contrast, our method excels in both controlling consistency and promoting diversity among the generated group dance motions. The outputs from our method demonstrate well-coordinated and realistic movements, implying that it can resolve the challenges of maintaining group coherence while delivering visually appealing results more effectively. 

Overall, the conducted quantitative analysis and visual comparisons reaffirm the superior performance of our proposed GCD to generate high-quality, synchronized, and visually pleasing group dance motions.

\subsubsection{Diversity and Consistency Analysis}
Table~\ref{tab:consistencyVsDiversity} also presents an in-depth analysis of our method's performance across seven evaluation metrics by adjusting the parameter $\gamma$ in Equation~\ref{eq_gamma} to control the consistency and diversity of the generated group choreographies. 
The findings reveal that our GCD with high consistency setting ($\gamma = 1$), performs better than other settings in terms of MMC, GMC, and TIR metrics, whereas the high diversity setting ($\gamma=-1$) achieves better results in the GenDiv metric. Meanwhile, the default model shows the best performance in both realism metrics (FID and GMR).   {It can also be seen that the model is relatively robust to the physical plausibility score (PFC) as there are no noticeable differences among the metric in the three settings. This implies that our model is able to create group animation with different consistency or diversity levels without compromising the plausibility of the movements too much.} More interestingly, we found that there are indeed positive correlations between the two measures MMC, GMR, and the trade-off parameter. It is clear that these metrics are better when the consistency level increases. This is reasonable as we expect higher correspondence between the motion and the music (MMC) or higher correlation of the group motions (GMR) when the consistency level grows, which also agrees with the definition of these metrics. 

\begin{figure}[t]
   \centering
\resizebox{\linewidth}{!}{
\setlength{\tabcolsep}{2pt}
\begin{tabular}{ccccc}

\shortstack{\rotatebox[origin=l]{90}{\hspace{-0.4cm} High Consistency
}}&
\shortstack{\includegraphics[width=0.33\linewidth]{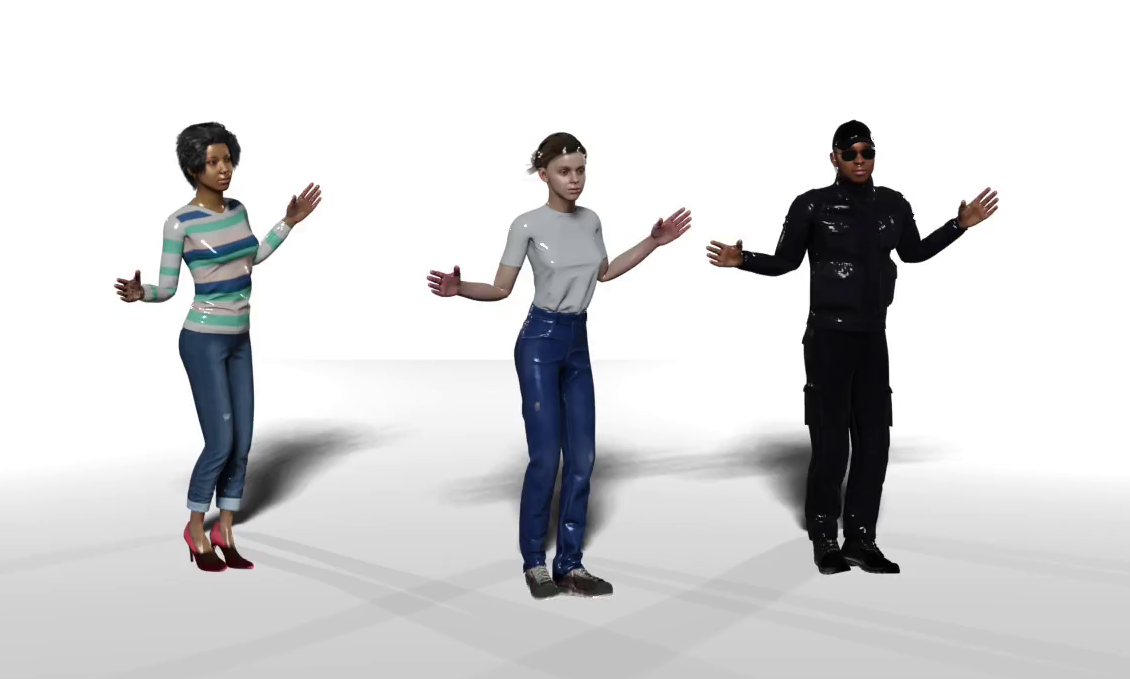}}&
\shortstack{\includegraphics[width=0.33\linewidth]{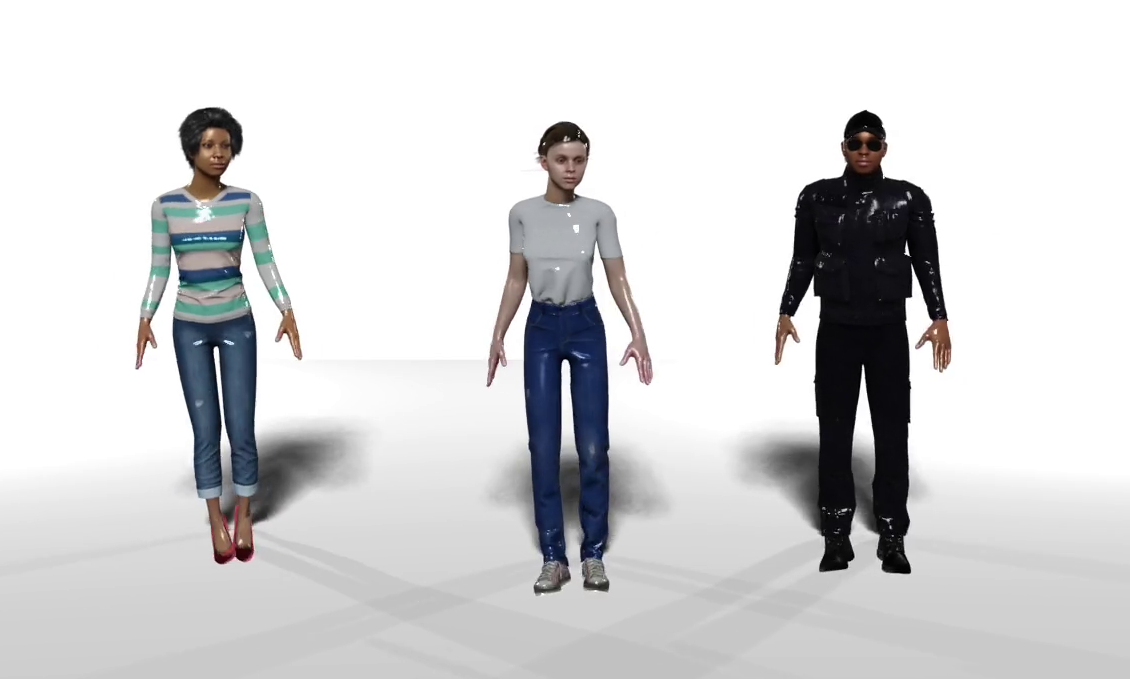}}&
\shortstack{\includegraphics[width=0.33\linewidth]{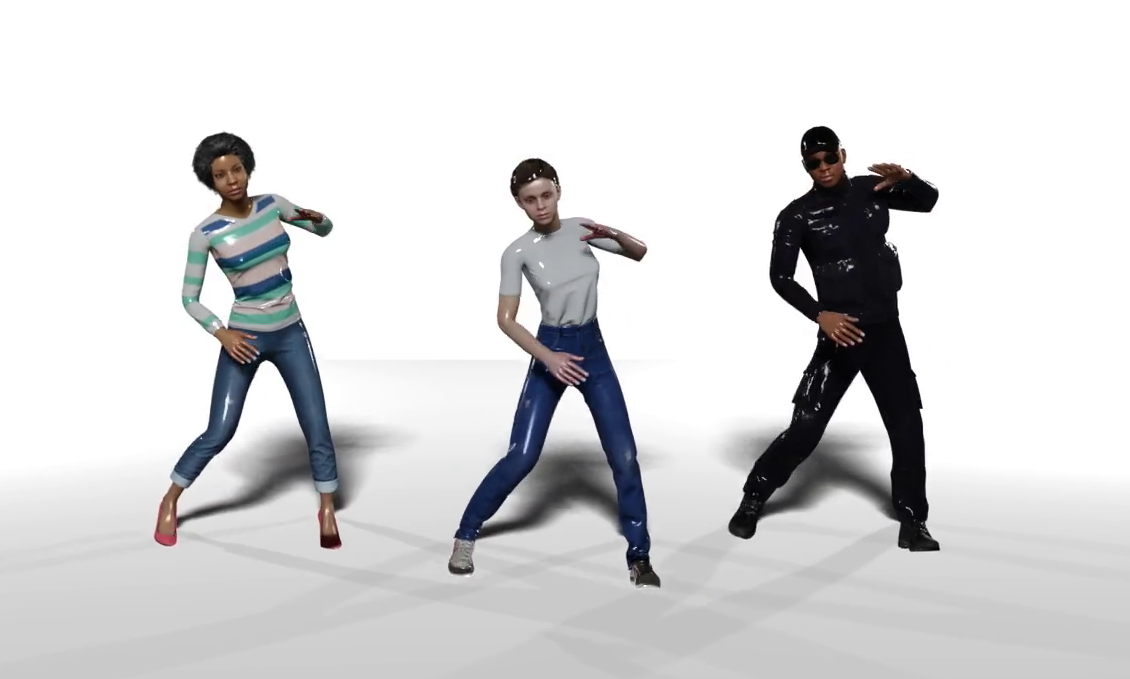}}\\[15pt]
\shortstack{\rotatebox[origin=l]{90}{\hspace{0.2cm} Neutral
}}&
\shortstack{\includegraphics[width=0.33\linewidth]{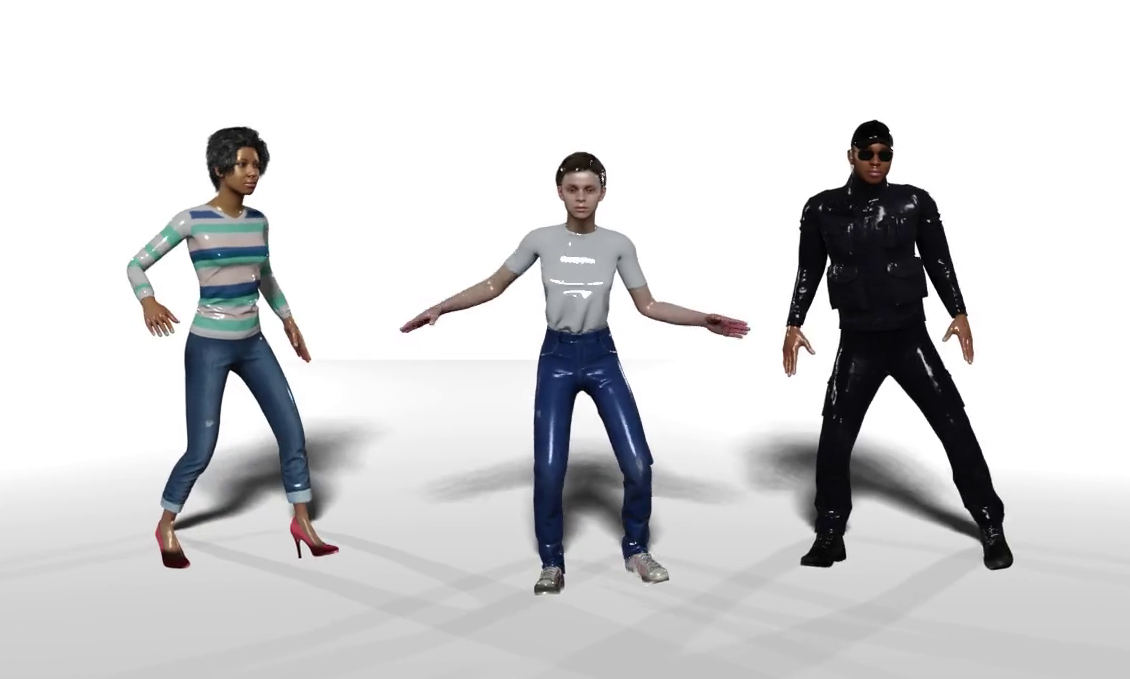}}&
\shortstack{\includegraphics[width=0.33\linewidth]{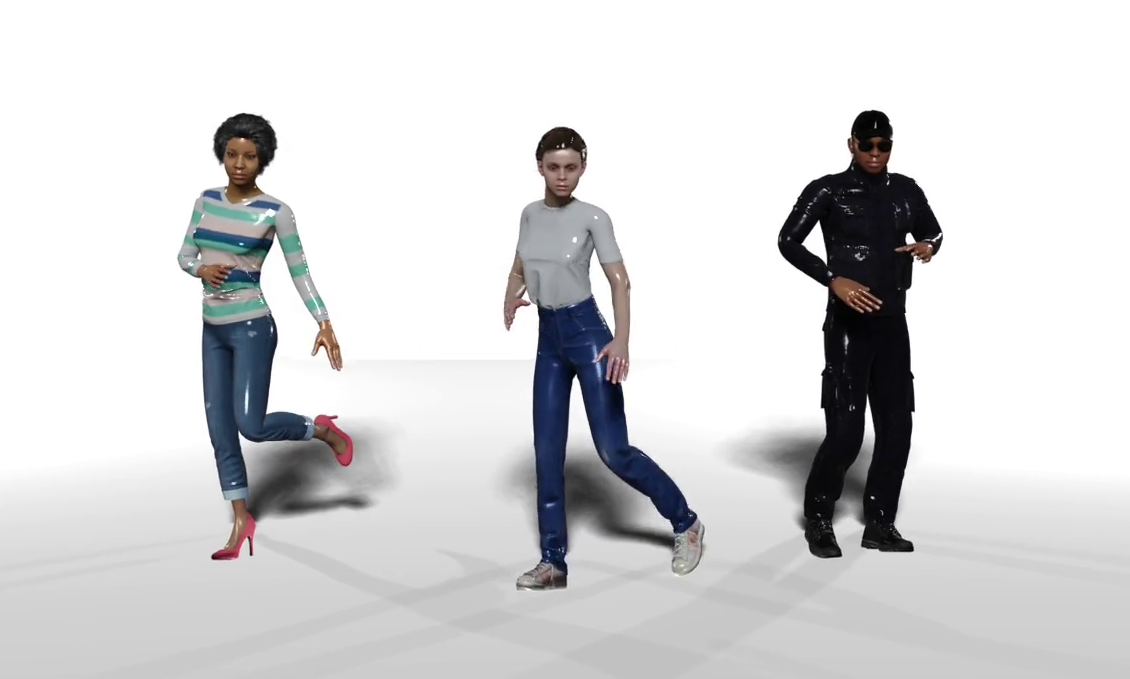}}&
\shortstack{\includegraphics[width=0.33\linewidth]{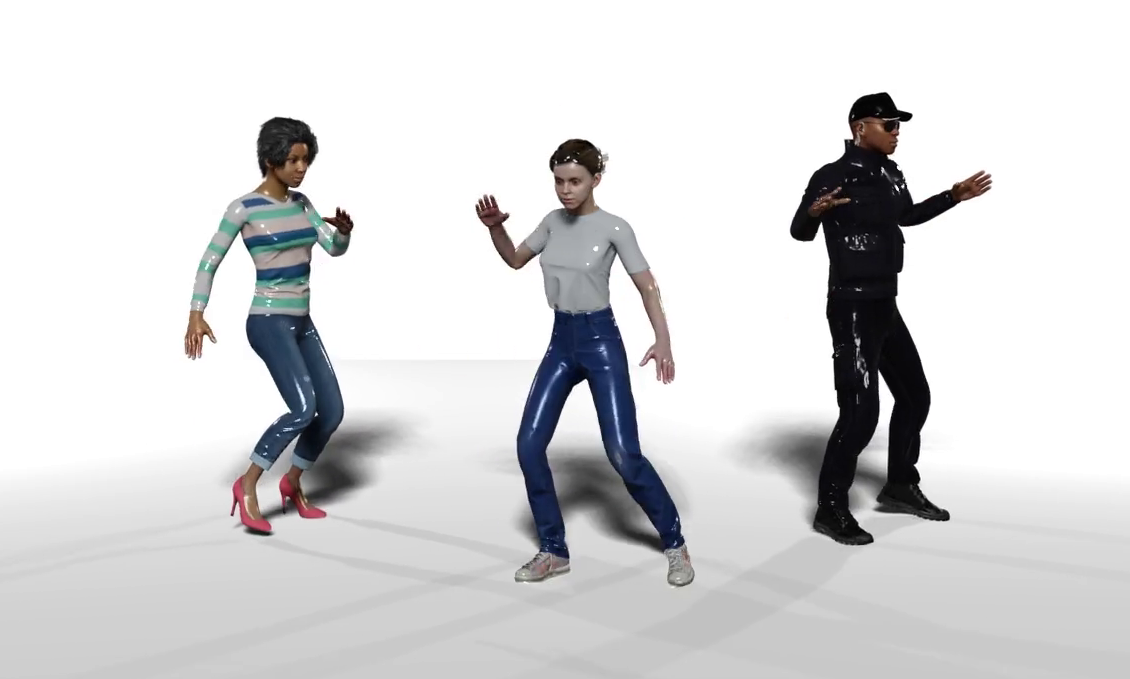}}\\[15pt]
\shortstack{\rotatebox[origin=l]{90}{\hspace{-0.1cm} High Diversity
}}&
\shortstack{\includegraphics[width=0.33\linewidth]{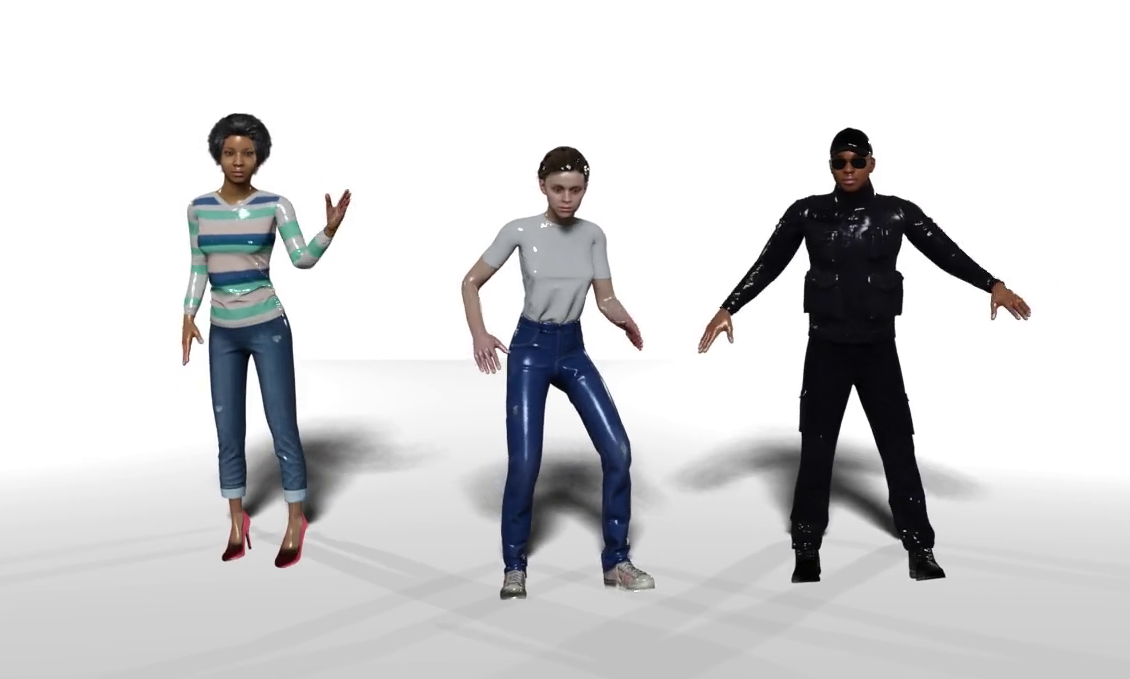}}&
\shortstack{\includegraphics[width=0.33\linewidth]{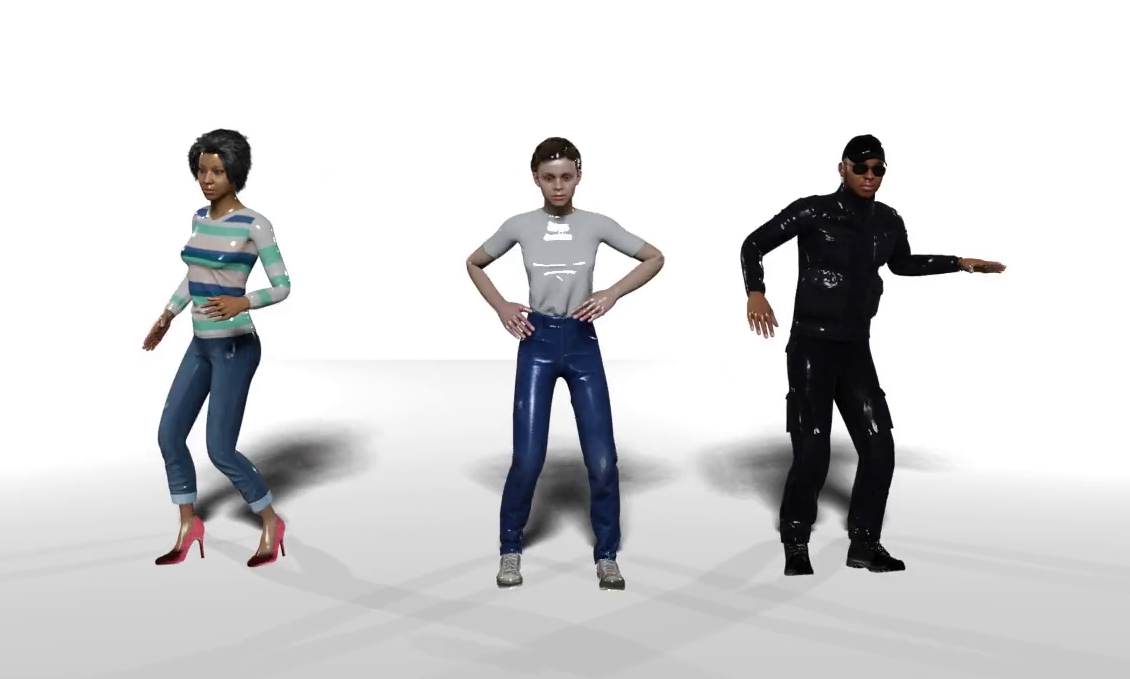}}&
\shortstack{\includegraphics[width=0.33\linewidth]{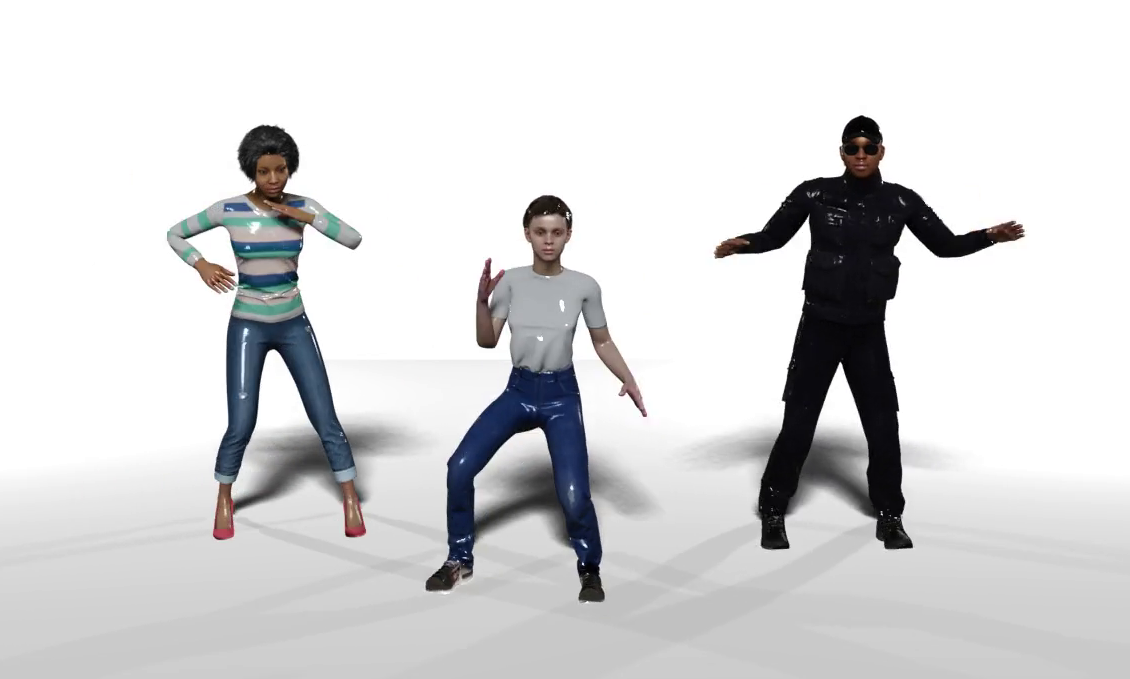}}\\[1pt]
\end{tabular}
}
    \caption{Consistency and diversity trade-off. High Consistency: parameter $\gamma=1$; High Diversity: parameter $\gamma=-1$; Neutral: parameter $\gamma=0$. }
    \label{fig:consistencyVsDiversity}
\end{figure}

\begin{figure}[t]
   \centering
\resizebox{\linewidth}{!}{
\setlength{\tabcolsep}{2pt}
\begin{tabular}{cc}
\shortstack{\rotatebox[origin=l]{90}{\hspace{0.0cm} High Consistency
}}&
\shortstack{\includegraphics[width=\linewidth]{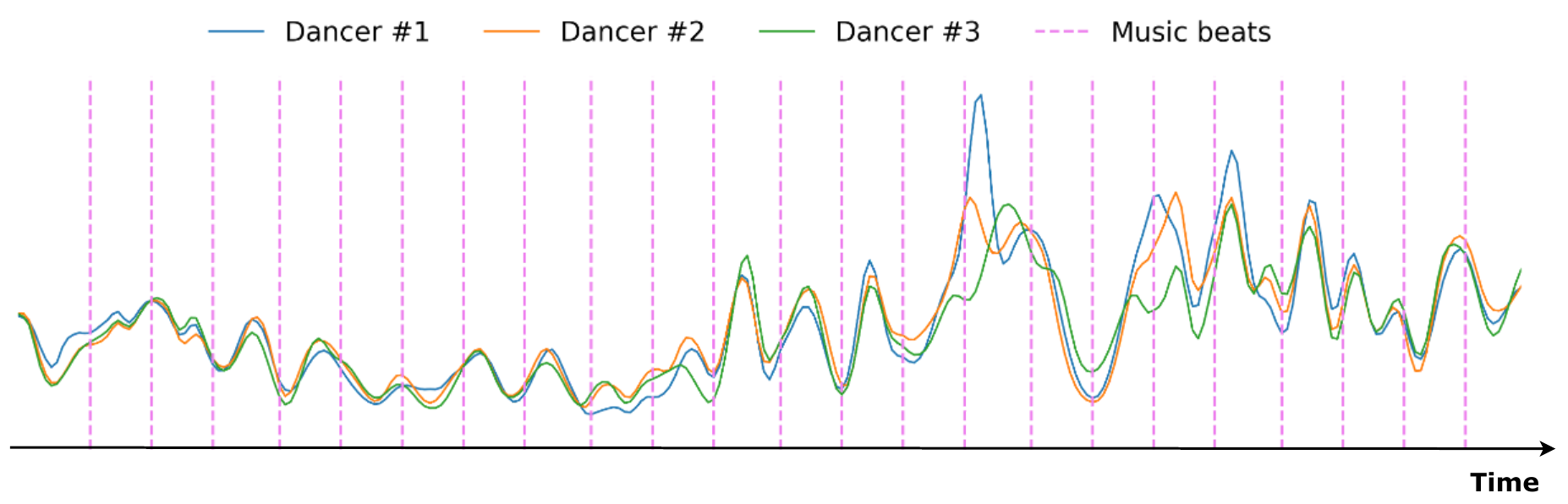}}\\[15pt]
\shortstack{\rotatebox[origin=l]{90}{\hspace{0.0cm} High Diversity
}}&
\shortstack{\includegraphics[width=\linewidth]{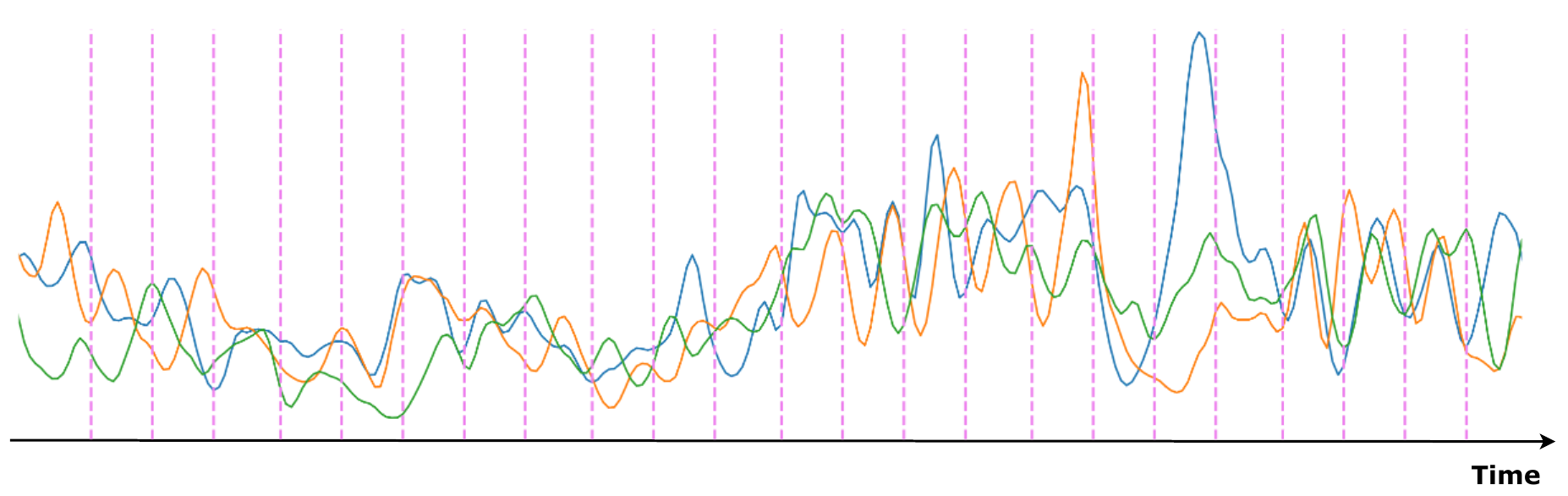}}\\[1pt]
\end{tabular}
}
    \caption{  {Correlation between the motion and music beats. The solid curve represents the kinetic velocity of each dancer over time and the vertical dashed line depicts the music beats of the sequence. The motion beats can be detected as the local extrema from the kinetic velocity curve.}}
    \label{fig:beat_align}
\end{figure}

Consistency setups in GCD lead to more similar movements between dancers. As a result, this similarity contributes to high scores in  MMC, GMR, and TIR metrics.  In contrast, the diversity setups can synthesize more complex motions with greater variation between dancers as measured by the GenDiv metric, but this also makes it more challenging to reach high values of FID, MMC, and TIR, compared with other setups. 
  {In addition, Figure~{\ref{fig:beat_align}} shows an example of correlations between motion beats and music beats under high consistency and diversity settings. The music beats are extracted using the beat tracking algorithm from the Librosa library {\cite{mcfee2015librosa}}. Notably, the velocity curves in high consistency setting display relatively similar shapes, whereas in the high diversity scenario, the curves are clearly distinguished among dancers. Despite the greater variations in high diversity setting, we can observe that the generated motions are matched with the music as the music beats are mostly located near the extrema of the motion curves in both settings.}
The experiment indicates that our model can faithfully capture different aspects of group choreography with different settings, including diversity and synchrony of the motions. This demonstrates the potential of our method towards various dance applications such as dance training or composing. Furthermore, our method can also produce distinctively different animation sequences under the same setting while adhering to the input music.  
It is also important to note that all three setups of GCD significantly outperform other baseline models. This verifies the effectiveness of our proposed approach and shows that it can create high-fidelity group dance animations in any setting.  For a more detailed visualization of the results, please refer to Figure~\ref{fig:consistencyVsDiversity} and our accompanying supplementary video.

\subsubsection{Number of Dancers Analysis}
Table~\ref{tab:n_dancers} provides insights into results obtained when generating arbitrarily different numbers of dancers using our proposed GCD in the neutral setting. In general, FID, GMR, and GMC metrics do not exhibit a clearly strong correlation with the number of generated dancers but display diverse and varied results. The MMC metric consistently shows its stability across all setups. 

As the number of generated dancers increases, the generation diversity (GenDiv) decreases while the trajectory intersection frequency (TIF) increases. However, it is worth noting that the differences observed in these metrics are relatively minor compared to those produced by GDanceR~\cite{le2023music}. This implies that our method can effectively control consistency and diversity, significantly reducing the chances of collisions between dancers and maintaining the overall quality of generated group dance motions.

For a detailed visualization of results, please refer to Figure~\ref{fig:numsDancers} and our accompanying supplementary video. These results underscore the robustness and flexibility of our method in generating group dance motions across varying numbers of dancers while ensuring consistency, diversity, and avoiding collisions between performers.

\begin{table}[!t]
\centering
\caption{Performance of group dance generation methods when we increase the number of generated dancers, compared with GDanceR. In GCD setup, Neutral mode with $\gamma = 0$ is used.}
\resizebox{\linewidth}{!}{
\setlength{\tabcolsep}{0.3 em} % for the horizontal padding
{\renewcommand{\arraystretch}{1.2}% for the vertical padding
\begin{tabular}{c|c|c|c|c|c|c|c}
  \hline
\textbf{Method} & \textbf{\begin{tabular}[c]{@{}c@{}}\#Generated\\ Dancers\end{tabular}} & FID$\downarrow$ & MMC$\uparrow$ & GenDiv$\uparrow$ & GMR$\downarrow$ & GMC$\uparrow$ & TIF$\downarrow$ \\   \hline
\multirow{4}{*}{GDanceR} 
 & 2 & 48.82  & 0.248  & 9.36  & 53.83  &75.44  & 0.086 \\
 & 3 & 44.47  & 0.245  & 9.36  & 55.85  & 74.07 & 0.104 \\ 
 & 4 & 47.32  & 0.248  & 9.24  & 58.79  & 77.71 & 0.162 \\ 
 & 5 & 44.19  & 0.249  & 8.99  & 55.05  & 78.72 & 0.218 \\   \hline\hline
\multirow{4}{*}{\textbf{\begin{tabular}[c]{@{}c@{}}GCD\\ (Ours)\end{tabular}}} 
 & 2 & 32.62 & 0.266 & 10.41 & 34.09 & 80.26 & 0.067 \\ 
 & 3 & 33.94 & 0.266 & 10.02 & 36.25 & 79.93 & 0.084 \\ 
 & 4 & 35.89 & 0.251 & 9.87 & 36.28 & 81.82 & 0.125 \\ 
 & 5 & 35.08 & 0.264 & 9.92  & 38.43 & 81.44 & 0.168 \\   \hline
\end{tabular}
}}
\vspace{2ex}
    \label{tab:n_dancers}
\vspace{-1ex}
\end{table}

\begin{figure}[t]
   \centering
\resizebox{\linewidth}{!}{
\setlength{\tabcolsep}{2pt}
\begin{tabular}{ccccc}

\shortstack{\rotatebox[origin=l]{90}{\hspace{0.5 cm}  \#2
}}&
\shortstack{\includegraphics[width=0.33\linewidth]{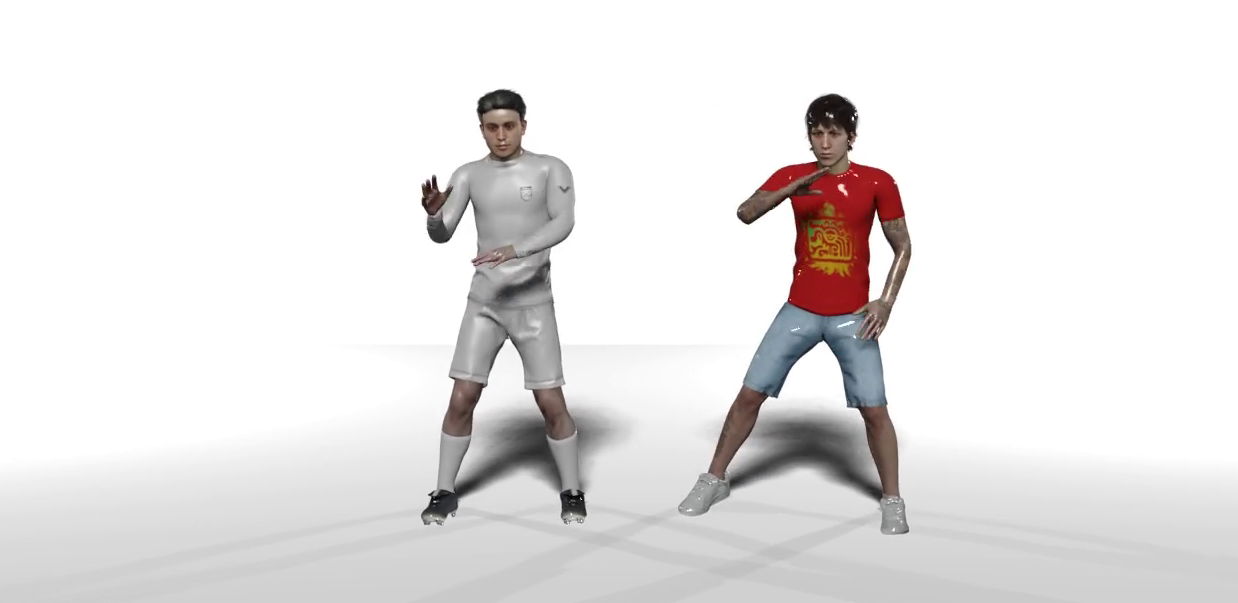}}&
\shortstack{\includegraphics[width=0.33\linewidth]{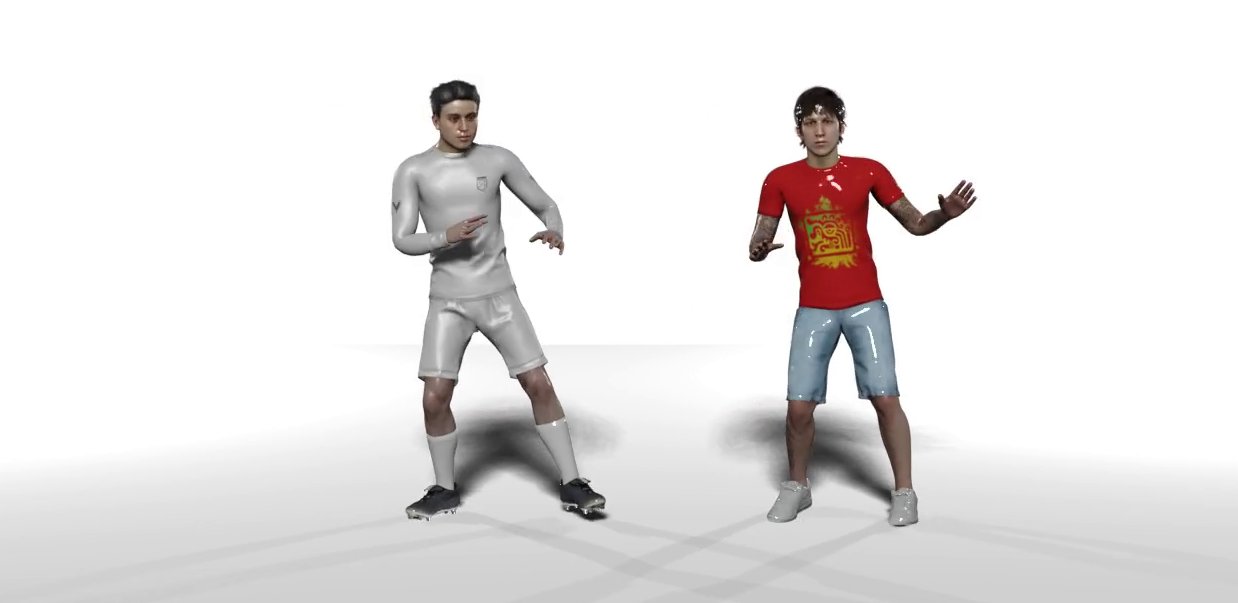}}&
\shortstack{\includegraphics[width=0.33\linewidth]{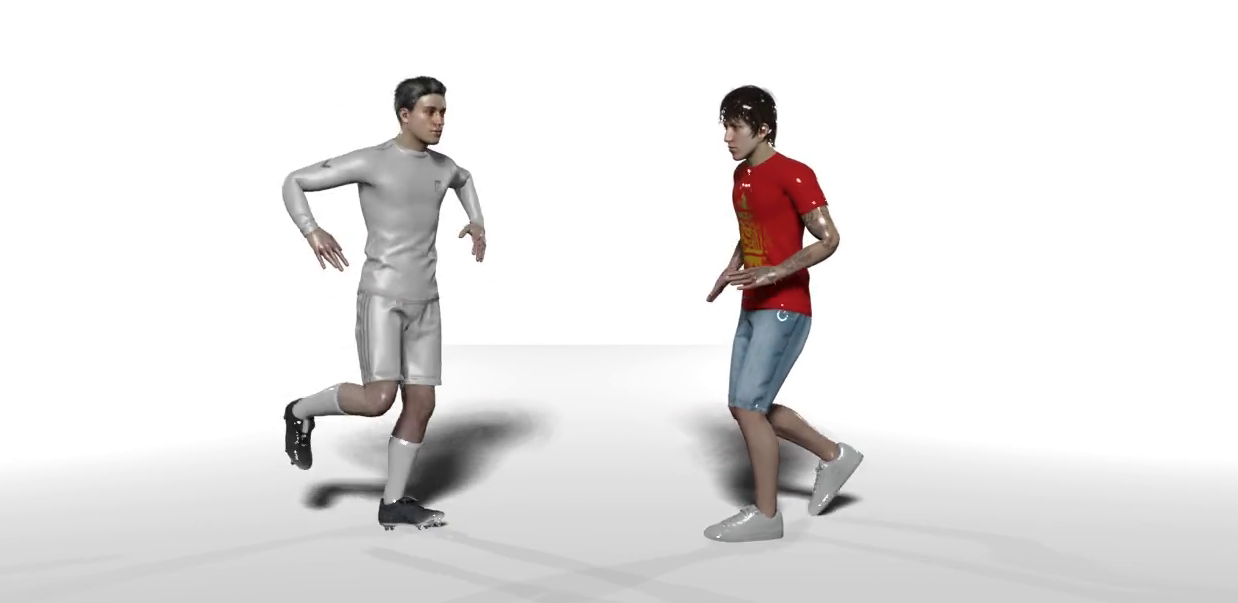}}\\[15pt]
%   ine\\
\shortstack{\rotatebox[origin=l]{90}{\hspace{0.5 cm}  \#3
}}&
\shortstack{\includegraphics[width=0.33\linewidth]{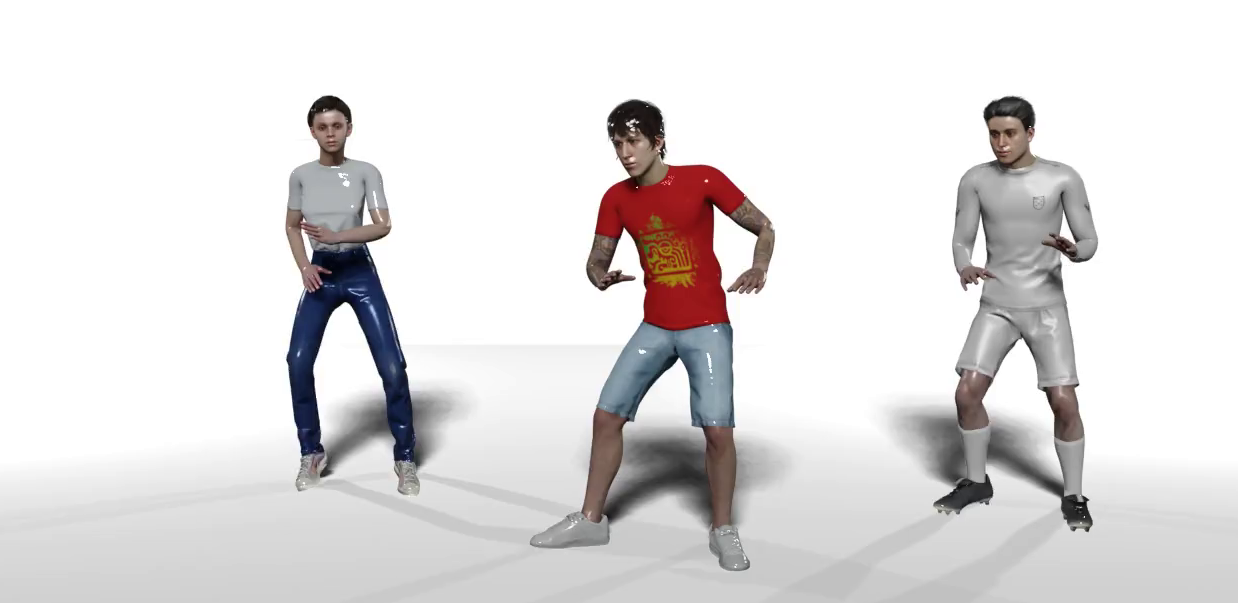}}&
\shortstack{\includegraphics[width=0.33\linewidth]{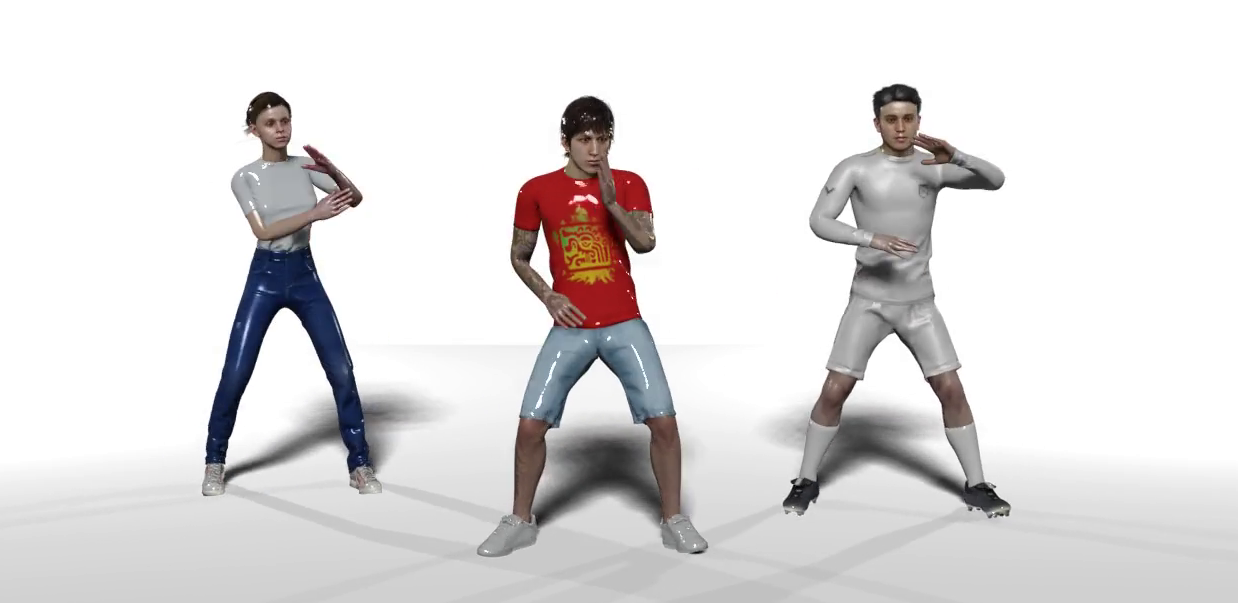}}&
\shortstack{\includegraphics[width=0.33\linewidth]{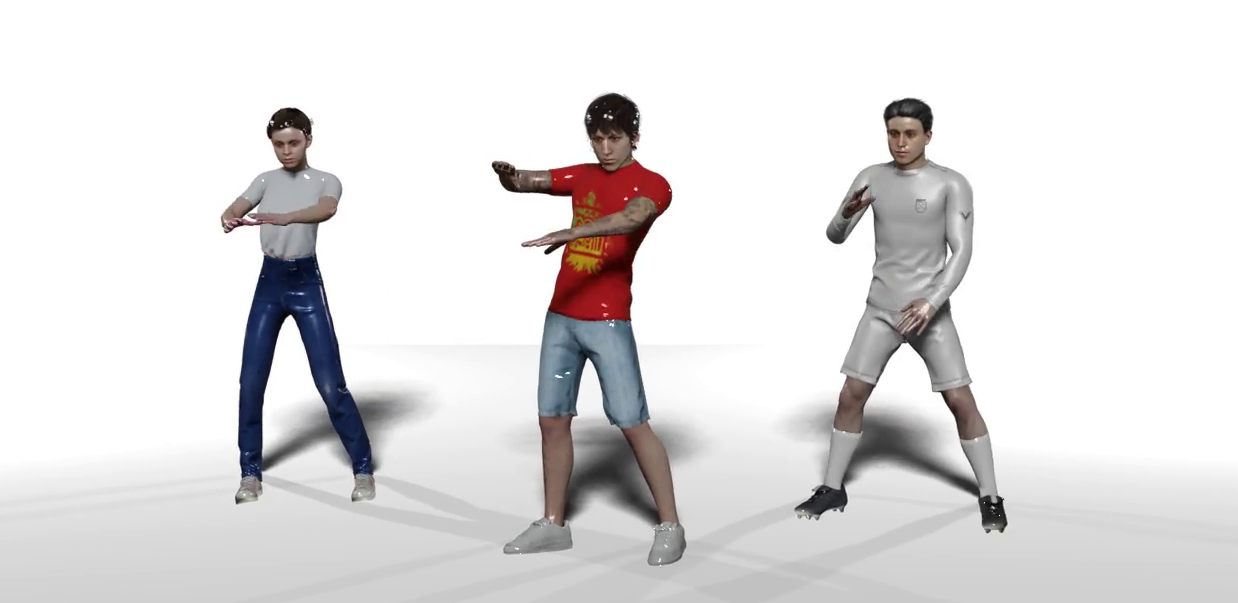}}\\[15pt]
\shortstack{\rotatebox[origin=l]{90}{\hspace{0.5 cm}  \#4
}}&
\shortstack{\includegraphics[width=0.33\linewidth]{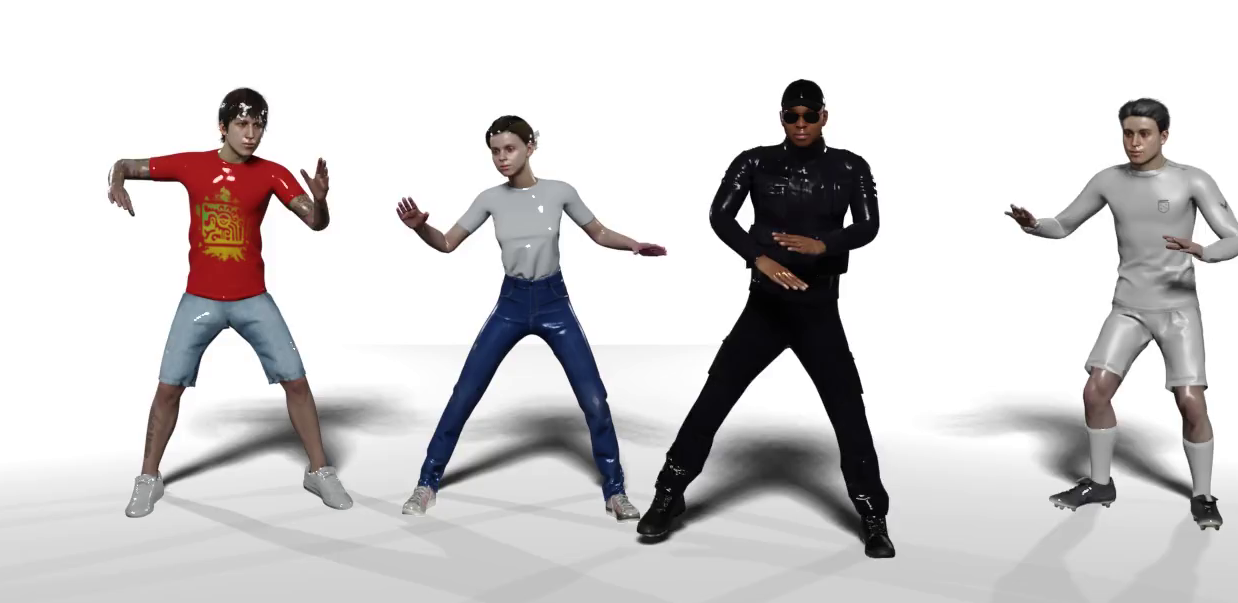}}&
\shortstack{\includegraphics[width=0.33\linewidth]{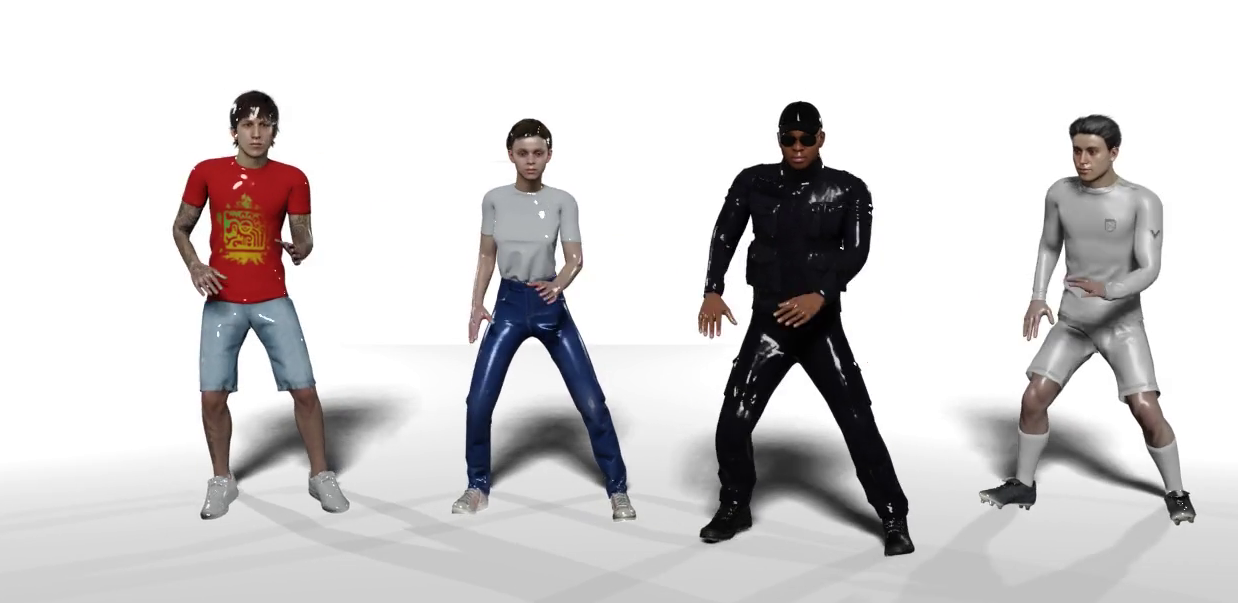}}&
\shortstack{\includegraphics[width=0.33\linewidth]{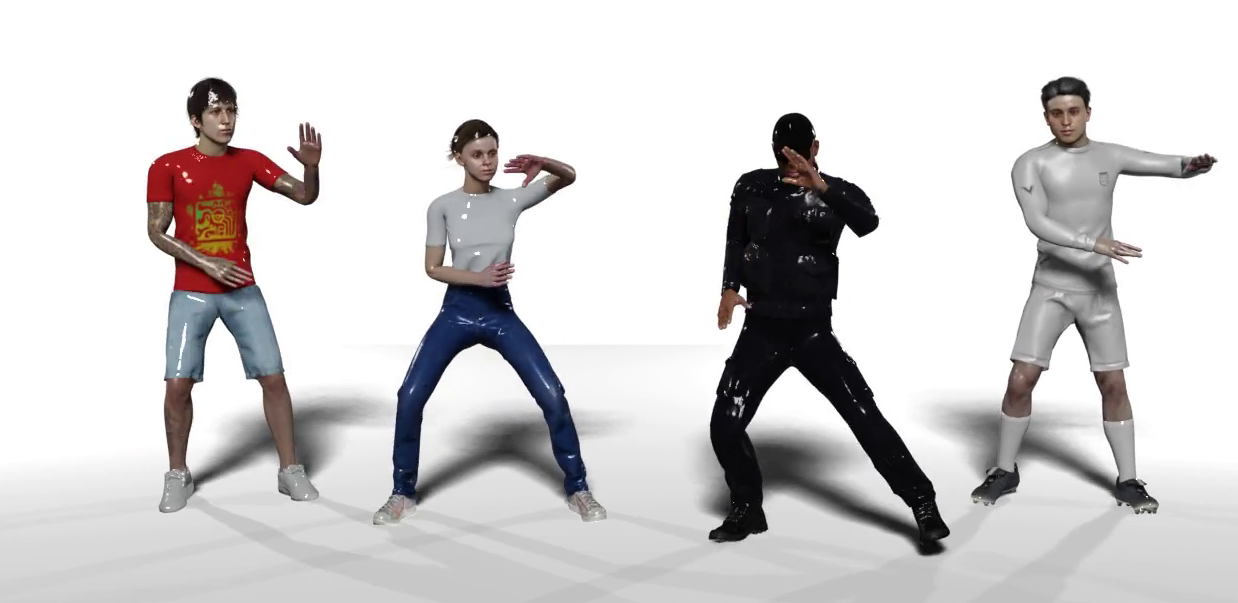}}\\[15pt]
%   ine\\
\shortstack{\rotatebox[origin=l]{90}{\hspace{0.5 cm} \#5
}}&
\shortstack{\includegraphics[width=0.33\linewidth]{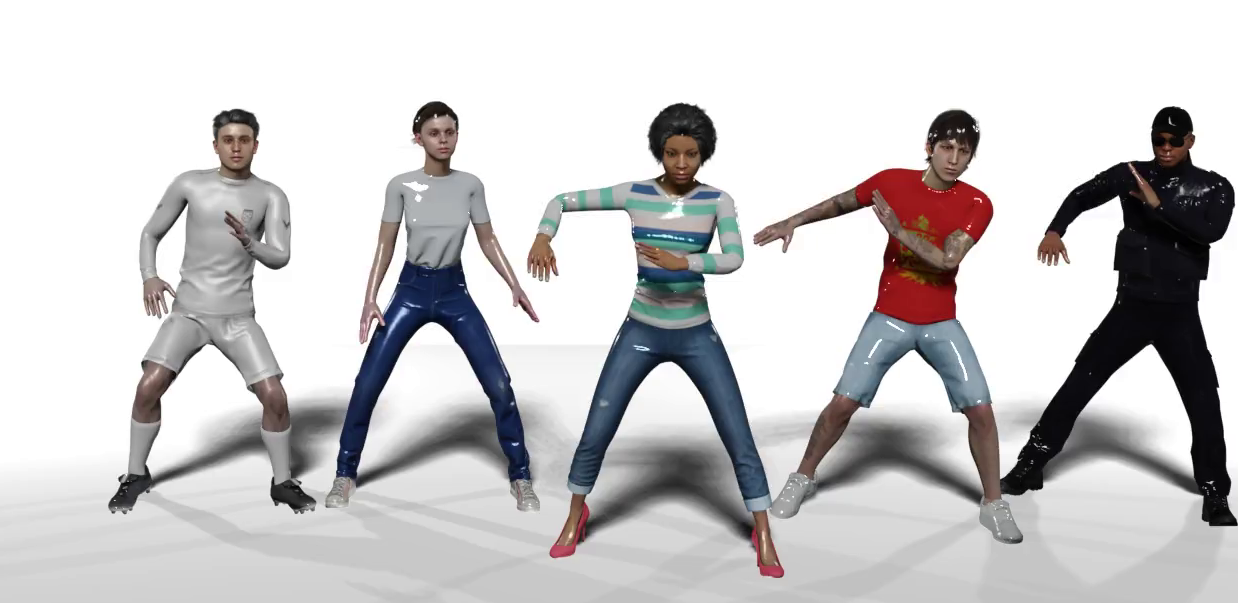}}&
\shortstack{\includegraphics[width=0.33\linewidth]{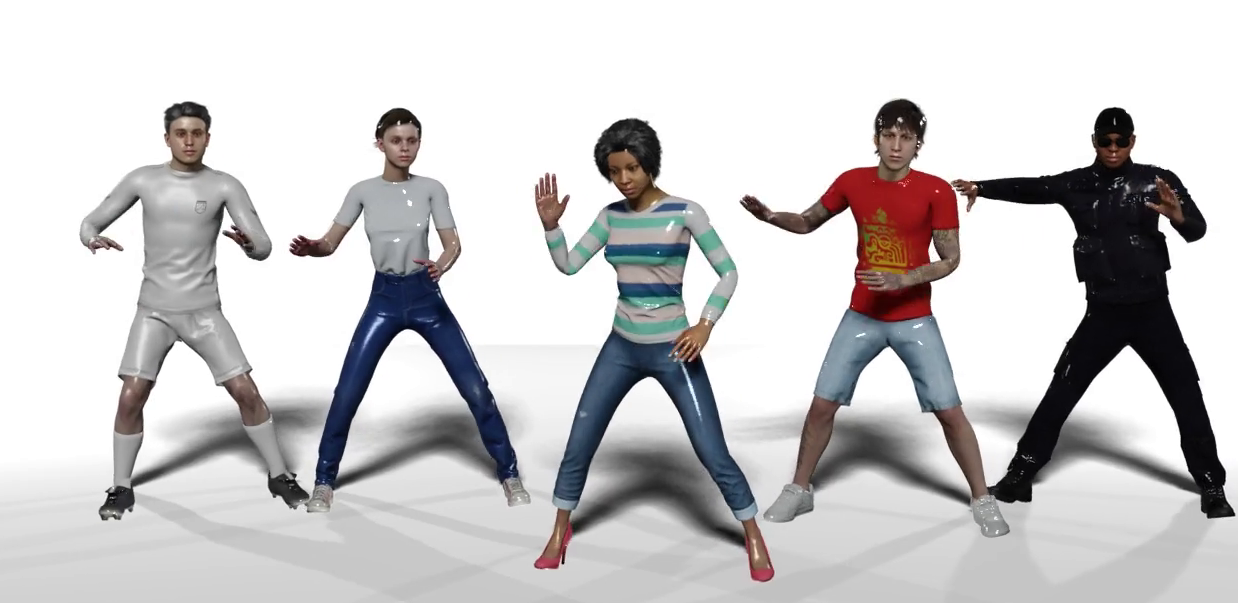}}&
\shortstack{\includegraphics[width=0.33\linewidth]{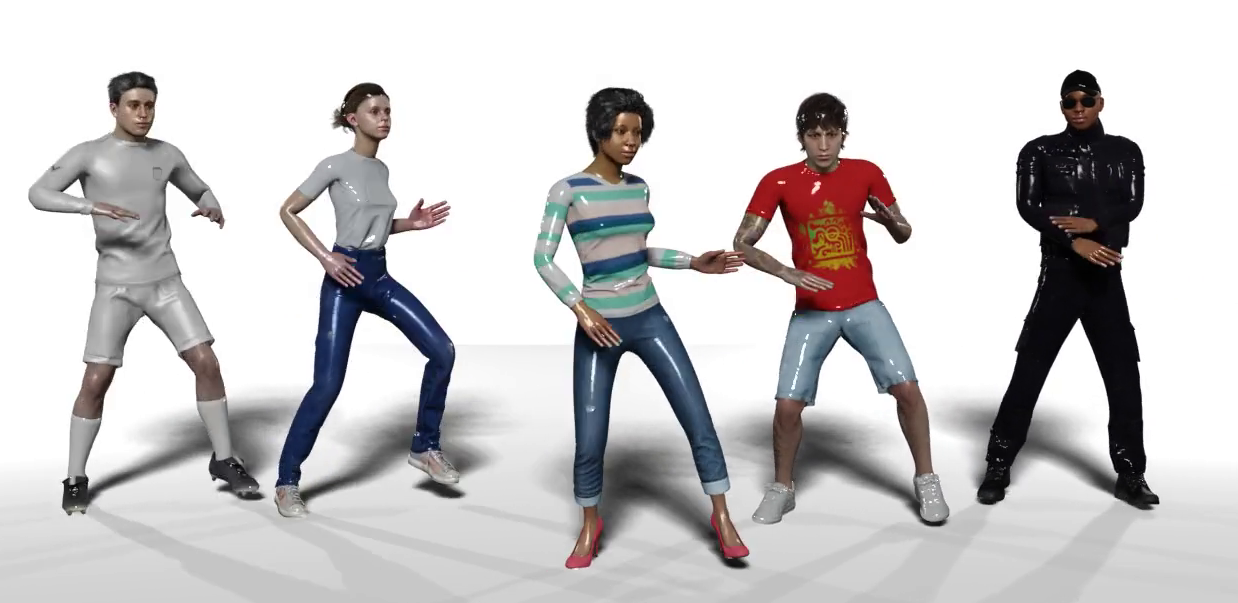}}\\[1pt]
\end{tabular}
}
    \caption{Group dance generation results of GCD in terms of different numbers of dancers.}
    \label{fig:numsDancers}
\end{figure}

\subsubsection{Long-term Analysis}

\begin{figure}[t]
   \centering
\resizebox{\linewidth}{!}{
\setlength{\tabcolsep}{2pt}
\begin{tabular}{ccccc}

\shortstack{\rotatebox[origin=l]{90}{\hspace{0 cm}  GDanceR
}}&
\shortstack{\includegraphics[width=0.33\linewidth]{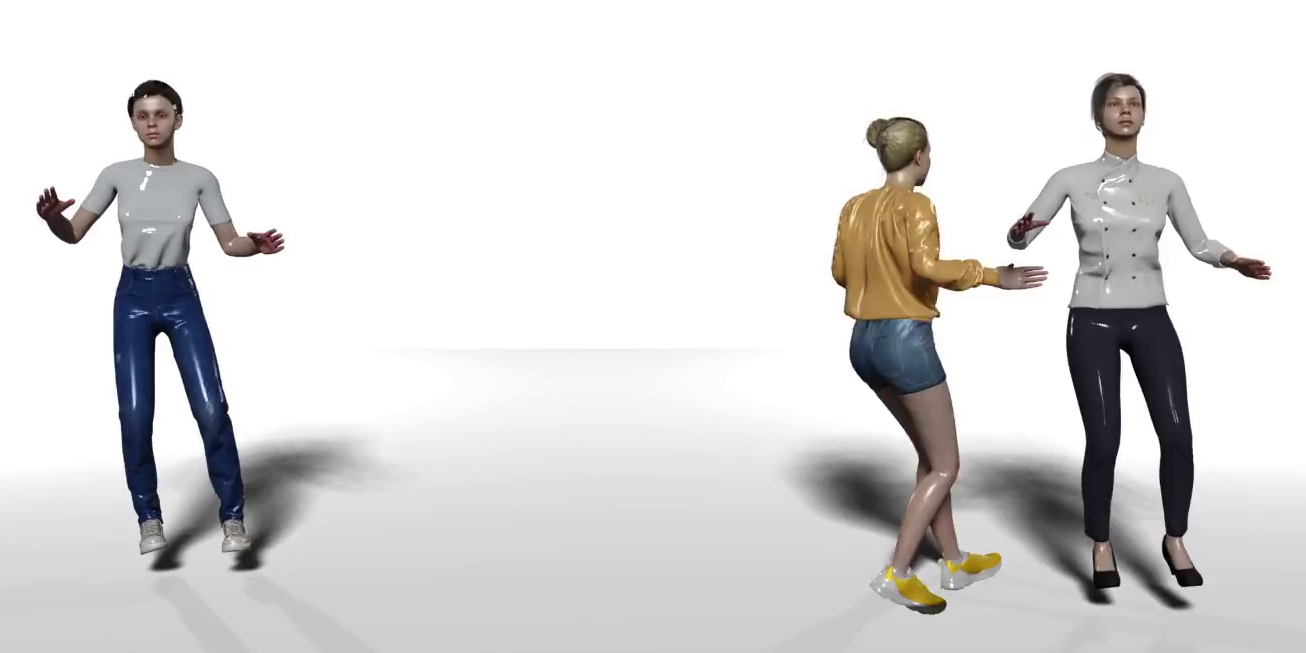}}&
\shortstack{\includegraphics[width=0.33\linewidth]{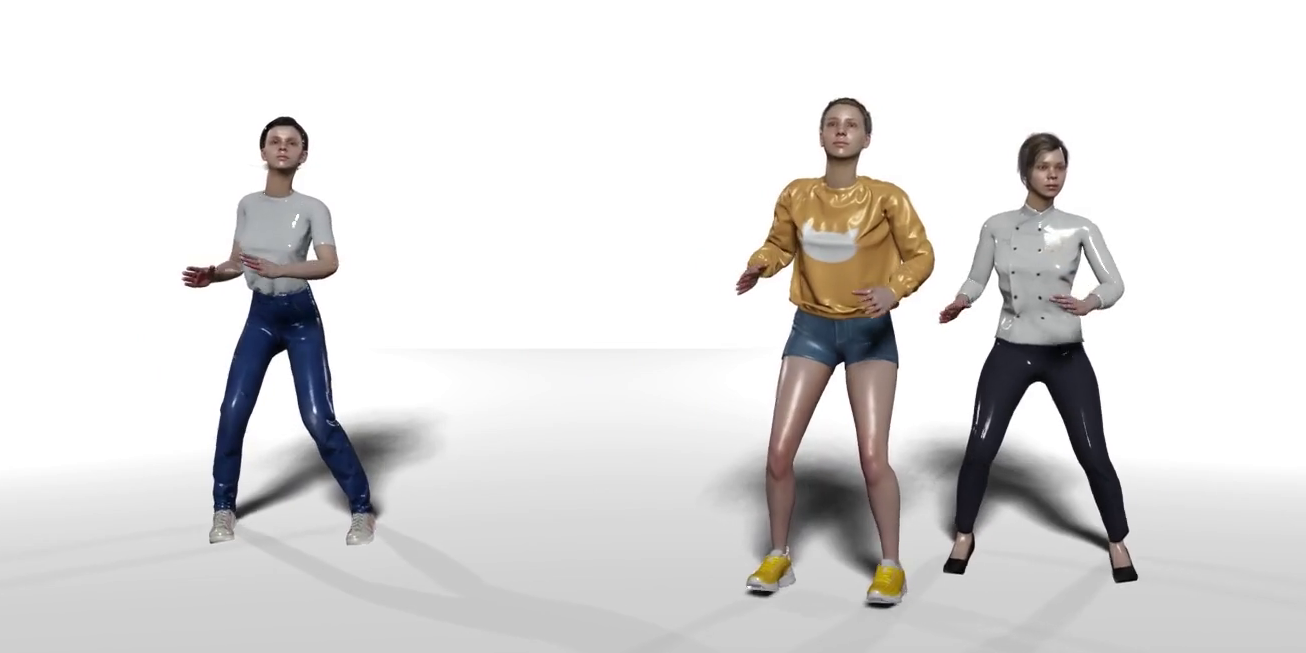}}&
\shortstack{\includegraphics[width=0.33\linewidth]{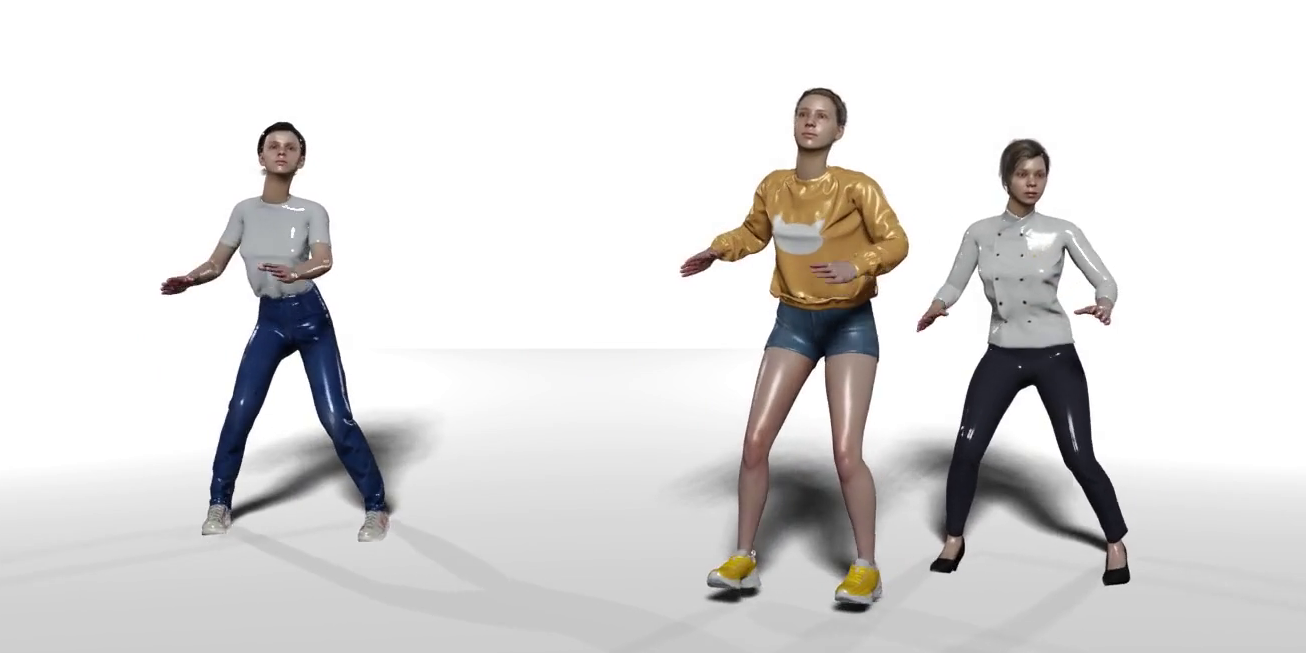}}&
\shortstack{\includegraphics[width=0.33\linewidth]{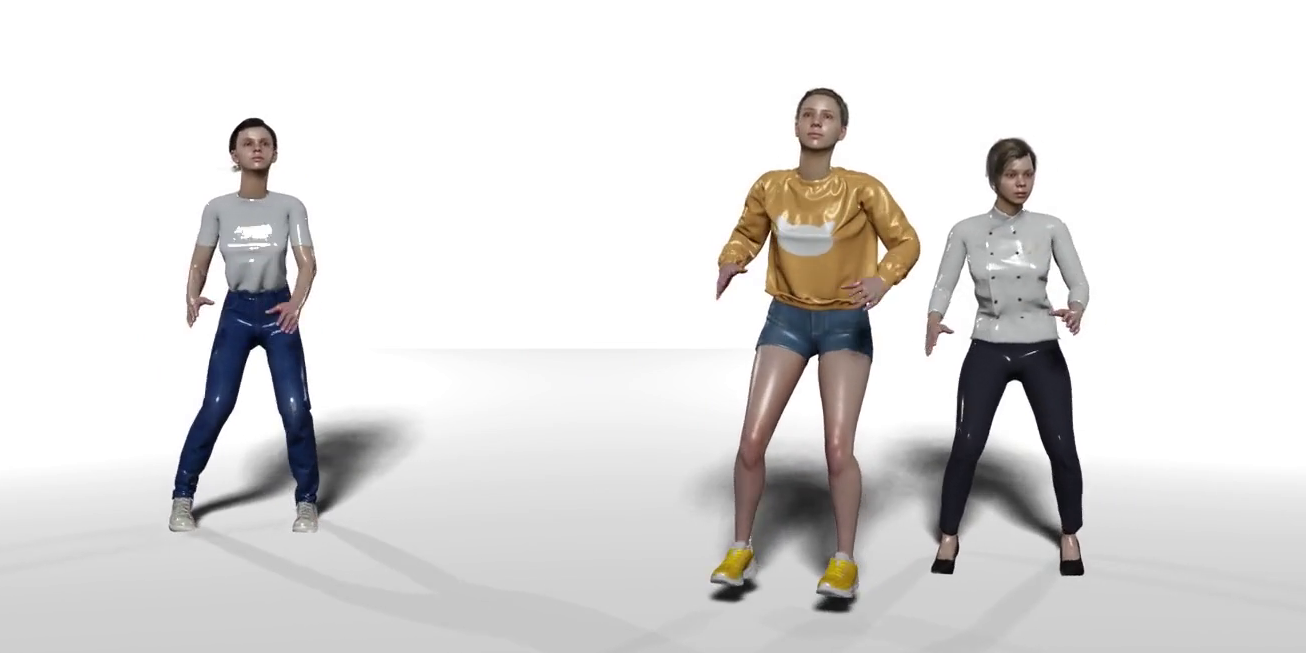}}\\[15pt]
\shortstack{\rotatebox[origin=l]{90}{\hspace{0 cm}  GCD
}}&
\shortstack{\includegraphics[width=0.33\linewidth]{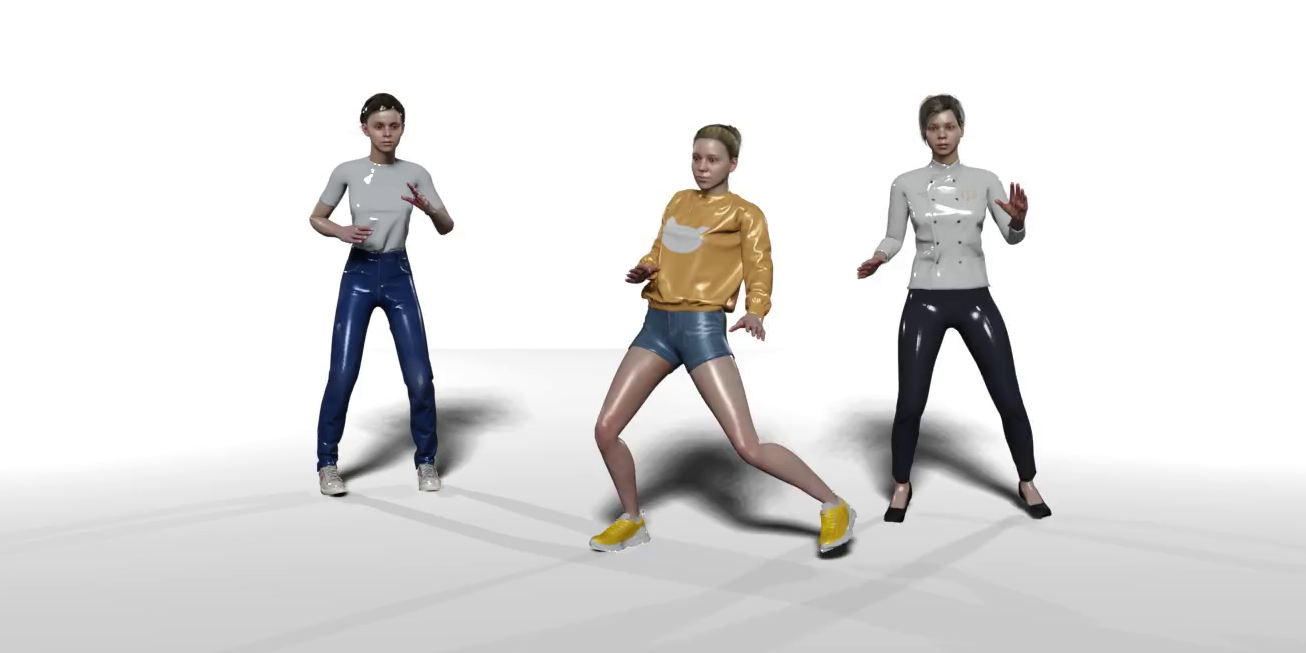}}&
\shortstack{\includegraphics[width=0.33\linewidth]{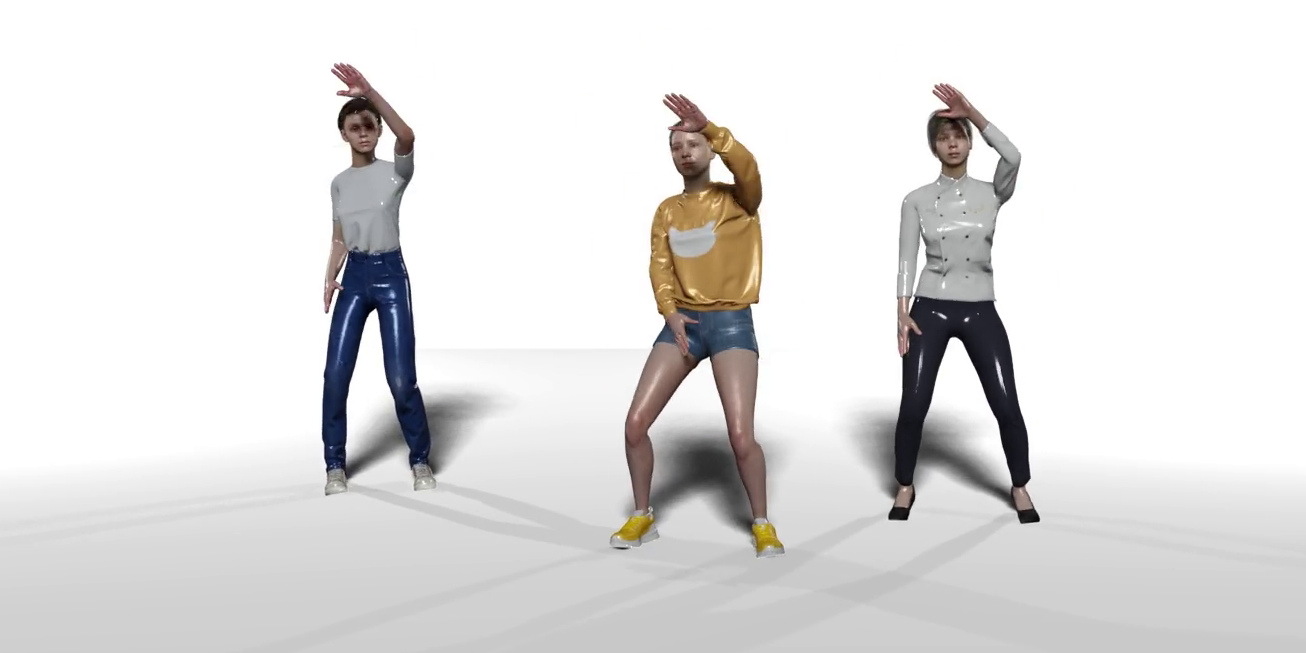}}&
\shortstack{\includegraphics[width=0.33\linewidth]{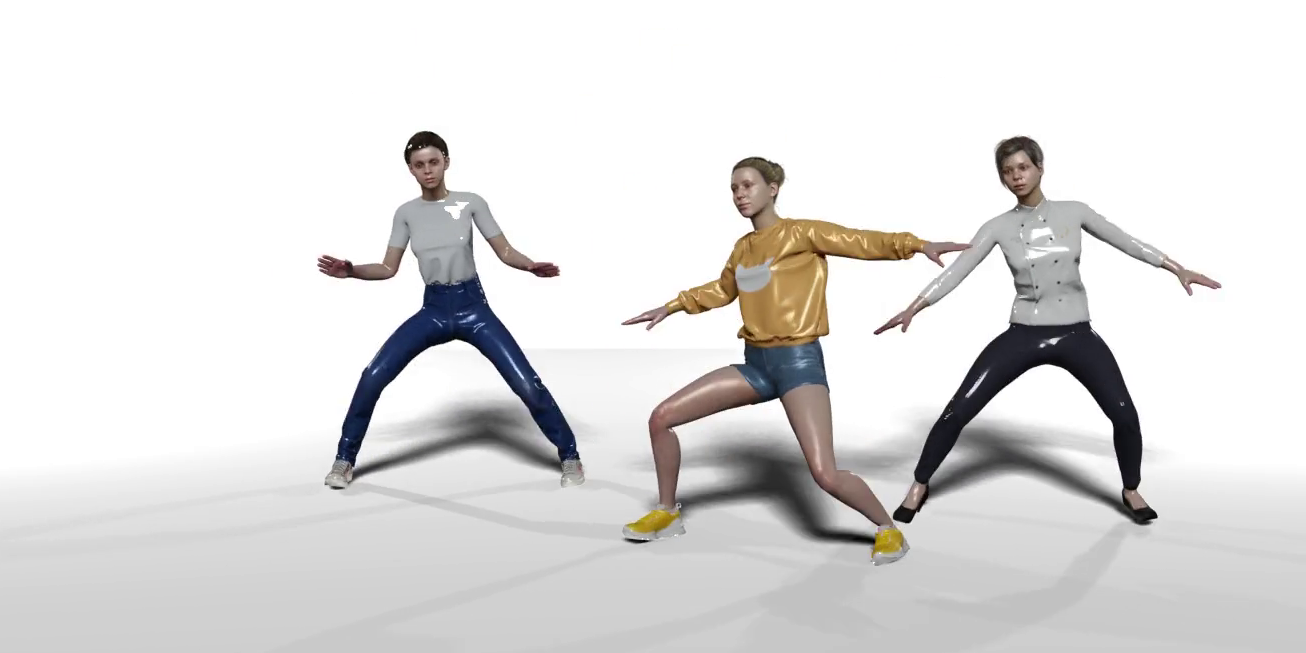}}&
\shortstack{\includegraphics[width=0.33\linewidth]{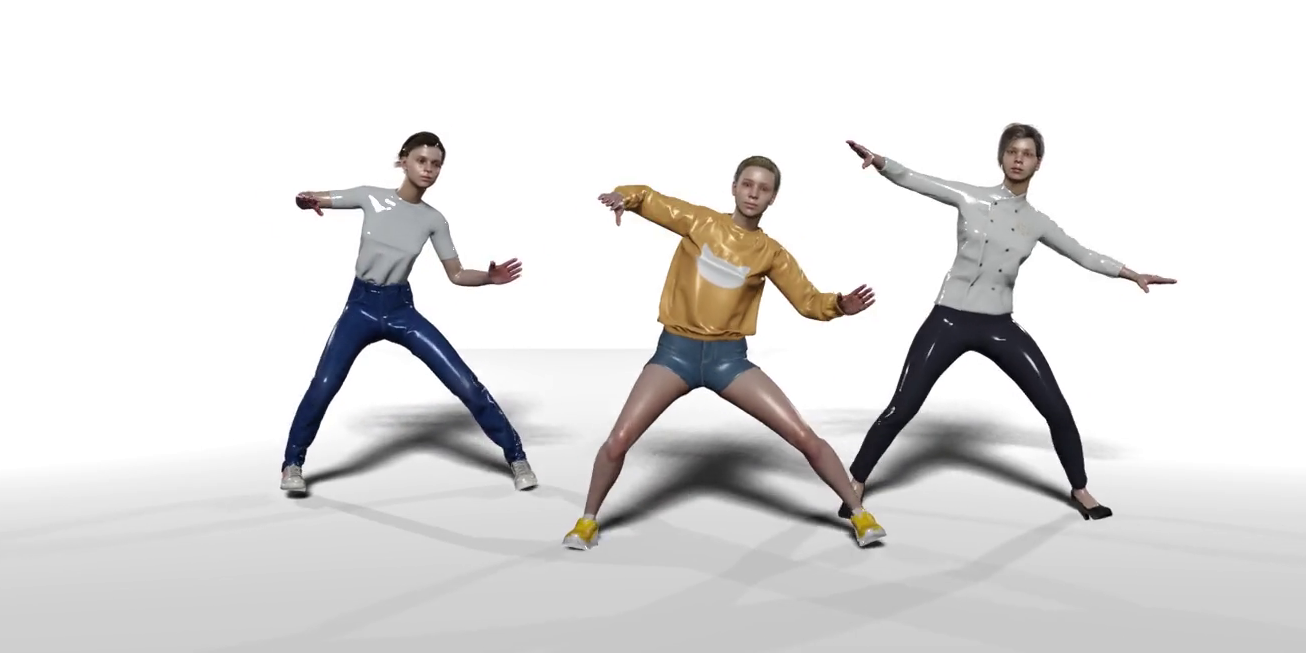}}\\[1pt]
\end{tabular}
}
    \caption{Long-term results of the 60-second clip. For clearer visualization, please visit our demo video.}
    \label{fig:longTerm}
\end{figure}

To evaluate the efficacy of the guidance signal in GCD for creating long-term group dance sequences, we conducted a comparative analysis between GCD and the baseline model GDanceR~\cite{le2023music}. The experiment involved musical pieces of different durations: 15 seconds, 30 seconds and 60 seconds. We show the results with the guidance parameter $\gamma = 0.5$ to enforce consistency with the music over a long duration. For more detailed information, please refer to Figure~\ref{fig:longTerm} and our supplementary video.

\begin{figure}[!t]
    \centering
    \includegraphics[width=0.47\textwidth, keepaspectratio=true]{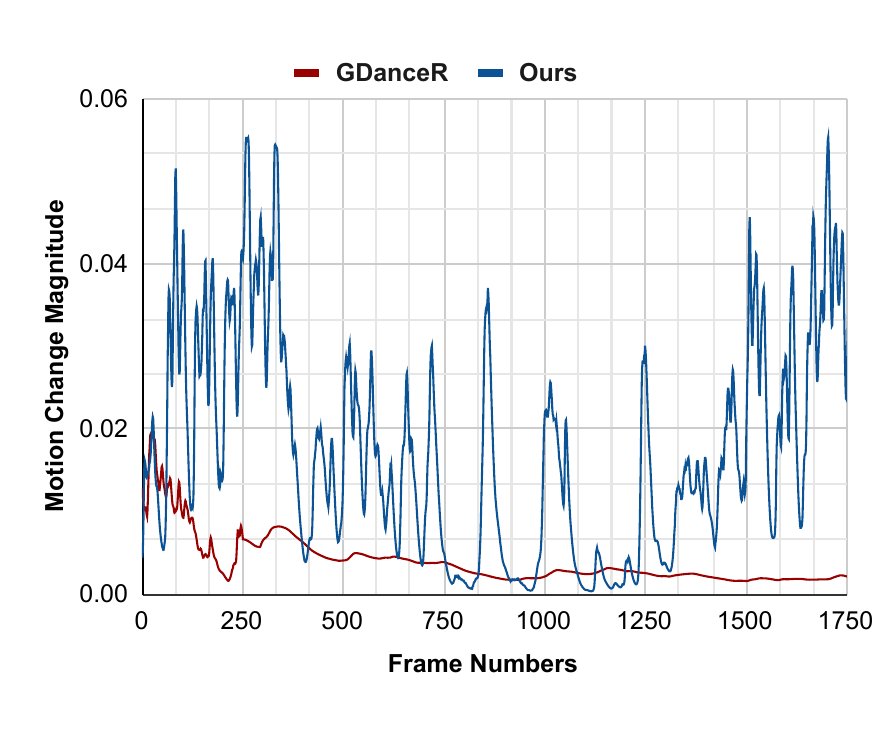}
    \vspace{-0.4cm}
    \caption{Motion changes comparison between our proposed GCD and GDanceR. The experiment is conducted on generated group dance results of $60$-second music pieces.}
    \label{fig:longTermGenDiv}
\end{figure}

While both methods produce satisfactory results in the first few seconds of the animations (e.g., about 5-6 seconds), GDanceR starts to exhibit floating and unrealistic movements or freeze into a mean pose in the later period of the sequence. Figure~\ref{fig:longTermGenDiv} shows the Motion changes comparison between our GCD method and GdanceR. The motion change magnitudes are calculated as average differences of the kinetic features~\cite{onuma2008fmdistance} between consecutive frames. It is evident that the motion change magnitude of GDanceR is gradually lower and approaching zero in the later half of the 60-second music piece, whereas our method can preserve high magnitudes and variations over time. This is because GDanceR generates almost frozen dance choreographies during this period. In contrast, the group dance motions produced by GCD remain natural with diverse movements throughout the entire duration of all music samples.

These findings confirm that our approach can effectively address the problem of  motion generation in long-horizon group dance scenarios. It maintains the motion quality and dynamics of the dance motions, ensuring that the created animations remain visually appealing throughout extended periods. This highlights the advantage of the contrastive strategy to enhance the consistency of the movements of dancers with their group and the music, resulting in significant improvements for long-term dance sequence generation compared to the baseline GDanceR.

\subsubsection{Ablation Analysis}\hfill\\
\textbf{Loss Terms.} The contribution of the geometric loss $\mathcal{L}_{\rm geo}$ and contrastive loss $\mathcal{L}_{\rm nce}$ in GCD is thoroughly analyzed and presented in Table~\ref{tab:lossAnalysis}. The results demonstrate that both losses play a crucial role in enhancing the overall performance across all four evaluation protocols. In particular, it can be seen that the effect of $\mathcal{L}_{\rm geo}$ on realism metrics (FID, GMR, and PFC) is significant. This observation can be attributed to the fact that this loss improves the physical plausibility and naturalness of the dance motions, empirically mitigating common artifacts such as jittery motion or foot skating. By enforcing the geometric constraints, GCD can generate faithful motions that are on par with real dances. Moreover, the contrastive loss $\mathcal{L}_{\rm nce}$  contributes positively to the favorable results in the synchrony measures (GMR and GMC). This loss term encourages the model to synchronize the movements of multiple dancers within a group, thus improving the harmonious coordination and cohesion of the generated choreographies. In general, the results of $\mathcal{L}_{\rm geo}$ and $\mathcal{L}_{\rm nce}$ validate their importance across various evaluation metrics.

\textbf{Group Global Attention.}
Results presented in first two lines of Table~\ref{tab:lossAnalysis}  demonstrate substantial improvements obtained by incorporating Group Global Attention into GCD. It clearly shows that without the Group Global Attention, the performance on the group dance metrics (GMR and GMC) is significantly degraded. We also observe that the removal of this block resulted in inconsistent movements across dancers, where they seem to dance in freestyle without any group choreographic rules and collide with each other in many cases, although they may still follow the rhythm of the music. Results suggest the vital importance of ensuring coherency and regulating collisions for visually appealing group dance animations.

\begin{table}[!t]
\centering
\caption{Global module contribution and loss analysis. Experiments are conducted on GCD with $\gamma = 0$ (neutral mode).}
\resizebox{\linewidth}{!}{
\setlength{\tabcolsep}{0.2 em} % for the horizontal padding
{\renewcommand{\arraystretch}{1.2}% for the vertical padding
\begin{tabular}{lc|ccc|cc}
\hline
\multicolumn{2}{c|}{\textbf{Method}} & \multicolumn{1}{c|}{FID$\downarrow$} & \multicolumn{1}{c|}{MMC$\uparrow$} & \multicolumn{1}{c|}{  {PFC}$\downarrow$} & \multicolumn{1}{c|}{GMR$\downarrow$} & \multicolumn{1}{c}{GMC$\uparrow$} \\  \hline
\multicolumn{2}{l|}{\textbf{GCD}} & \multicolumn{1}{c|}{31.16} & \multicolumn{1}{c|}{0.261} & \multicolumn{1}{c|}{2.53} & \multicolumn{1}{c|}{31.47} & \multicolumn{1}{c}{80.97} \\ 
\multicolumn{2}{l|}{GCD w/o Group Global Attention} & \multicolumn{1}{c|}{31.35} & \multicolumn{1}{c|}{0.263}  & \multicolumn{1}{c|}{2.62} & \multicolumn{1}{c|}{62.23} & \multicolumn{1}{c}{60.72}\\ 
\hline
\multicolumn{2}{l|}{GCD w/o $\mathcal{L}_{\rm geo}$}  & \multicolumn{1}{c|}{39.27} & \multicolumn{1}{c|}{0.254} & \multicolumn{1}{c|}{2.95} & \multicolumn{1}{c|}{37.73} & \multicolumn{1}{c}{80.11} \\ 
\multicolumn{2}{l|}{GCD w/o $\mathcal{L}_{\rm nce}$}  & \multicolumn{1}{c|}{35.10} & \multicolumn{1}{c|}{0.241} & \multicolumn{1}{c|}{2.57} & \multicolumn{1}{c|}{47.47} & \multicolumn{1}{c}{71.82} \\ 
\multicolumn{2}{l|}{GCD w/o  ($\mathcal{L}_{\rm geo}$ \& $\mathcal{L}_{\rm nce}$) } & \multicolumn{1}{c|}{40.99} & \multicolumn{1}{c|}{0.232} & \multicolumn{1}{c|}{2.98} & \multicolumn{1}{c|}{49.98} & \multicolumn{1}{c}{71.02} \\ 
 \hline
\end{tabular}
}}
\vspace{2ex}
    \label{tab:lossAnalysis}
\end{table}

\subsection{User Study}

Qualitative user studies are important for evaluating generative models
as the perception of users tends to be the most relevant metric for many downstream applications. Therefore, we conduct user studies to evaluate our approach in terms of group choreography generation. We organized two separate studies and enlisted roughly 50 individuals with diverse backgrounds to participate in our experiment.  Each participant should have some relevant experience in music and dance (at least 1 month of studying or working in dance-related professions). The age of participants varied
between 20 and 50, with approximately 55\% female and 45\% male.

In the initial study, we requested the participants to evaluate the dancing animations based on three criteria: the naturalness of the dancing motions (Realism), how well the movements match the music (Music-Motion Correspondence), and how well the dancers interact or synchronize with each other (Synchronization between Dancers). Participants were asked to rate scores from 0 to 10 for each criterion,   { ranging from \textit{(0)-very poor}, \textit{(5)-acceptable}, to \textit{(10)-very good}}. The collected scores were then normalized to range $[0,1]$.

\begin{figure}[!t]
    \centering
    \includegraphics[width=0.42\textwidth, keepaspectratio=true]{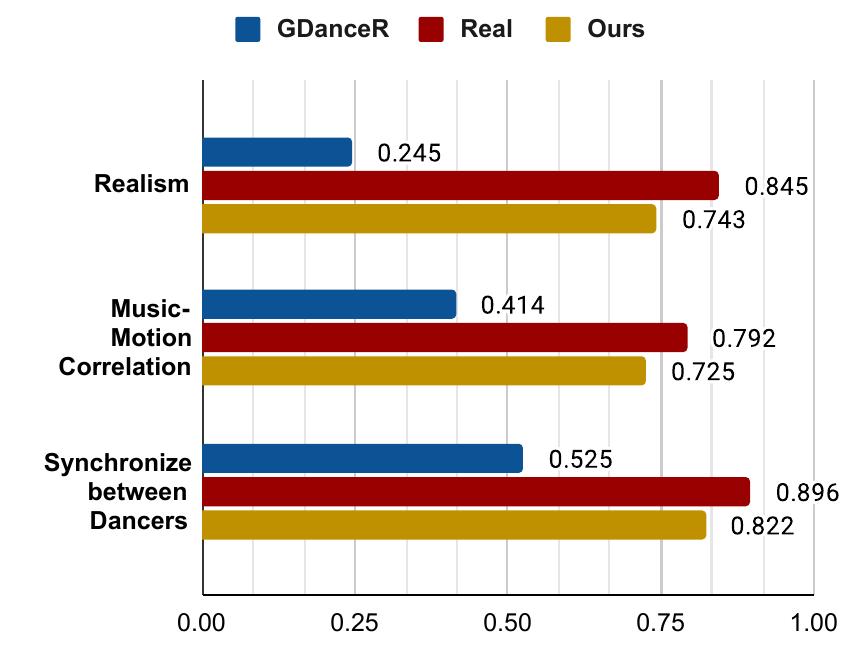}
    \caption{User study results in three criteria: Realism; Music-Motion Correspondence; and Synchronization between Dancers.}
    \vspace{-0.2cm}
    \label{fig:UserStudy1}
\end{figure}

This user study encompassed a total of $189 * 3$ samples with songs that are not present in the train set, including those generated from GDanceR~\cite{le2023music}, real dance clips from the dataset, and generated results from our proposed method in neutral mode. Figure~\ref{fig:UserStudy1} shows average scores for all mentioned targets across three experiments. Notably, the ratings of our method are significantly higher than GDanceR across all three criteria.   {We also perform Tukey honest significance tests to determine the significant differences among the three methods. For the first two criteria (Realism and Music-Motion Correspondence), we observe that the mean scores of all methods are significantly different with $p < 0.05$. For synchronization critera, the differences are significant except for the scores between our method and real dances ($p\approx0.07$). }
This highlights that our method can even achieve comparable scores with real dances, especially in the synchronization evaluation. This can be attributed to the proposed contrastive diffusion strategy, which can effectively maintain a balance between the consistency of the movements and the group/audio context, as well as diversity in generated dances.

In the second study, we aim to assess the diversity and consistency of the generated dance outputs and determine if they met the expectations of the users. Specifically, participants were asked to assign scores ranging from 0 to 10 to evaluate the consistency and diversity of each dance clip, i.e., how synchronized or how distinctive movements between dancers does the group dance present. A lower score indicated higher consistency, while a higher score indicated greater diversity. These scores were subsequently normalized to $[-1,1]$ to align with the studying range of the control parameter employed in our proposed method.

\begin{figure}[!t]
    \centering
    \includegraphics[width=0.40\textwidth, keepaspectratio=true]{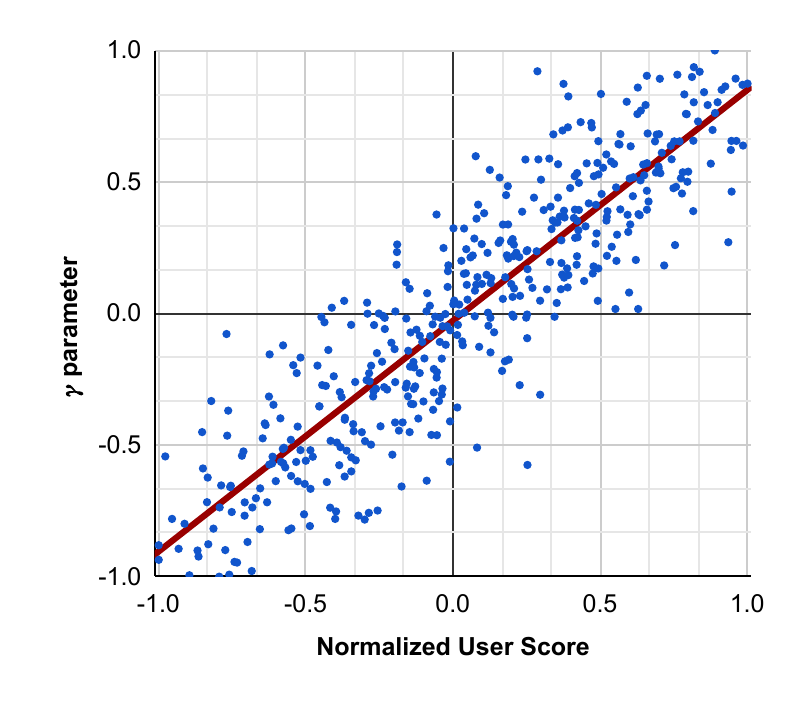}
        \vspace{-0.4 cm}
    \caption{Correlation between the controlling consistency/diversity and the  scores provided by the users .}
    \vspace{-0.2 cm}
    \label{fig:UserStudy2}
\end{figure}

Figure~\ref{fig:UserStudy2} depicts a scatter plot illustrating the relationship between the scores provided by the participants and the $\gamma$ parameter that was used to generate the dance samples. The parameter values were randomly drawn from a uniform distribution with range $[-1,1]$ to create the animations along with randomly sampled musical pieces. The survey shows a strong correlation between the user scores and the control parameter, in which we calculated the correlation coefficient to be approximately 0.88.  The results indicate that the diversity and consistency level of the generated group choreography samples is mostly in agreement with the user evaluation, as indicated by the scores obtained.
 
\section{Discussion and Conclusion}

While controlling consistency and diversity in group dance generation by using our proposed GCD has numerous advantages and potentials, there are certain limitations.
Firstly, it requires tuning the parameters and a complex system that is not trivial to train, to ensure that the generated dance motions can produce the desired level of similarity among dancers while still presenting enough variation to avoid repetitive or monotonous movements. This may involve long inference processes and may require significant computational resources in both the training and testing phases. 

Secondly, over-controlling consistency and diversity may introduce constraints on the creative freedom of generated dances. While enforcing consistency can lead to synchronized and harmonious group movements, it may limit the possibility of exploring unconventional or new experimental dance styles. On the other hand, promoting diversity results in unique and innovative dance sequences, but it may sacrifice coherence and coordination among dancers.

  {Although our model can synthesize semantically faithful group dance animation with effective coordination among dancers, it does not capture clear physical contact between dancers such as hand touching. This is because the data we used in training does not contain such detailed hand motion information. We think that exploring group dance with realistic physical hand interactions is a promising area for future work. Additionally, while our method offers a trade-off between diversity and consistency, achieving perfect alignment between high-diversity movements and music remains a challenging task. The diversity level among dancers and the alignment with music are also heavily influenced by the training data. We believe further efforts are required to reach this.}

Lastly, the subjective nature of evaluating consistency and diversity poses a challenge. Metrics for measuring these aspects may not be best fitted. We believe it is essential to consider diverse perspectives and demand domain experts to validate the effectiveness and quality of the generated dance motions.

To conclude, we have introduced GCD, a new method for audio-driven group dance generation that effectively controls the consistency and diversity of generated choreographies. By using contrastive diffusion along with the guidance technique, our approach enables the generation of a flexible number of dancers and long-term group dances without compromising fidelity. Through our experiments, we have demonstrated the capability of GCD  to produce visually appealing and synchronized group dance motions. The results of our evaluation, including comparisons with existing methods, highlight the superior performance of our method across various metrics including realism and synchronization. By enabling control over the desired levels of consistency and diversity while preserving fidelity, our work has the potential for applications in entertainment, virtual performances, and artistic expression, advancing the effectiveness of deep learning in generative choreography.


% \begin{acks}
% Rap Video and Vocal: Lethal, M. (2017, November 20). Mac Lethal – 400 Words in 1 Minute (Fast Rap). Genius. Retrieved October 28, 2021, from https://genius.com/Mac-lethal-400-words-in-1-minute-fast-rap-lyrics

% Opera Video and Vocal: Symphony, H. (2017, October 24). Andrea Bocelli: In His Own Words. Houston Symphony. Retrieved November 18, 2021, from https://houstonsymphony.org/andrea-bocelli-in-his-own-words/

% Ballad Video: 
% Adele: NPR Music Tiny Desk Concert. (2011, February 14). YouTube. Retrieved October 11, 2021, from https://www.youtube.com/watch?v=XfzpYcwiUrA

% Easy on Me Song: Adele - Easy On Me (Official Video). (2021, October 14). YouTube. Retrieved November 12, 2021, from https://www.youtube.com/watch?v=U3ASj1L6_sY
% \end{acks}
% \nocite{*}

\bibliographystyle{ACM-Reference-Format}
\bibliography{acmart}

% %%
% %% If your work has an appendix, this is the place to put it.
% \appendix



% \lipsum[1]


% \lipsum[1]

% \lipsum[1]
\end{sloppypar}
\end{document}